%% file: neurips_2026.tex
\documentclass{article}

\usepackage[preprint]{neurips_2026}
\setcitestyle{numbers,square,comma}

\usepackage[utf8]{inputenc}
\usepackage[T1]{fontenc}
\usepackage{hyperref}
\usepackage{url}
\usepackage{booktabs}
\usepackage{amsfonts}
\usepackage{amsmath}
\usepackage{nicefrac}
\usepackage{microtype}
\usepackage[table]{xcolor}
\usepackage{graphicx}
\usepackage{subcaption}
\usepackage{multirow}
\usepackage{enumitem}
\usepackage{wrapfig}
\usepackage{enumitem}
\usepackage{pifont}
\usepackage{float}
\usepackage[most]{tcolorbox}

\definecolor{codeblue}{HTML}{2E86C1}
\definecolor{codered}{HTML}{C0392B}
\definecolor{codegreen}{HTML}{27AE60}
\newcommand{\ours}{\textsc{SimWorld Studio}}
\definecolor{checkgreen}{RGB}{30,160,80}
\definecolor{crossred}{RGB}{180,70,70}
\newcommand{\cmark}{\textcolor{checkgreen}{\ding{51}}}
\newcommand{\xmark}{\textcolor{crossred}{\ding{55}}}

\definecolor{rowgray}{HTML}{F7F7F7}
\usepackage{booktabs}
\usepackage{multirow}
\usepackage{makecell}
\usepackage{array}
\usepackage{graphicx}
\usepackage{siunitx}
\usepackage{algorithmicx}
\usepackage{algpseudocode}
\usepackage{tabularx}
\usepackage{listings}
\usepackage{needspace}

\lstdefinestyle{pythoncasestyle}{
    language=Python,
    basicstyle=\ttfamily\scriptsize,
    keywordstyle=\bfseries,
    commentstyle=\itshape,
    stringstyle=\ttfamily,
    showstringspaces=false,
    breaklines=true,
    breakatwhitespace=false,
    columns=fullflexible,
    keepspaces=true,
    frame=single,
    rulecolor=\color{black!18},
    backgroundcolor=\color{black!3},
    xleftmargin=1mm,
    xrightmargin=1mm,
    framexleftmargin=1mm,
    framexrightmargin=1mm,
    aboveskip=4pt,
    belowskip=6pt
}

\definecolor{standoutred}{HTML}{E74C3C}

\definecolor{darkgreen}{RGB}{0,100,0}

\definecolor{murraypink}{RGB}{220,80,160}

\newcommand{\ucsd}{$^{1}$}
\newcommand{\nyush}{$^{2}$}
\newcommand{\colead}{$^{*}$}

\title{\ours{}: \\ Automatic Environment Generation with Evolving Coding Agent for Embodied Agent Learning}

\author{%
    Haoqiang Kang\ucsd\colead\hspace{6pt}
    Xiaokang Ye\ucsd\colead\hspace{6pt}
    Yuhan Liu\nyush\hspace{6pt}
    Siddhant Hitesh Mantri\ucsd\hspace{6pt} \\
    {\bf
    Lingjun Mao\ucsd\hspace{6pt}
    James Fleming\ucsd\hspace{6pt}
    Drishti Regmi\ucsd\hspace{6pt}
    Lianhui Qin\ucsd} \\
    \ucsd UC San Diego\;\hspace{6pt}
    \nyush New York University
}

\begin{document}

\maketitle

\let\thefootnote\relax\footnotetext{*~Equal contribution.}



\begin{center}
    \vspace{-30pt}
    \includegraphics[width=\linewidth]{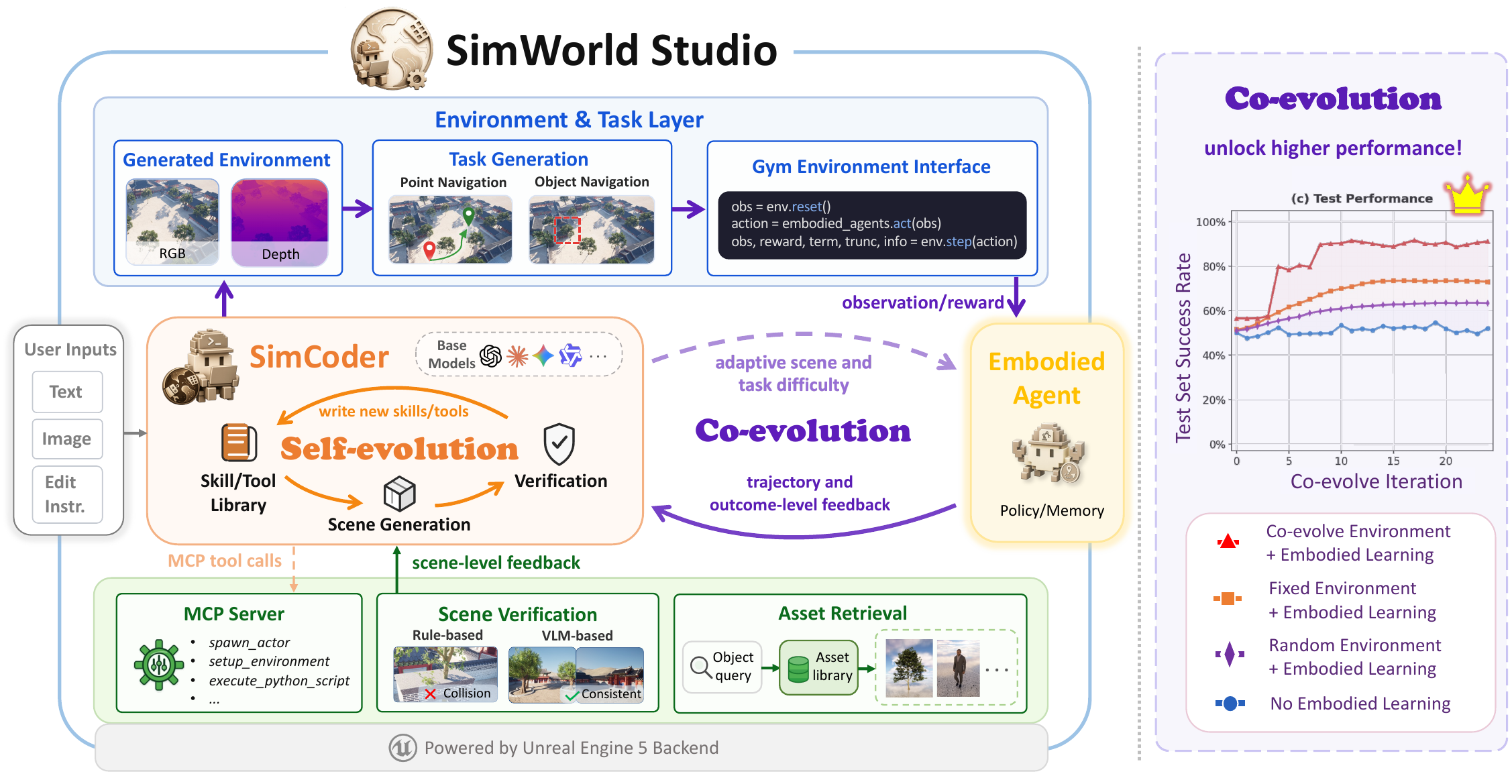}
    \captionof{figure}{
        \textbf{\ours{}:}
        {\bf (Left)} \textsc{SimCoder} automatically generates UE5 interactive environments with  realistic 3D scenes, learning tasks, and Gym interfaces. 
        {\bf (Right)} Co-evolving environment generation with embodied learning substantially improves test success over both fixed-environment training and the untrained-agent baseline.
    }
    \vspace{-1mm}
    \label{fig:teaser}
\end{center}

\input{Section/0-Abstract}


\input{Section/1-Introduction}
\input{Section/3-SimWorld_Studio}


\input{Section/4-Experiment}

\input{Section/2-Related_Work}

\input{Section/5-Conclusion}

\bibliographystyle{plainnat}
\bibliography{references}

\appendix
\input{Appendix/A-SimWorld_Studio}

\input{Appendix/Additional-Related-Work}

\input{Appendix/B-CaseStudy_1}

\input{Appendix/C-CaseStudy_2}

\input{Appendix/D-CaseStudy_3}

\input{Appendix/E-Examples}

\end{document}

%% file: Section/0-Abstract.tex
\begin{abstract}
\vspace{-2mm}
LLM/VLM-based digital agents have advanced rapidly thanks  to scalable sandboxes for coding, web navigation, and computer use, which provide rich interactive training grounds. In contrast, embodied agents still lack abundant, diverse, and automatically generated 3D environments for interactive learning. Existing embodied simulators rely on manually crafted scenes or procedural templates, while recent LLM-based 3D generation systems mainly produce static scenes rather than deployable environments with verifiable tasks and standard learning interfaces. We introduce \ours{}, an open-source platform built on Unreal Engine 5 for generating evolving embodied learning environments. At its core is \textsc{SimCoder}, a tool/skill-augmented coding agent that writes and executes engine-level code to construct physically grounded 3D worlds from language/image instructions. \textsc{SimCoder} {\it self-evolves} by using verifier feedback (e.g., compilation errors, physics checks, VLM critiques) to revise environments and autonomously add reusable tools and skills to its library. Generated worlds are exported as Gym-style environments for embodied agent learning. \ours{} further enables {\it co-evolution} between environment generation and embodied learning: agent performance feedback guides \textsc{SimCoder} to generate adaptive curricula near the learner's capability frontier, so that environments become increasingly challenging as the embodied agent improves. Three case studies on embodied navigation show that self-evolution improves generation reliability, generated environments substantially improve embodied agent performance that generalizes to unseen benchmarks, and co-evolution yields an 18-point success-rate gain over fixed-environment learning and a 40-point gain over an untrained agent.\footnote{Code is available at \href{https://github.com/SimWorld-AI/SimWorld-Studio}{\texttt{https://github.com/SimWorld-AI/SimWorld-Studio}}.}

\end{abstract}

%% file: Section/1-Introduction.tex
\vspace{-6mm}
\section{Introduction}
\label{sec:intro}
\vspace{-2mm}
Large language and vision models have recently made striking progress as \emph{digital agents}: they can write and debug code, operate graphical user interfaces, navigate the web, and complete multi-step tasks in software environments. A key enabler of this progress is the availability of scalable interactive digital sandboxes, such as code execution environments and operating-system simulators, in which agents can act, receive feedback, and improve through repeated experience \cite{dong2026agentworldscalingrealworldenvironment,guo2025genenvdifficultyalignedcoevolutionllm, zhao2024expel}. By contrast, progress toward similarly capable \emph{embodied agents} remains comparatively limited. Although LLMs and VLMs provide powerful priors for perception, reasoning, and planning in 3D worlds~\citep{driess2023palmeembodiedmultimodallanguage, zitkovich2023rt2}, embodied learning still lacks the kind of abundant, diverse, and automatically generated interactive environments that digital agents increasingly rely on.

A central bottleneck is the difficulty of simulating embodied environments at scale. Training and evaluating embodied agents require not only visually plausible 3D scenes, but also physically grounded worlds in which agents can be deployed, take actions, observe consequences, and receive task feedback. Existing embodied platforms, such as AI2-THOR~\citep{kolve2022ai2thorinteractive3denvironment}, Habitat~\citep{puig2023habitat30cohabitathumans}, CARLA~\citep{dosovitskiy2017carlaopenurbandriving}, ThreeDWorld~\citep{gan2021threedworldplatforminteractivemultimodal}, and iGibson~\citep{li2021igibson}, provide important infrastructure for embodied AI, but they largely depend on manually designed scene collections that are expensive to construct, limited in diversity, and fixed once released. Procedurally generated platforms such as ProcTHOR~\citep{deitke2022procthorlargescaleembodiedai} and Infinigen~\citep{raistrick2023infinite} improve scalability, yet their diversity is still bounded by hand-designed templates or rules. Meanwhile, a growing line of work explores LLM- or coding-agent-based 3D scene generation, either by predicting layouts or by writing executable code against a game engine~\citep{yang2024holodecklanguageguidedgeneration, hu2024scenecraftllmagentsynthesizing, zhang2026code2worlds, xia2026sagescalableagentic3d, kuang2026vulcantoolaugmentedmultiagents, liu2026imaginecitycitygenagentprocedural, pfaff2026scenesmith}. However, these systems primarily generate \emph{static scenes}: their outputs are typically evaluated as visual or geometric artifacts, rather than as deployable interactive environments. 

The distinction between \emph{scene generation} and \emph{environment generation} is crucial. For embodied agent learning, a generated world must be more than a visually plausible arrangement of objects: it must be an interactive system in which agents can perceive, act, and receive feedback. Such environments should expose observations and actions through a standard interface, define verifiable tasks, provide reward signals, and support training and evaluation without manual integration.
Moreover, the environment generator itself should not remain fixed. As an embodied agent improves, the simulator should be able to generate more diverse, complex, and challenging environments informed by the agent's current capabilities. Such a closed loop would turn environment generation from a one-shot content-creation problem into an adaptive curriculum mechanism, where the worlds generated for training evolve together with the agents learning inside them.

We introduce \ours{}, an open-source platform built on Unreal Engine~5 for automatic generation of evolving interactive embodied learning environments (Figure~\ref{fig:teaser}). At its core is \textsc{SimCoder}, a tool-augmented coding agent that creates realistic, physically grounded UE5 environments from natural-language instructions, image guidance, and editing requests. 
Rather than merely placing static assets, \textsc{SimCoder} writes and executes engine-level code to construct diverse environments, ranging from simple street corners to full city districts. It uses rich verifier feedback, including compilation errors, collision reports, physics checks, and VLM critiques, to revise generated environments for improved validity (Figure~\ref{fig:platform}). Over time, \textsc{SimCoder} can also autonomously author new tools and reusable skills, add them to its own library for reuse in future generations, thereby improving reliability and scalability.
Similar to previous tool-making LLMs \citep{cai2024largelanguagemodelstool, wang2023voyageropenendedembodiedagent}, 
this mechanism closes a {\it self-evolution} loop for the coding agent without manual intervention.

Every environment generated by \ours{} can be seamlessly exported as a standardized Gymnasium-style embodied environment, with \texttt{reset()}, \texttt{step()}, and task-dependent observation spaces, action spaces, and reward signals. In this work, we use navigation as a representative case study: tasks are automatically derived from the generated scene structure, including traversable regions, obstacles, goals, and spatial relations. This allows LLM-based or other embodied agents to be deployed directly in generated worlds and trained on verifiable downstream tasks. Crucially, \ours{} also supports a {\it co-evolution} loop between the coding agent and the embodied agent. Performance signals from the embodied learner, such as task success, failure modes, and exploration coverage, are fed back to \textsc{SimCoder}, steering future generation toward environments near the frontier of the learner's current ability. In this way, \ours{} aims to provide not only a scalable source of embodied training environments, but also an adaptive platform in which environment generation and embodied agent learning improve together. Compared with preliminary attempts~\citep{zala2024envgengeneratingadaptingenvironments}, which use LLMs to adapt predefined simple game environments for small RL agents, \ours{} provides a flexible and realistic platform for environment-agent co-adaptation.

Across three case studies (Figure~\ref{fig:case_studies}), we show that 
(i) \textsc{SimCoder} reliably generates physically valid and prompt-aligned environments, with structured tools, verification, and self-evolution each contributing measurably to quality; 
(ii) embodied agents trained in the generated environments achieve substantial improvements that transfer to unseen navigation benchmarks, with environment diversity directly driving generalization; and 
(iii) closing the co-evolution loop between \textsc{SimCoder} and the embodied agent via an adaptive curriculum yields an 18-point Success Rate gain over fixed-environment training and a 40-point gain over an untrained agent, showing that generated environments become more effective for embodied learning when shaped by agent feedback.\footnote{
Additional UI views, running cases, prompts, and generated tools/skills/examples are provided in Appendices~\ref{app:platform}, \ref{app:prompts}, and~\ref{app:examples}. 
}
We will open source the platform and all experiments upon acceptance.

%% file: Section/3-SimWorld_Studio.tex
\section{\ours{}}
\vspace{-3mm}
\label{sec:platform}

\ours{} 
is built on the open-source Unreal Engine 5 based \textsc{SimWorld} library \citep{ye2025simworld} by inheriting its assets, runtime, and Python wrapper on the UE5 backend, enabling highly realistic, physically grounded environments.
\ours{} makes two main methodological contributions: \textbf{(1) Automatic environment generation} (\S\ref{sec:env_generation}): a coding agent that synthesizes executable 3D scenes, evolves its own skill and tool library from verifier feedback, and exports each scene as a Gymnasium-compatible embodied environment.
\textbf{(2) Co-evolution as an adaptive curriculum mechanism} (\S\ref{sec:coevolution}): embodied agent performance is fed back into environment generation, so new environments target the agent’s current weaknesses and remain near the boundary of its capabilities. See our main UI page in Fig~\ref{fig:main_ui}.


\begin{figure}[t]
    \centering
    \includegraphics[width=1.0\textwidth]{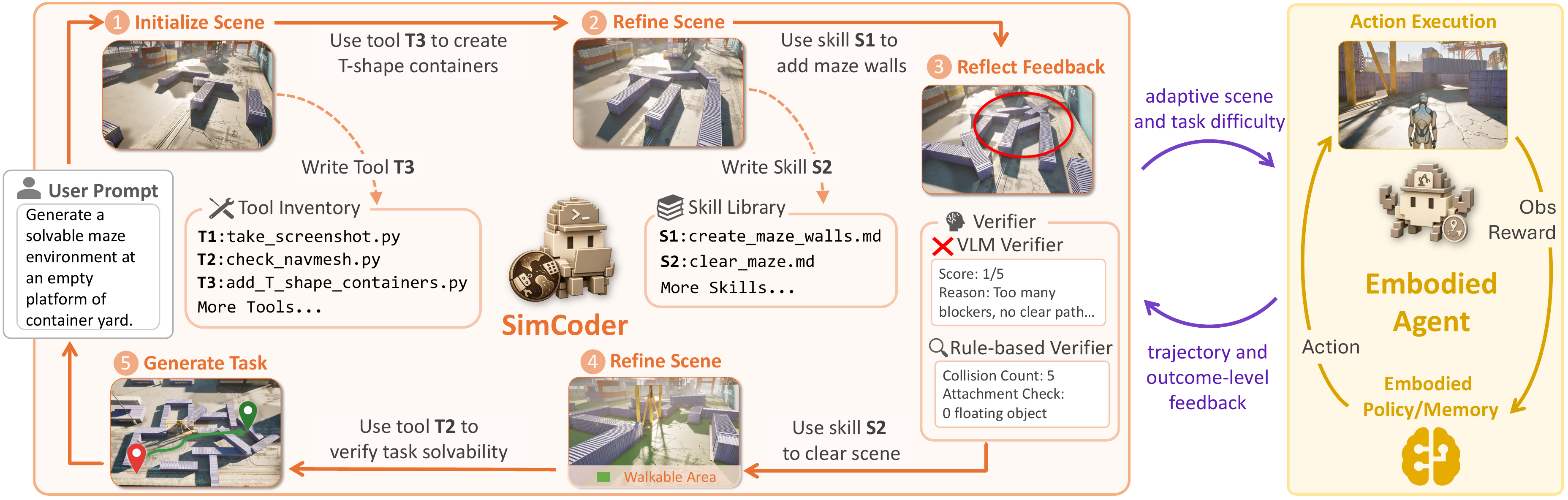}
    \vspace{-10pt}
    \caption{\textsc{SimCoder} turns a user prompt into an interactive environment through an automatic self-evolving loop: it writes tools, creates reusable skills, reuses them across iterations, and refines the scene with verifier feedback. NavMesh-based tools are used to generate solvable navigation tasks. 
    \textsc{SimCoder} furthers uses embodied-agent feedback to autonomously adapt environment difficulty and co-evolve with the embodied agent (purple loop).} 
    \vspace{-3mm}
    \label{fig:platform}
\end{figure}

\vspace{-3mm}
\subsection{\textsc{SimCoder}: Coding Agent for Automatic Environment Generation}
\vspace{-2mm}
\label{sec:env_generation}

As shown in Figure~\ref{fig:teaser}(Left), \ours{} comprises three components:  \textbf{\textsc{SimCoder}}, an LLM coding agent that drives generation; \textit{tool and skill libraries}, including an inventory of Python functions as tools and a library of skills which are reusable procedures, exposed through a Model Context Protocol~\citep{anthropic2024mcp} (MCP) bridge;  and \textit{verifiers} that return verification signals (rule- and VLM-based) to guide scene construction and revision. 
As illustrated in Figure~\ref{fig:platform} with a maze-generation task, generation flows in a loop: given a user prompt (text, image, or edit instruction), \textsc{SimCoder} issues tool calls or skill retrievals through MCP; the backend executes them and returns a state update or a verifier signal, which \textsc{SimCoder} consumes as the next observation and either continues building the scene, revises in place, or, when a fix proves broadly useful, writes a new tool or skill back into its library so future generations can reuse it. 
Once the scene passes verification, \textsc{SimCoder} derives a task from it and exports it as a Gymnasium environment, allowing embodied agents to interact with it.



\newtcolorbox{interfacecasebox}[1]{
    enhanced,
    breakable,
    width=\textwidth,
    colback=black!3,
    colframe=black!18,
    boxrule=0.6pt,
    arc=1.5mm,
    left=1mm,
    right=1mm,
    top=1.2mm,
    bottom=1.2mm,
    before skip=5pt,
    after skip=6pt,
    title=#1,
    coltitle=black,
    colbacktitle=black!8,
    fonttitle=\small\bfseries,
    attach boxed title to top left={xshift=1.5mm,yshift=-1.2mm},
    boxed title style={
        colback=black!8,
        colframe=black!18,
        boxrule=0.4pt,
        arc=1.2mm,
        left=1.2mm,
        right=1.2mm,
        top=0.6mm,
        bottom=0.6mm
    }
}

\Needspace{0.35\textheight}
\begin{interfacecasebox}{\textsc{SimWorld Studio} User Interface}
    \vspace{-0.5mm}
    \centering
    \includegraphics[width=\linewidth]{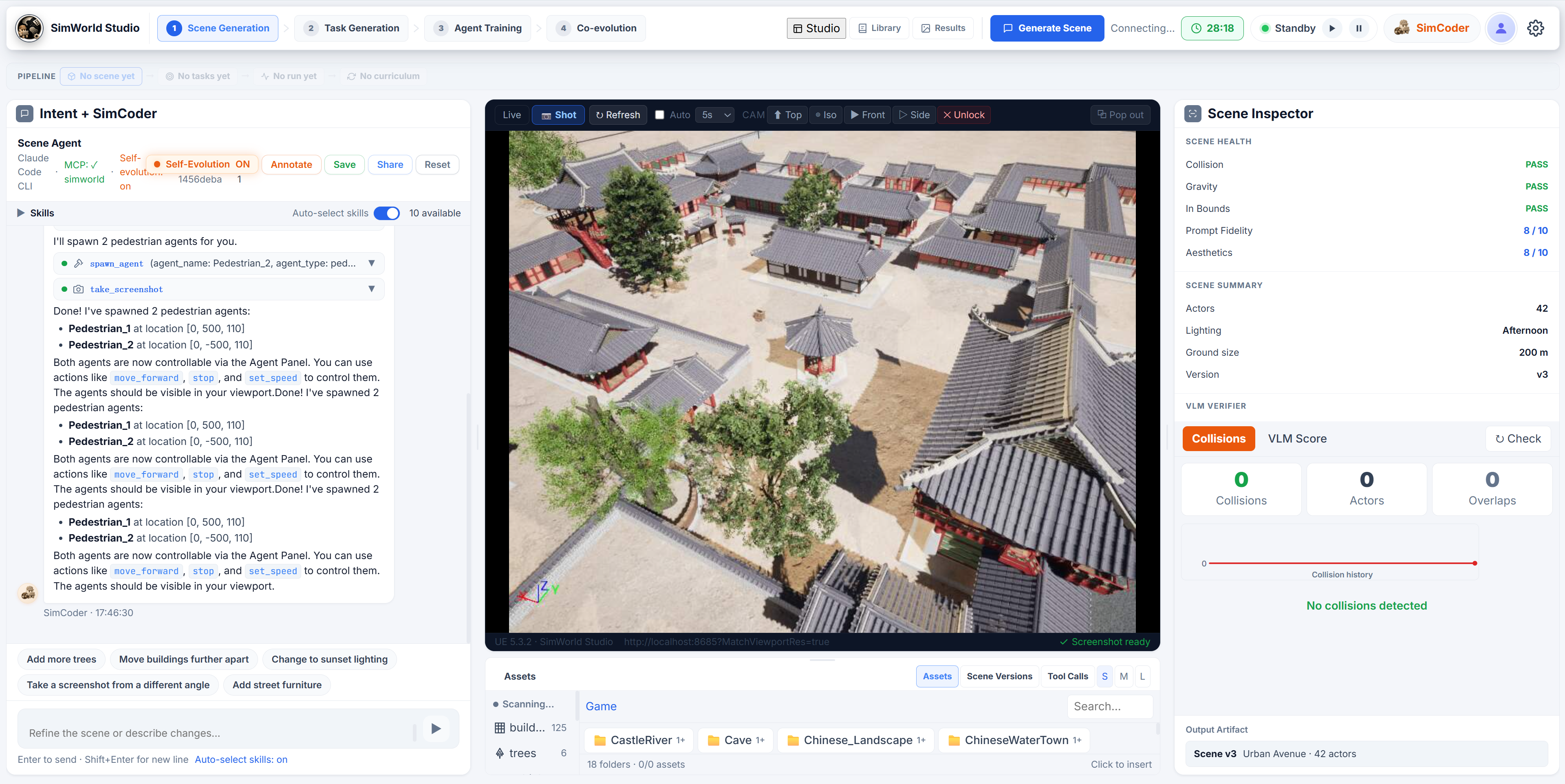}

    \vspace{-1mm}
    \captionof{figure}{The workspace supports the full environment-generation workflow in a single view. The top pipeline bar tracks progress across \textit{scene generation}, \textit{task generation}, \textit{agent training}, and \textit{co-evolution}. In this scene generation page, the left panel combines natural-language interaction with \textsc{SimCoder}, executable scene and agent-control skills, and quick editing actions. The central viewport streams the live Unreal Engine scene, while the bottom asset browser exposes available scene assets and saved scene versions. The right \emph{Scene Inspector} shows scene health, scene summary, and VLM-based verification.}
    \label{fig:main_ui}
\end{interfacecasebox}

\vspace{-2mm}
\paragraph{MCP Tools.}\label{sec:coding_agent}
Tools are Python function calls that \textsc{SimCoder} invokes through the MCP bridge to act on the UE backend. The inventory has two parts. \emph{Primitive tools} are the fixed, predefined set of operations needed to author a scene end-to-end (e.g., actor management, environment and asset management, and scene evaluation). \emph{Extensible tools} cover everything outside this fixed set: a Python escape hatch runs arbitrary Unreal Engine Python for one-off operations, and any pattern that proves useful across runs is promoted via self-evolution into a named wrapper that the bridge registers as a first-class MCP tool, indistinguishable from the primitives at call time. Step~1 of Figure~\ref{fig:platform} shows one such wrapper (\texttt{add\_T\_shape\_containers.py}) being invoked from the Tool Inventory. The full primitive inventory is in Appendix~\ref{app:mcp_tools}.
\vspace{-3mm}
\paragraph{Skill Library.}
Skills sit one layer above tools. Each skill is a Markdown document that records how to use a tool (or a sequence of tools) to accomplish a particular composition goal; \textsc{SimCoder} retrieves applicable skills at the start of each episode and issues the underlying tool calls itself, so skills tell it how to compose tools rather than bypass them. As with tools, the library has two parts: a small set of \emph{primitive skills} ships with the platform (covering common composition goals such as building placement, city layout, and screenshot capture for the VLM judge), and \emph{extensible skills} accumulate over time through self-evolution. Step~2 of Figure~\ref{fig:platform} shows \textsc{SimCoder} retrieving an evolved skill (\texttt{create\_maze\_walls.md}) to add walls to the partially-built maze.
\vspace{-3mm}
\paragraph{Verification Loop.}\label{sec:verifier}
\ours{} verifies generated scenes through two complementary verifiers (Step~3 of Figure~\ref{fig:platform}). A \emph{rule-based verifier} computes physical and geometric metrics (e.g., collisions, vertical support, in-bounds placement) from the scene graph and is invoked on every actor-modifying tool call. A \emph{VLM-based verifier} captures multi-view screenshots and asks a vision-language model to score semantic alignment against the prompt, returning structured feedback after each block of construction. Verifier responses re-enter the trajectory as the next observation, and \textsc{SimCoder} revises in place. In the maze episode of Figure~\ref{fig:platform}, for example, the rule-based verifier reports a collision count of~5 and the VLM verifier scores the scene 1/5 (\textit{``too many blockers, no clear path…''}); \textsc{SimCoder} then retrieves the \texttt{clear\_maze.md} skill and removes the redundant containers before continuing. Full metric definitions are in Appendix~\ref{app:metrics}.
\vspace{-3mm}
\paragraph{Self-evolution.}\label{sec:self_evolution}
Self-evolution turns one-off fixes into permanent capabilities. When a verifier failure recurs across attempts, \textsc{SimCoder} restates the failure at the level of a \emph{class} of cases and authors a new tool or skill that addresses the class rather than the specific instance, writing it to the registry so all subsequent runs can retrieve it~\citep{cai2024largelanguagemodelstool, qian2023creatortoollibcreationllmagent}. Step~4 of Figure~\ref{fig:platform} illustrates one such update: after the maze fails verification, \textsc{SimCoder} writes a new skill (\texttt{clear\_maze.md}) that generalizes the corrective procedure (i.e., removing redundant blockers from any container layout) and the skill is then available for all future episodes. Representative authored entries are in Appendix~\ref{app:skill_library}.



\vspace{-2mm}
\paragraph{Task Generation.}\label{sec:task_gen}\label{sec:embodied_agent}
\textsc{SimCoder} also generates a task on top of a generated scene, using the same tool-call interface to query scene structure (e.g., NavMesh for traversable regions). Step~5 of Figure~\ref{fig:platform} shows the maze scene compiled into a navigation task with a sampled start--goal pair on the walkable area. We instantiate two canonical navigation families as a representative case: \emph{point navigation}~\citep{anderson2018evaluation} (goal = coordinate) and \emph{object navigation}~\citep{batra2020objectnav} (goal = semantic target). Task solvability is guaranteed by NavMesh connectivity, and verifiability follows from the same scene-query tools: during execution, we directly query the agent pose and target location and check success based on distance to the target.
\vspace{-3mm}
\paragraph{Gymnasium Compilation.}
A generated environment then exports as a standard Gymnasium environment, with \texttt{env.reset()} and \texttt{env.step(action)} returning RGB-D observations, agent pose, and reward (top of Figure~\ref{fig:teaser}(Left)). Because the contract is the standard one, any off-the-shelf RL algorithm (e.g., PPO~\citep{schulman2017ppo}) or training-free LLM policy (e.g., ReAct~\citep{yao2022react}) plugs in without modification, making each generated scene a first-class training substrate for embodied agent learning. 
\vspace{-3mm}
\subsection{Co-Evolution: An Adaptive Curriculum Mechanism}
\label{sec:coevolution}
\vspace{-2mm}
So far the generator runs open-loop: it produces environments without knowing how the embodied agent fares in them. Co-evolution closes this loop and turns environment generation from a one-shot content-creation problem into an adaptive curriculum mechanism, where the scenes generated for training evolve together with the agents learning inside them. One round alternates two updates: the embodied agent trains on a batch of \textsc{SimCoder}-generated environments, and \textsc{SimCoder} then updates based on the resulting performance before producing the next batch. The two agents update individually, through different mechanisms.
\vspace{-3mm}
\paragraph{Embodied Agent Evolving.}
From the embodied agent's perspective, co-evolution differs from fixed-environment training only in that the scene distribution drifts between rounds; the agent's update rule is unchanged. \ours{} reuses the Gym interface of \S\ref{sec:task_gen} without modification, so an RL policy (e.g., PPO~\citep{schulman2017ppo}) updates via standard policy gradients on the reward returned by \texttt{step()}, while an LLM-based policy updates through in-context mechanisms such as incremental rule accumulation or reflection-style memory~\citep{wang2023voyageropenendedembodiedagent, shinn2023reflexion}.
\vspace{-3mm}
\paragraph{Coding Agent Evolving.}
\textsc{SimCoder}'s update is in-context: between rounds the embodied agent's performance is fed back as context for the next generation episode, and \textsc{SimCoder} reweights its skill retrievals and tool invocations to raise difficulty where success rates plateau, lower it where the agent stalls, and oversample structural features the agent has not yet mastered. The underlying LLM weights are not modified. The performance signal is read through three feedback channels at increasing abstraction: \emph{scene-level} feedback reports physical validity and prompt alignment of scenes; \emph{outcome-level} feedback provides task success and return statistics for difficulty-matching objectives~\citep{wang2019poet, dennis2020emergent}; and \emph{trajectory-level} feedback exposes the agent's per-episode experience for reflection-based updates to \textsc{SimCoder}'s generation principles~\citep{wang2023voyageropenendedembodiedagent}. A specific co-evolution \emph{recipe} selects a subset of these channels and pairs it with the embodied agent's learning rule.

Section~\ref{sec:case_study_3} instantiates this recipe for navigation tasks, using outcome-level agent outcomes to adapt \textsc{SimCoder}'s difficulty schedule while the agent improves through incremental rule accumulation. The resulting adaptive curriculum outperforms fixed-environment training.

%% file: Section/4-Experiment.tex
\begin{figure}[t]
    \centering
    \includegraphics[width=1.0\linewidth]{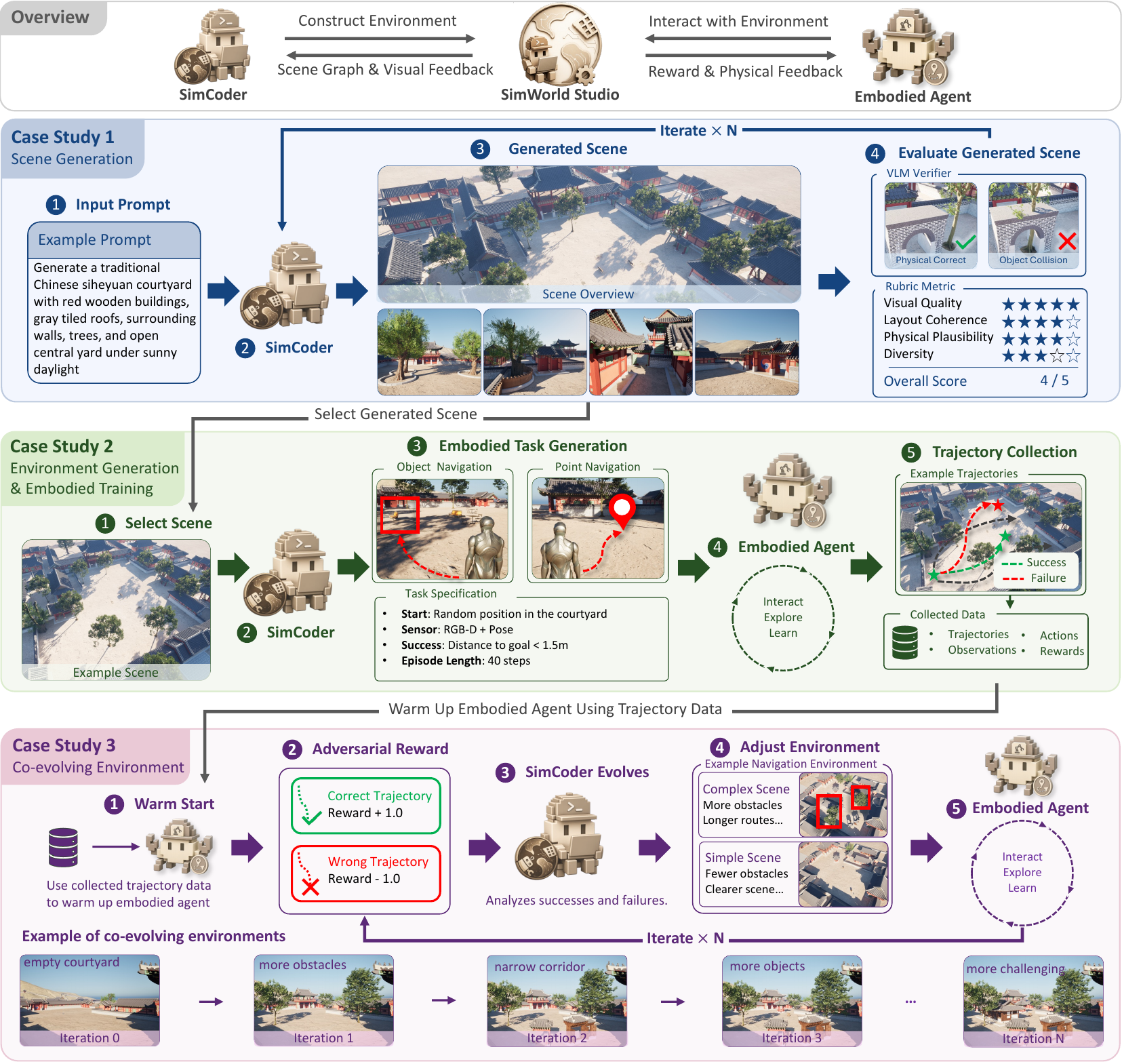}
    \vspace{-15pt}
    \caption{
        \textbf{Three case studies evaluating \ours{}.}
        Case~1 evaluates \textsc{SimCoder}'s scene generation quality across settings and LLM backbones.
        Case~2 trains embodied navigation agents in generated environments.
        Case~3 studies co-evolution where \textsc{SimCoder} and the embodied agent iteratively improve each other.
        }
    \label{fig:case_studies}
    \vspace{-5mm}
\end{figure}

\section{Experiments and Analysis}
\label{sec:experiments}
\vspace{-3mm}

We analyze \ours{} through three case studies of increasing scope (Figure~\ref{fig:case_studies}): environment generation quality (\S\ref{sec:case_study_1}), embodied agent learning in generated environments (\S\ref{sec:case_study_2}), and co-evolution between the environment generation and the embodied agent (\S\ref{sec:case_study_3}).

\vspace{-3mm}
\subsection{Case Study 1: Can \textsc{SimCoder} generate valid and diverse environments?}
\label{sec:case_study_1}
\vspace{-2mm}
This case study evaluates whether \textsc{SimCoder} can generate diverse, physically plausible 3D environments from natural language prompts, reference images, and editing instructions.
As illustrated in Figure~\ref{fig:case_studies} (case study 1 left), \textsc{SimCoder} receives a text prompt (e.g., ``build a residential neighborhood with parallel streets and a park''), invokes MCP tools to spawn and arrange assets in the UE5 environment, and iteratively refines the scene through screenshot-based verification (\S\ref{sec:verifier}).

\textbf{Settings.}
We evaluate across three settings of increasing complexity:
\textbf{(S1)} \emph{Text-to-Scene}: generate a scene from a natural language prompt alone;
\textbf{(S2)} \emph{Image+Text-to-Scene}: generate with an additional reference image (hand-drawn sketch or aerial photo);
\textbf{(S3)} \emph{Scene Editing}: modify an existing scene by adding, removing, or rearranging objects without rebuilding from scratch.
Each setting is tested at three difficulty levels (easy, medium, hard), yielding 9 evaluation scenes total.
We use the two-axis evaluation from \S\ref{sec:verifier}: rule-based metrics for physical validity (e.g., collision-free placement, gravity consistency, in-bounds placement) and VLM-as-judge metrics for semantic alignment (e.g., prompt fidelity, spatial fidelity, layout aesthetics); full definitions are in Appendix~\ref{app:metrics}.

\textbf{Base Models.}
We benchmark four LLM backbones, including Claude Opus 4.6~\citep{anthropic2026opus46}, Claude Sonnet 4.6~\citep{anthropic2026sonnet46}, and Qwen3.5-27B/9B~\citep{qwen2026qwen35}, all through the Claude Code agent framework~\citep{anthropic2026claudecode} with the same MCP tool interface, verification loop, and skill library (\S\ref{sec:env_generation}).
All agents differ only in the underlying LLM, isolating the contribution of model capability from platform infrastructure.

\definecolor{metphys}{HTML}{1565C0}
\definecolor{metquant}{HTML}{2E7D32}
\definecolor{metsem}{HTML}{E65100}
\definecolor{metaes}{HTML}{AD1457}

\vspace{-3mm}
\subsubsection{Results}
\vspace{-2mm}

Table~\ref{tab:main_results} reports performance averaged across difficulty levels; full breakdowns are in Appendix~\ref{app:full_results}.

\begin{table}[t]
\centering
\caption{\textbf{Scene generation quality (averaged across difficulty levels).} All metrics $\in [0,1]$ ($\uparrow$). Bold = best per column. Metric colors: \textcolor{metquant}{quantity}, \textcolor{metphys}{physical validity}, \textcolor{metsem}{semantic}, \textcolor{metaes}{aesthetic}. }
\label{tab:main_results}
\vspace{2pt}
\scriptsize
\setlength{\tabcolsep}{3.5pt}
\renewcommand{\arraystretch}{1.12}
\resizebox{\textwidth}{!}{%
\begin{tabular}{@{}l cccc cc c cccc cc c cccc cc c@{}}
\toprule
& \multicolumn{7}{c}{\textbf{S1: Text-to-Scene}} & \multicolumn{7}{c}{\textbf{S2: Image+Text-to-Scene}} & \multicolumn{7}{c}{\textbf{S3: Scene Editing}} \\
\cmidrule(lr){2-8} \cmidrule(lr){9-15} \cmidrule(lr){16-22}
& \multicolumn{4}{c}{\textit{Rule-Based}} & \multicolumn{2}{c}{\textit{VLM}} &
  & \multicolumn{4}{c}{\textit{Rule-Based}} & \multicolumn{2}{c}{\textit{VLM}} &
  & \multicolumn{4}{c}{\textit{Rule-Based}} & \multicolumn{2}{c}{\textit{VLM}} & \\
\cmidrule(lr){2-5} \cmidrule(lr){6-7} \cmidrule(lr){9-12} \cmidrule(lr){13-14} \cmidrule(lr){16-19} \cmidrule(lr){20-21}
\textbf{LLM}
  & \rotatebox{70}{\textcolor{metquant}{Count}} & \rotatebox{70}{\textcolor{metquant}{Diversity}} & \rotatebox{70}{\textcolor{metphys}{No Collision}} & \rotatebox{70}{\textcolor{metphys}{Gravity}}
  & \rotatebox{70}{\textcolor{metsem}{Fidelity}} & \rotatebox{70}{\textcolor{metaes}{Aesthetics}} & \rotatebox{70}{\textbf{Avg}}
  & \rotatebox{70}{\textcolor{metquant}{Count}} & \rotatebox{70}{\textcolor{metphys}{No Collision}} & \rotatebox{70}{\textcolor{metphys}{Gravity}} & \rotatebox{70}{\textcolor{metphys}{In-Bounds}}
  & \rotatebox{70}{\textcolor{metsem}{Fidelity}} & \rotatebox{70}{\textcolor{metsem}{Style}} & \rotatebox{70}{\textbf{Avg}}
  & \rotatebox{70}{\textcolor{metphys}{Preserve}} & \rotatebox{70}{\textcolor{metquant}{Edit Count}} & \rotatebox{70}{\textcolor{metphys}{No Collision}} & \rotatebox{70}{\textcolor{metsem}{Coherence}}
  & \rotatebox{70}{\textcolor{metsem}{Edit Compl.}} & \rotatebox{70}{\textcolor{metaes}{Layout}} & \rotatebox{70}{\textbf{Avg}} \\
\midrule
Qwen3.5-9B
  & .00 & .50 & .37 & .58 & .20 & .13 & .36
  & .60 & .33 & .53 & 1.0 & .13 & .13 & .45
  & 1.0 & .00 & .37 & .23 & .23 & .20 & .34 \\
Qwen3.5-27B
  & .17 & .46 & .99 & 1.0 & .43 & .37 & .59
  & .73 & .75 & .83 & 1.0 & .38 & .30 & .67
  & 1.0 & .00 & .99 & .47 & .13 & .50 & .52 \\
Sonnet 4.6
  & .33 & \textbf{.80} & \textbf{1.0} & 1.0 & .53 & .37 & .70
  & .80 & .81 & .87 & 1.0 & .50 & .37 & .73
  & 1.0 & 1.0 & .98 & .40 & \textbf{.67} & .40 & .74 \\
Opus 4.6
  & \textbf{.75} & .72 & .99 & 1.0 & \textbf{.63} & \textbf{.47} & \textbf{.77}
  & \textbf{.83} & \textbf{.86} & \textbf{.92} & 1.0 & \textbf{.62} & \textbf{.47} & \textbf{.79}
  & 1.0 & 1.0 & .98 & \textbf{.50} & .50 & \textbf{.53} & \textbf{.75} \\
\bottomrule
\end{tabular}
}
\vspace{-4mm}
\end{table}



\textbf{\textsc{SimCoder} with different coding models generates physically valid environments; quality scales with model capability}.
Near-perfect physical validity holds across all settings, Opus~4.6 and Sonnet~4.6 maintain collision-free rates $\geq$0.98 regardless of input modality or difficulty (see Appendix~\ref{app:full_results}).
Semantic quality scales with model size: Opus~4.6 leads across all three settings (S1:~0.77, S2:~0.79, S3:~0.75), and image guidance consistently boosts smaller models (Qwen3.5-27B: S1~0.59$\to$S2~0.67) by anchoring spatial layout.
\vspace{-1mm}
\begin{figure}[h]
        \centering
        \vspace{-0mm}
    \includegraphics[width=0.75\linewidth]{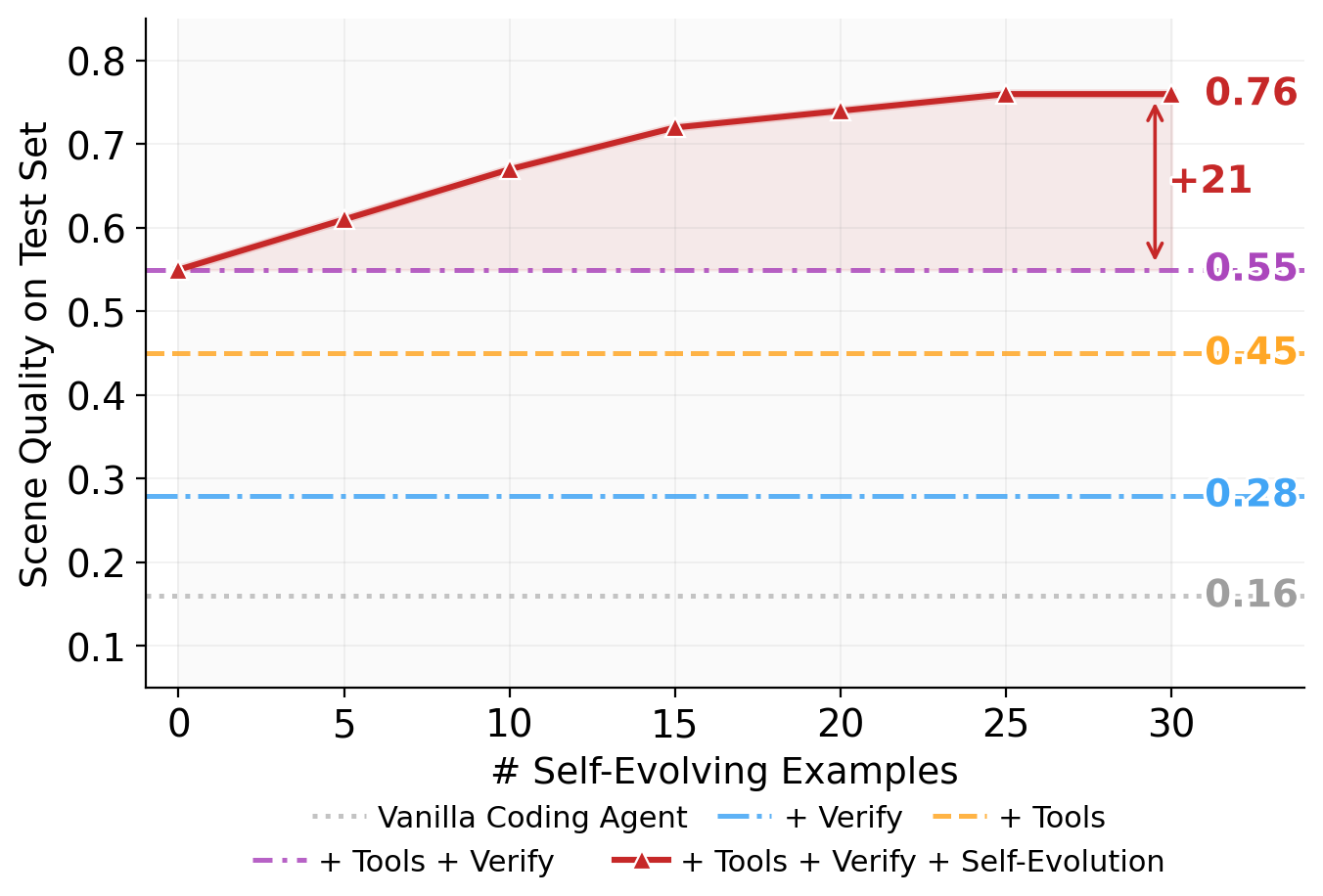}
    \vspace{-2mm}
        \caption{Ablation study results for \textsc{SimCoder} in Case Study 1.}
        \label{fig:ablation}
    \vspace{-0mm}
\end{figure}
\vspace{-4mm}
\paragraph{Ablation studies on key platform components.}

Figure~\ref{fig:ablation} ablates three platform components beyond the vanilla coding agent: MCP tools, verification loop, and self-evolution. The evaluation is conducted on a held-out test set of 9~scenes across S1/S2/S3.
First, we observe that the \textit{vanilla coding agent} fails to construct reliable environments (scoring 0.16). Then, adding \textit{customized MCP tools} raises quality to 0.45 ($+$0.29), providing the structured action space needed for reliable asset interaction. Moreover, adding the \textit{verification loop} improves quality by $+$0.10, as iterative screenshot-based correction catches spatial errors that single-pass generation misses. We find that \textit{self-evolution} can break the plateau shared by all static configurations, further raising a $+$0.21 quality improvement by accumulating reusable placement strategies across generations. Together, these results show that structured tool access is a hard prerequisite, while self-evolution provides the largest quality gain by converting experience into reusable knowledge. Full ablation details are in Appendix~\ref{app:cs1_abl_res}.

\vspace{-0mm}
\subsubsection{Qualitative Example: Text-to-Scene Generation}
\vspace{-1mm}

Figure~\ref{fig:qualitative_medieval_outputs} presents a concrete example of how \textsc{SimCoder} transforms a natural-language prompt into a complete UE5 scene. Figure~\ref{fig:qualitative_medieval_tool_skill_code} further illustrates the corresponding tool calls, skills, and code used to construct the scene. Additional qualitative examples are in Appendix~\ref{app:examples}.

\definecolor{metquant}{HTML}{B66A00}
\definecolor{metphys}{HTML}{1F7A5A}
\definecolor{metsem}{HTML}{2F63B7}
\definecolor{metaes}{HTML}{8A4FBF}

\definecolor{softCream}{HTML}{FFF9F0}
\definecolor{softLine}{HTML}{F1D7B8}
\definecolor{softBrown}{HTML}{5E4026}
\definecolor{softMuted}{HTML}{8A6B4D}

\newcommand{\MetricHeader}[2]{%
    \textcolor{#1}{\scriptsize\bfseries #2}%
}

\newcommand{\ScoreTable}[7]{%
\begin{tabular}{@{}lccccccc@{}}
\toprule
& \MetricHeader{metquant}{Count}
& \MetricHeader{metquant}{Div.}
& \MetricHeader{metphys}{No Col.}
& \MetricHeader{metphys}{Grav.}
& \MetricHeader{metsem}{Fid.}
& \MetricHeader{metaes}{Aes.}
& \textbf{Avg} \\
\midrule
& #1 & #2 & #3 & #4 & #5 & #6 & \textbf{#7} \\
\bottomrule
\end{tabular}
}

\newcommand{\SceneRow}[9]{%
\begin{tcolorbox}[
    enhanced,
    colback=white,
    colframe=softLine,
    boxrule=0.55pt,
    arc=2.2mm,
    left=2mm,
    right=2mm,
    top=2mm,
    bottom=2mm,
    drop shadow={orange!8!white},
    before skip=2mm,
    after skip=2mm
]
\begin{tabularx}{\linewidth}{@{}>{\centering\arraybackslash}X@{\hspace{3mm}}p{0.36\linewidth}@{}}
\begin{minipage}[c][0.235\textheight][c]{\linewidth}
    \centering
    {\sffamily\bfseries\small\textcolor{softBrown}{#1}}\par\vspace{1mm}
    \includegraphics[width=\linewidth,height=0.205\textheight,keepaspectratio]{#2}
\end{minipage}
&
\begin{minipage}[c][0.235\textheight][c]{\linewidth}
    \centering
    {\sffamily\scriptsize\textcolor{softMuted}{S1: Text-to-Scene metrics}}\par\vspace{1.5mm}
    \sffamily\scriptsize
    \setlength{\tabcolsep}{2.6pt}
    \renewcommand{\arraystretch}{1.12}
    \ScoreTable{#3}{#4}{#5}{#6}{#7}{#8}{#9}
\end{minipage}
\end{tabularx}
\end{tcolorbox}
}

\definecolor{toolblue}{HTML}{E8F0FE}
\definecolor{toolborder}{HTML}{4285F4}
\definecolor{skillgreen}{HTML}{E6F4EA}
\definecolor{skillborder}{HTML}{34A853}
\definecolor{codegray}{HTML}{F5F5F5}
\definecolor{codeborder}{HTML}{5F6368}

\newtcbox{\scorebadge}[1][]{
    on line,
    enhanced,
    colback=white,
    colframe=orange!18!white,
    boxrule=0.35pt,
    arc=1mm,
    left=0.7mm,
    right=0.7mm,
    top=0.35mm,
    bottom=0.35mm,
    boxsep=0pt,
    nobeforeafter,
    #1
}

\newcommand{\CompactScores}[7]{%
\centering
{\sffamily\tiny
\scorebadge{\textcolor{metquant}{\bfseries Cnt.}~#1}\hspace{0.7mm}
\scorebadge{\textcolor{metquant}{\bfseries Div.}~#2}\hspace{0.7mm}
\scorebadge{\textcolor{metphys}{\bfseries NoCol.}~#3}\hspace{0.7mm}
\scorebadge{\textcolor{metphys}{\bfseries Grav.}~#4}\hspace{0.7mm}
\scorebadge{\textcolor{metsem}{\bfseries Fid.}~#5}\hspace{0.7mm}
\scorebadge{\textcolor{metaes}{\bfseries Aes.}~#6}\hspace{0.7mm}
\scorebadge[colback=black!3,colframe=black!14]{\textbf{Avg.}~#7}
}
}



\definecolor{metquant}{HTML}{B66A00}
\definecolor{metphys}{HTML}{2E7D5A}
\definecolor{metsem}{HTML}{355FB8}
\definecolor{metaes}{HTML}{8D56C6}

\definecolor{softLine}{HTML}{EFD7BC}
\definecolor{softBrown}{HTML}{6E4A2D}
\definecolor{softMuted}{HTML}{8E6D50}



\definecolor{metquant}{HTML}{A76500}
\definecolor{metphys}{HTML}{2E7D5A}
\definecolor{metsem}{HTML}{355FB8}
\definecolor{metaes}{HTML}{8D56C6}

\definecolor{softLine}{HTML}{EFD7BC}
\definecolor{softBrown}{HTML}{6E4A2D}
\definecolor{softMuted}{HTML}{8E6D50}

\newtcbox{\ScoreChip}[1][]{
    on line,
    enhanced,
    colback=orange!2!white,
    colframe=orange!14!white,
    boxrule=0.3pt,
    arc=0.8mm,
    left=0.55mm,
    right=0.55mm,
    top=0.25mm,
    bottom=0.25mm,
    boxsep=0pt,
    nobeforeafter,
    #1
}

\newcommand{\InlineScores}[7]{%
{\sffamily\fontsize{5.8}{6.0}\selectfont
\ScoreChip{\textcolor{metquant}{\bfseries Count}~#1}
\ScoreChip{\textcolor{metquant}{\bfseries Diversity}~#2}
\ScoreChip{\textcolor{metphys}{\bfseries  No Collision}~#3}
\ScoreChip{\textcolor{metphys}{\bfseries Gravity}~#4}
\ScoreChip{\textcolor{metsem}{\bfseries Fidelity}~#5}
\ScoreChip{\textcolor{metaes}{\bfseries Aesthetics}~#6}
\ScoreChip[colback=black!3,colframe=black!12]{\bfseries Avg.~#7}
}
}

\newcommand{\SceneCard}[9]{%
\begin{tcolorbox}[
    enhanced,
    colback=white,
    colframe=softLine,
    boxrule=0.55pt,
    arc=2.3mm,
    outer arc=2.3mm,
    left=1.6mm,
    right=1.6mm,
    top=1.1mm,
    bottom=1.2mm,
    drop shadow={orange!5!white},
    before skip=0.55mm,
    after skip=0.55mm
]
\begin{tabularx}{\linewidth}{@{}Xr@{}}
{\sffamily\bfseries\textcolor{softBrown}{#1}}
&
\InlineScores{#3}{#4}{#5}{#6}{#7}{#8}{#9}
\end{tabularx}

\vspace{0.5mm}

\centering
\includegraphics[
    width=\linewidth,
    height=0.24\textheight,
    keepaspectratio
]{#2}
\end{tcolorbox}
}


\begin{figure}[p]
\centering
\vspace*{-7mm}

\begin{minipage}[t][\textheight][s]{\textwidth}
\centering

\begin{tcolorbox}[
    enhanced,
    colback=orange!3!white,
    colframe=orange!22!white,
    boxrule=0.5pt,
    arc=2mm,
    outer arc=2mm,
    left=2.2mm,
    right=2.2mm,
    top=1.1mm,
    bottom=1.2mm,
    drop shadow={orange!6!white},
    borderline west={1.3pt}{0pt}{orange!32!white},
    before skip=0pt,
    after skip=0pt,
    nobeforeafter,
    before upper={
        \sffamily
        \noindent
        {\bfseries\color{softBrown}User Prompt}
        \hfill
        {\scriptsize\color{softMuted}Text-to-Scene Setting}
        \par\vspace{0.35mm}
        {\color{orange!22!white}\hrule height 0.3pt}
        \vspace{1.5mm}
    }
]
\sffamily
\fontsize{7.5}{6.9}\selectfont
\setlength{\parindent}{0pt}
\setlength{\parskip}{0pt}
\color{brown!18!black}
Build a believable medieval village street scene, in daytime. A narrow cobblestone lane runs east--west; timbered houses and a tavern flank \textcolor{orange!70!black}{\bfseries both sides shoulder-to-shoulder}. Outside each house there's the household's everyday clutter, a stack of barrels, a wheelbarrow, some grain sacks, and a tree or two grows between certain houses. A well sits at a small square, and one or two houses have a market canopy fused to the front wall like a stall opening onto the street. Villagers cluster around the well and the market. The aesthetic mood, counts, exact placement, and which-house-gets-what are all your creative call. Everything else is yours to decide: lane length, per-house clutter style, number of canopies, villager density, time-of-day mood.
\vspace{-1mm}
\end{tcolorbox}

\vspace{0.5mm}

\SceneCard
    {Claude Opus 4.6}
    {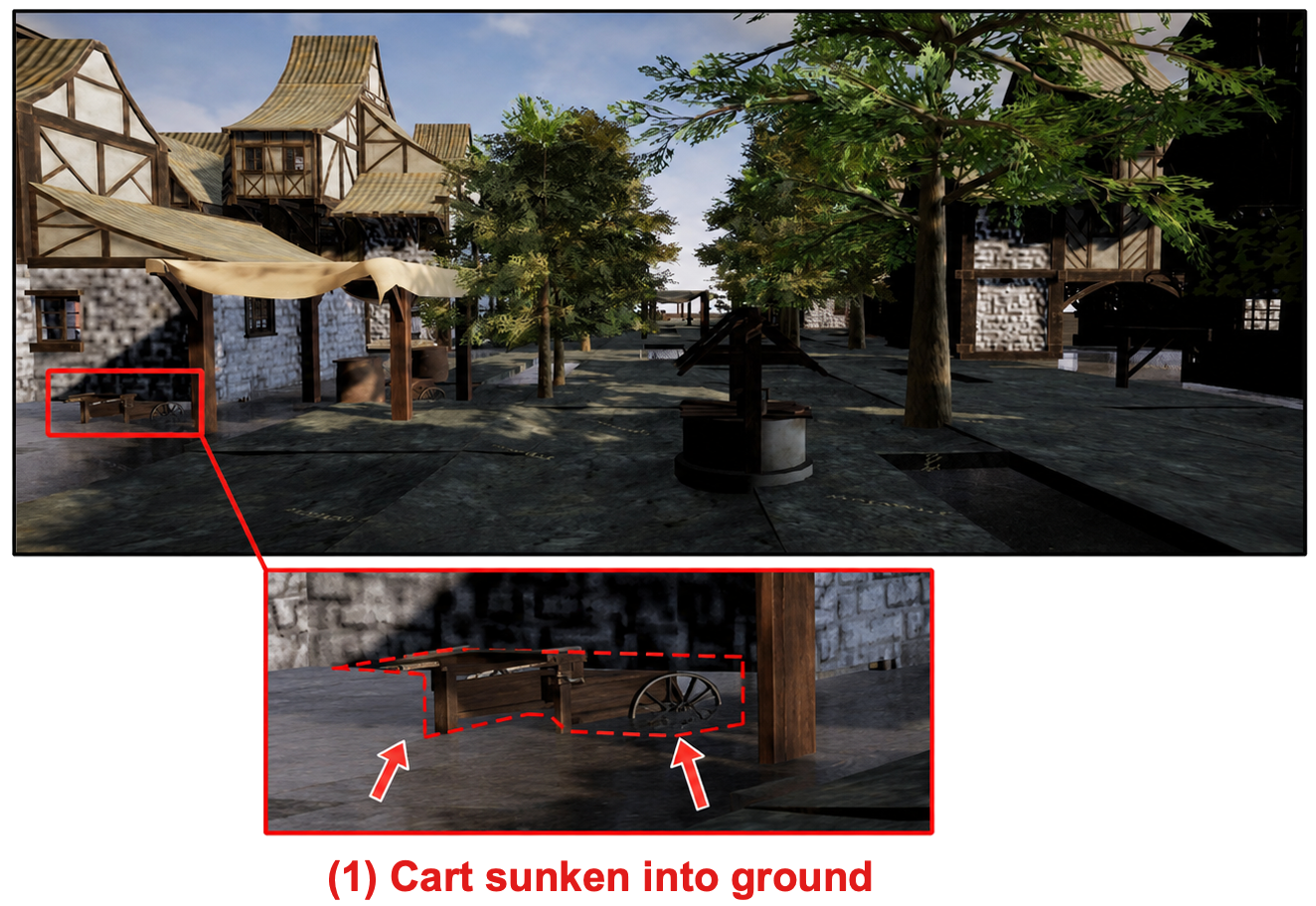}
    {0.72}{0.70}{0.78}{0.74}{0.64}{0.66}{0.71}

\vspace{0.35mm}

\SceneCard
    {Qwen3.5-27B}
    {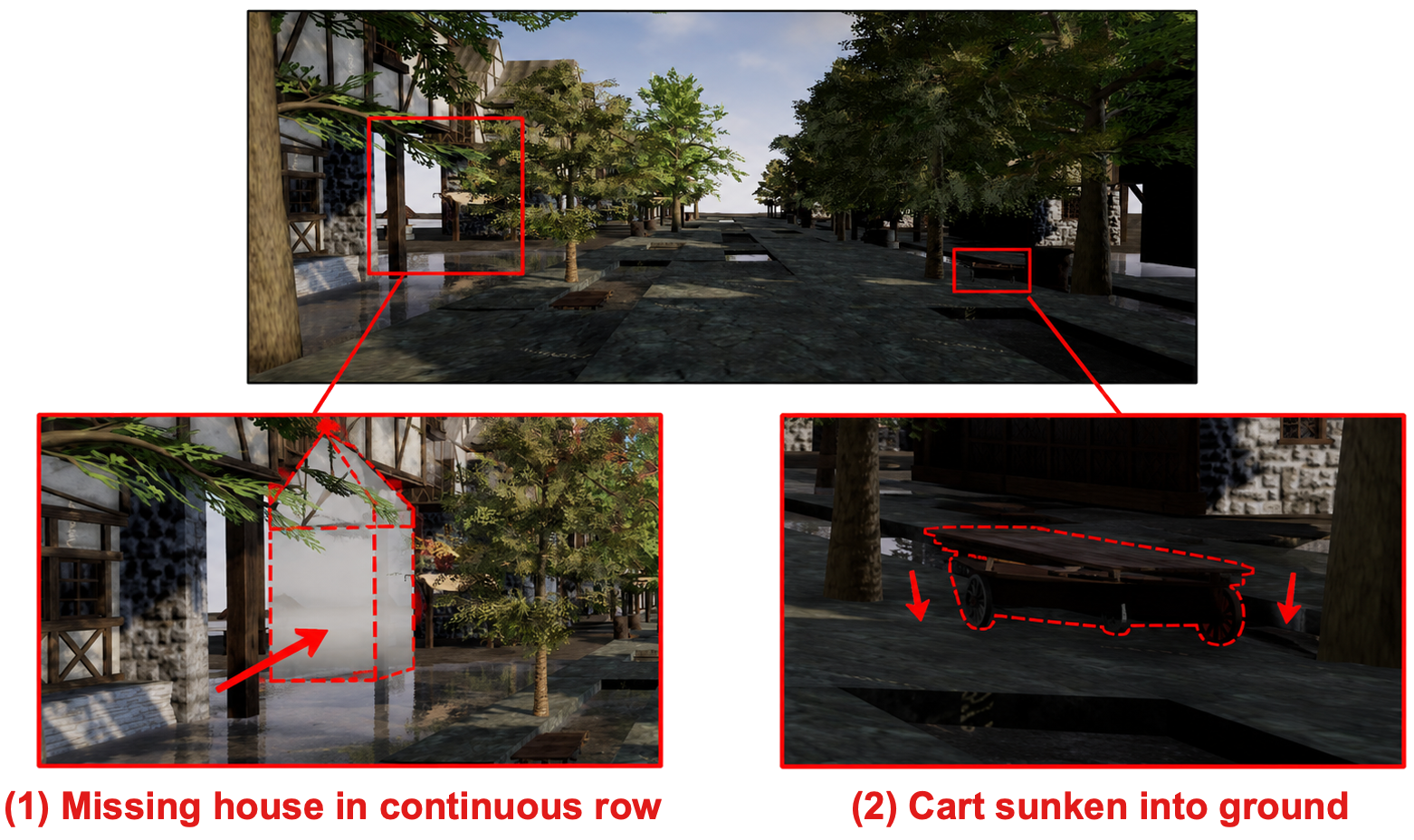}
    {0.52}{0.58}{0.70}{0.66}{0.48}{0.56}{0.58}

\vspace{0.35mm}

\SceneCard
    {Qwen3.5-9B}
    {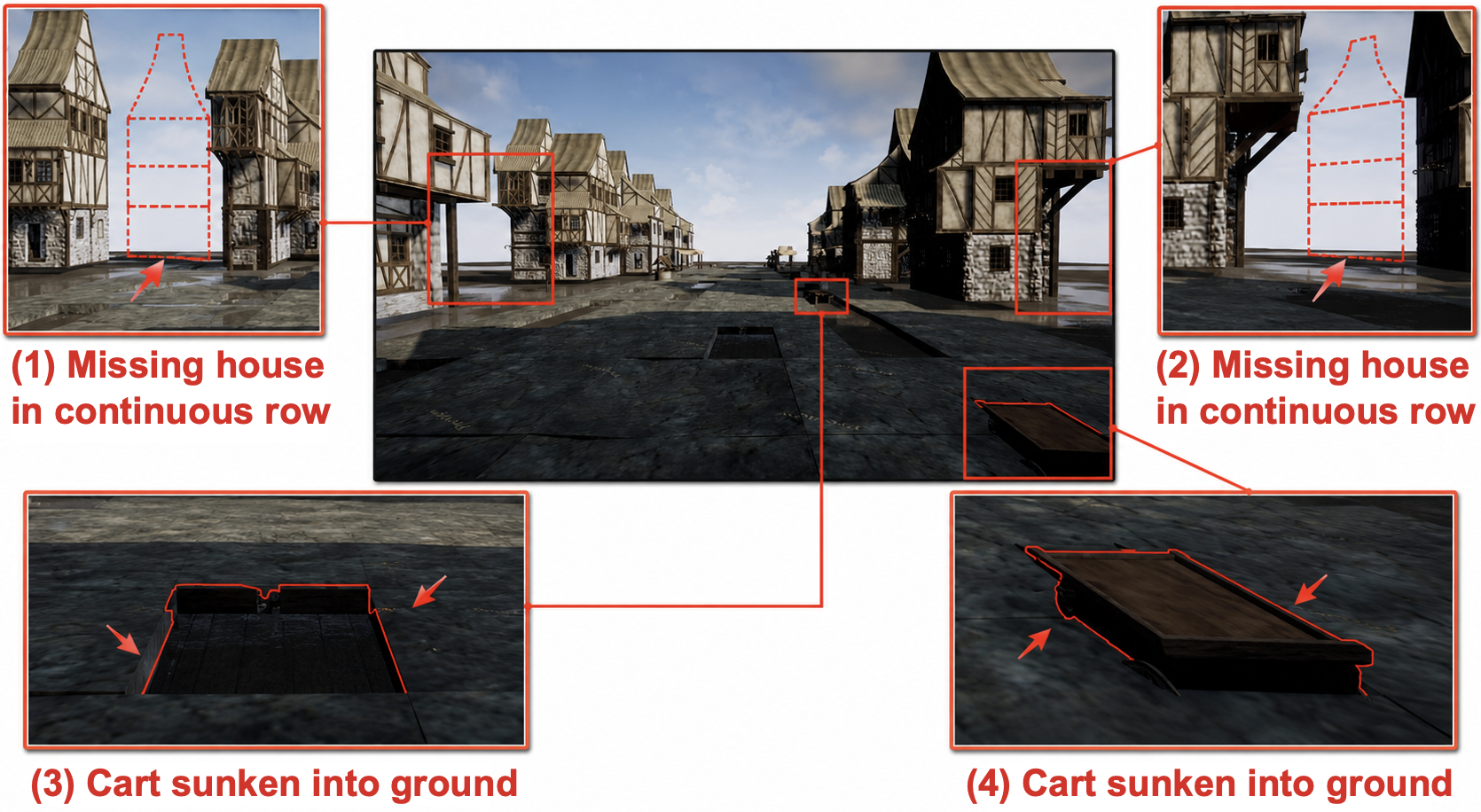}
    {0.34}{0.46}{0.45}{0.38}{0.30}{0.42}{0.39}

\vspace{0.4mm}

\caption{\textbf{Qualitative medieval village scene comparison.}
Each generated scene is paired with its corresponding text-to-scene evaluation scores. Scores are normalized to $[0,1]$; higher is better.}
\label{fig:qualitative_medieval_outputs}

\end{minipage}
\vspace*{-6mm}
\end{figure}

\clearpage
\definecolor{boxBg}{HTML}{FFFFFF}
\definecolor{boxFrame}{HTML}{D0D7DE}
\definecolor{boxTitleBg}{HTML}{F6F8FA}
\definecolor{boxText}{HTML}{24292F}
\definecolor{boxMuted}{HTML}{57606A}

\definecolor{promptAccent}{HTML}{6F42C1}
\definecolor{promptSoft}{HTML}{FFF8F2}

\definecolor{mcpAccent}{HTML}{8250DF}
\definecolor{mcpBlue}{HTML}{0969DA}
\definecolor{mcpGreen}{HTML}{1A7F37}
\definecolor{mcpSoft}{HTML}{F6F2FF}

\definecolor{skillAccent}{HTML}{0969DA}
\definecolor{skillSoft}{HTML}{F6F8FA}

\definecolor{codeBg}{HTML}{F6F8FA}
\definecolor{codeComment}{HTML}{1A7F37}

\tcbset{
    unifiedbox/.style={
        enhanced,
        colback=boxBg,
        colframe=boxFrame,
        boxrule=0.7pt,
        arc=2.5mm,
        left=4mm,
        right=4mm,
        top=3mm,
        bottom=3mm,
        before skip=1.5mm,
        after skip=2mm,
        fonttitle=\sffamily\small\bfseries,
        coltitle=boxText,
        colbacktitle=boxTitleBg,
        boxed title style={
            colback=boxTitleBg,
            colframe=boxFrame,
            boxrule=0.4pt,
            arc=1.5mm,
            left=2mm,
            right=2mm,
            top=1mm,
            bottom=1mm
        }
    }
}

\lstdefinestyle{simcoderPython}{
    language=Python,
    basicstyle=\ttfamily\scriptsize,
    keywordstyle=\color{mcpAccent}\bfseries,
    stringstyle=\color{mcpBlue},
    commentstyle=\color{codeComment}\itshape,
    identifierstyle=\color{boxText},
    showstringspaces=false,
    breaklines=true,
    columns=fullflexible,
    keepspaces=true,
    frame=none,
    backgroundcolor=\color{codeBg},
    aboveskip=0pt,
    belowskip=0pt
}

\begin{center}
\begin{minipage}{\linewidth}
\centering


\begin{tcolorbox}[
    unifiedbox,
    title={
        (a) MCP Tool
        \hfill
    }
]
\newcommand{\req}{\textsuperscript{*}}

\begin{tcolorbox}[
    enhanced,
    colback=mcpSoft,
    colframe=boxFrame,
    boxrule=0.4pt,
    arc=1.5mm,
    left=2.5mm,
    right=2.5mm,
    top=1.2mm,
    bottom=1.2mm
]
\ttfamily\small
\textcolor{mcpAccent}{def}
\textcolor{mcpBlue}{ execute\_python\_script}\textcolor{boxText}{(}
\textcolor{mcpGreen}{script}\textcolor{boxText}{: str, }
\textcolor{mcpGreen}{timeout}\textcolor{boxText}{: int = 60}
\textcolor{boxText}{) -> dict}
\end{tcolorbox}

\vspace{1mm}

\sffamily\scriptsize
Runs a Python script inside UE5's embedded interpreter. The script can access \texttt{unreal} and staged skill modules such as \texttt{road}, \texttt{buildings}, \texttt{street\_dressing}, \texttt{people}, and \texttt{lighting}.

\vspace{1mm}

\scriptsize
\setlength{\tabcolsep}{5pt}
\renewcommand{\arraystretch}{1.1}
\begin{tabularx}{\linewidth}{@{}
    >{\ttfamily\bfseries\color{mcpGreen}}p{0.20\linewidth}
    >{\ttfamily\color{mcpAccent}}p{0.12\linewidth}
    X@{}}
\textcolor{boxMuted}{parameter} &
\textcolor{boxMuted}{type} &
\textcolor{boxMuted}{description} \\
\midrule
script\req & str & Python source to execute; one call builds the scene. \\
timeout & int & Max execution time in seconds. Default: \texttt{60}. \\
\end{tabularx}

\end{tcolorbox}

\vspace{-1mm}


\begin{tcolorbox}[
    unifiedbox,
    title={
        (b) Skill
        \hfill
    }
]

\sffamily\scriptsize

{\bfseries\textcolor{skillAccent}{\# pack\_flush\_row}}

\vspace{0.8mm}

\textcolor{boxMuted}{
Theme-agnostic shoulder-to-shoulder building packer. Given a pool of Blueprint IDs, it places mixed-width buildings along \texttt{+X} with facades flush to a wall line.
}

\vspace{1.5mm}

\setlength{\tabcolsep}{5pt}
\renewcommand{\arraystretch}{1.1}
\begin{tabularx}{\linewidth}{@{}
    >{\ttfamily\bfseries}p{0.18\linewidth}
    >{\ttfamily}p{0.12\linewidth}
    X@{}}
\toprule
\textbf{Param} & \textbf{Type} & \textbf{Use} \\
\midrule
asset\_ids & list & Blueprint pool controlling the visual style. \\
x\_range & tuple & Packing extent \texttt{(x\_min, x\_max)}. \\
flush\_y & float & Wall-line Y coordinate. \\
side & str & \texttt{"south"} or \texttt{"north"} controls side and yaw. \\
seed & int & Randomizes ordering without mirroring rows. \\
\bottomrule
\end{tabularx}

\end{tcolorbox}

\vspace{-1mm}


\begin{tcolorbox}[
    unifiedbox,
    title={
        (c) Generated Code
        \hfill
    }
]

\begin{tcolorbox}[
    enhanced,
    colback=codeBg,
    colframe=boxFrame,
    boxrule=0.4pt,
    arc=1.5mm,
    left=2mm,
    right=2mm,
    top=1.2mm,
    bottom=1.2mm
]
\begin{lstlisting}[style=simcoderPython]
# Narrow cobble lane centered on the origin
x_min, x_max, y_min, y_max, _ = lay_cobble_path(
    length_tiles=26, width_tiles=5,
    start_x=-4550, start_y=-875, seed=42
)

# Pack medieval buildings tightly along both street edges
MED_POOL = [MED_BP_TAVERN, MED_BP_HOUSE_1, MED_BP_HOUSE_2, MED_BP_HOUSE_3]

pack_flush_row(MED_POOL, (x_min, x_max), flush_y=y_min,
    side="south", margin=350, seed=11, label_prefix="Bldg_Med_S_")
pack_flush_row(MED_POOL, (x_min, x_max), flush_y=y_max,
    side="north", margin=350, seed=29, label_prefix="Bldg_Med_N_")

# Add household clutter and finish the scene
dress_each_house("Bldg_Med_S_", side="south", props_per_house_range=(0, 4), seed=101)
dress_each_house("Bldg_Med_N_", side="north", props_per_house_range=(0, 4), seed=202)
place_well(x=-2730, y=0, label="Med_Well_Square")
apply_daytime_medieval_lighting(time_of_day="morning")  # daytime preset
\end{lstlisting}
\end{tcolorbox}

\end{tcolorbox}

\captionof{figure}{\textbf{Example of tool, skill, and generated script used in the example in Fig~\ref{fig:qualitative_medieval_outputs}.}
The MCP tool is presented as a Python function, the skill as Markdown-style documentation, and the generated scene program as executable Python.}
\label{fig:qualitative_medieval_tool_skill_code}

\end{minipage}
\end{center}
\clearpage

\clearpage

\subsection{Case Study 2: Can embodied agents learn useful navigation abilities from \ours{}-generated environments?}
\vspace{-2mm}
\label{sec:case_study_2}

We evaluate whether LLM-based embodied agents can learn navigation policies in
\ours{}-generated environments and generalize to unseen scenes. As shown in
Figure~\ref{fig:case_studies} (Case Study 2), an agent navigates a
\textsc{SimCoder}-generated city from RGB, bearing, and distance observations.
Navigation serves as a functional probe of environment quality, since failures
may reflect either weak policies or defects in the generated world.
\vspace{-1mm}

\textbf{Settings.}
We consider two outdoor tasks: \emph{Point Navigation} (PointNav),
where the agent navigates to a target coordinate, and \emph{Object Navigation}
(ObjectNav), where it locates a target object category. Episodes are generated on
\ours{} maps via UE5 NavMesh by sampling start--goal pairs, computing
shortest-path references, and filtering by reachability and path length.
For each task, we sampled 1.2K episodes as our training set and evaluated agents on 329 held-out \ours{} episodes; we additionally evaluate on the external SimWorld-MMNav benchmark~\citep{zhuang2026simworldrobotics} for cross-benchmark transfer. To assess how scene diversity affects downstream learning, we vary the number of training environments from 1 to 30 with the training budget fixed at 200 episodes.
\vspace{-1mm}

\textbf{Metrics.}
We report Success Rate (SR) for \emph{task success}, Success weighted by Path Length (SPL) and Soft Success weighted by Path Length (SoftSPL) for
\emph{path efficiency}, and normalized Dynamic Time Warping (nDTW) for \emph{trajectory fidelity}. Definitions are in Appendix~\ref{app:cs2_metric_def}.
\vspace{-1mm}

\textbf{Models.}
We evaluate Qwen3.5-2B/9B/27B~\citep{qwen2026qwen35} with and without agent memory. Our memory design is inspired by Generative Agents~\citep{park2023generativeagentsinteractivesimulacra} and ExpeL~\citep{zhao2024expel}, but extends them with \emph{multi-level} updates at step, trajectory, and task granularities, allowing distilled strategies to be retrieved at the appropriate scope during inference (Appendix~\ref{app:memory}).
\vspace{-2mm}
\subsubsection{Results}
\vspace{-1mm}
\input{Tables/case_study_2/main}

\begin{figure}[h]
    \vspace{0mm}
    \centering
    \begin{subfigure}[t]{0.46\linewidth}
        \centering
        \includegraphics[width=\linewidth]{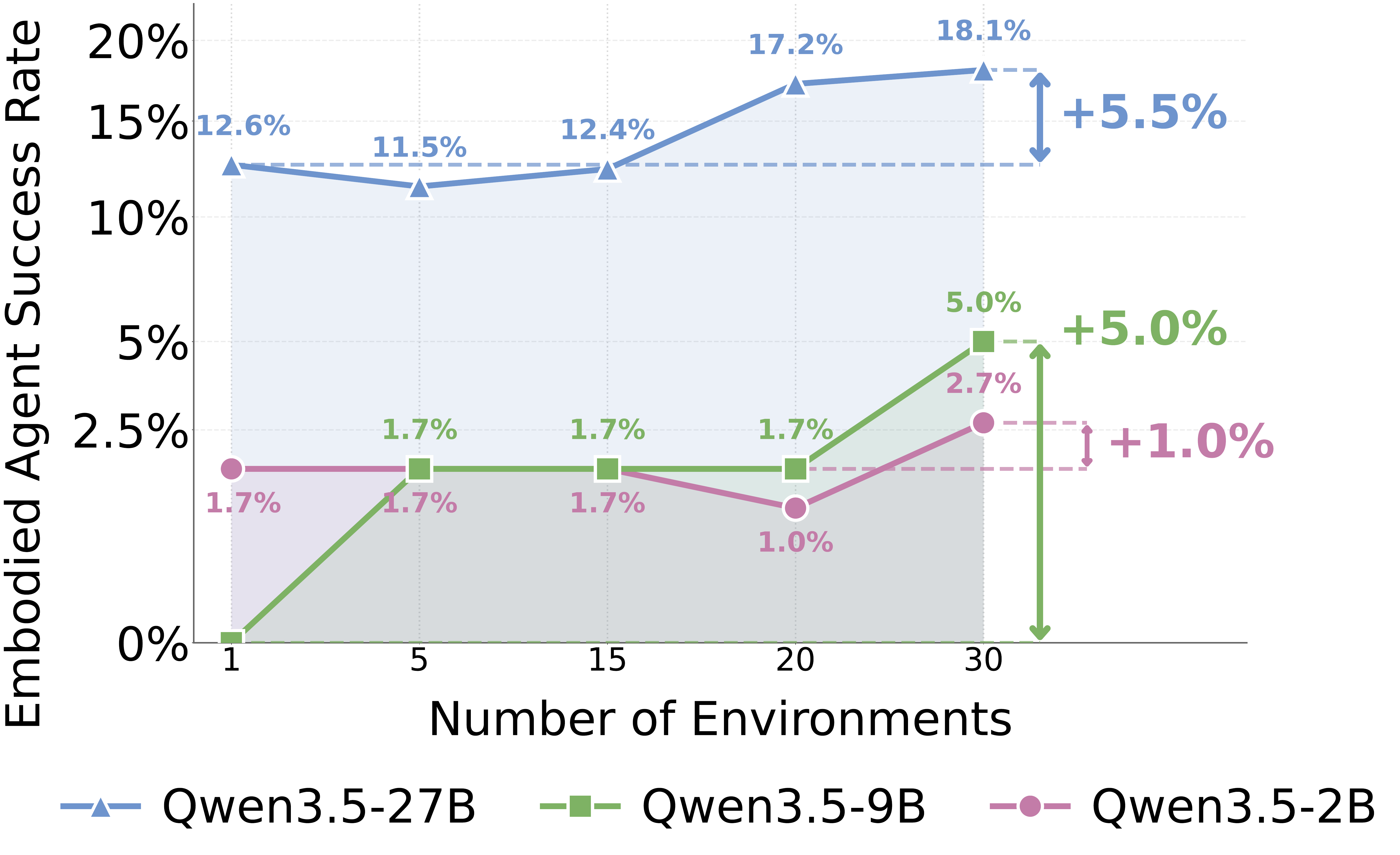}
        \label{fig:ablation_env}
    \end{subfigure}
    \hfill
    \vspace{3mm}
    \begin{subfigure}[t]{0.45\linewidth}
        \centering
        \includegraphics[width=0.95\linewidth]{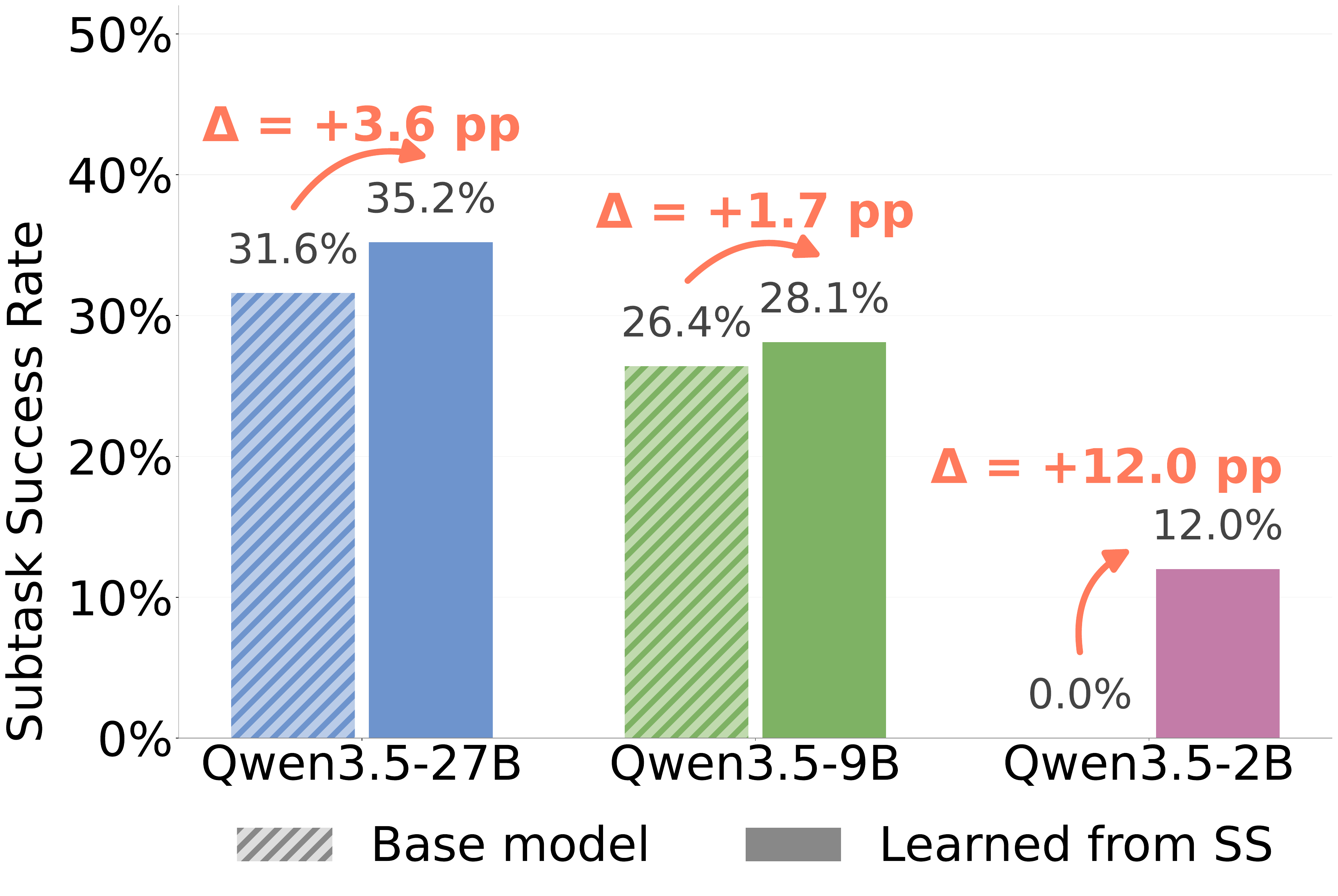}
        \label{fig:ablation_generalization}
    \end{subfigure}
    \vspace{-25pt}
    \caption{\textbf{Generalization analysis.}
    \textbf{(Left)} More diverse \ours{} environments yield stronger test-time
    generalization. \textbf{(Right)} Embodied agents learned in \ours{} transfer to SimWorld-MMNav across model scales.}
    \label{fig:generalization_analysis}
    \vspace{-2pt}
\end{figure}

\textbf{\ours{} improves the embodied agents in both held-out environments and external benchmark.}
On held-out \ours{} test environments (Table~\ref{tab:cs2_main_results}),
agents trained in \ours{} substantially outperform the no-training baseline
across all model scales (e.g., Qwen3.5-9B: $+14.62\%$ SR on ObjectNav).
These gains transfer to the external SimWorld-MMNav benchmark (Figure~\ref{fig:generalization_analysis}(Right); $+12.0$pp on 2B), indicating that \ours{} captures transferable navigation knowledge rather than environment-specific shortcuts.

\textbf{\ours{}'s open-ended environment generation capability translates into stronger embodied agent
generalization.} A central capability of \ours{} is generating an essentially unbounded number of distinct environments on demand, and we find this has a measurable downstream effect: test success rises with the number of distinct \ours{}-generated training environments (Figure~\ref{fig:generalization_analysis}(Left); $+5.5$pp on Qwen3.5-27B). 
Since the training budget is fixed, these gains are attributable to scene diversity rather than additional experience, which means the platform's scalable generation directly converts into stronger embodied agents. We additionally ablate observation modalities, finding that \ours{}'s built-in RGB-D interface outperforms text-only inputs by providing complementary geometric and semantic cues (Table~\ref{tab:ablation_obs}, Appendix~\ref{app:obs_abl}).

\vspace{-3mm}
\subsection{Case Study 3: Can \textsc{SimCoder} co-evolve with embodied agents with agent feedback?}
\vspace{-2mm}
\label{sec:case_study_3}

\begin{figure}[h]
    \centering
    \includegraphics[width=\linewidth]{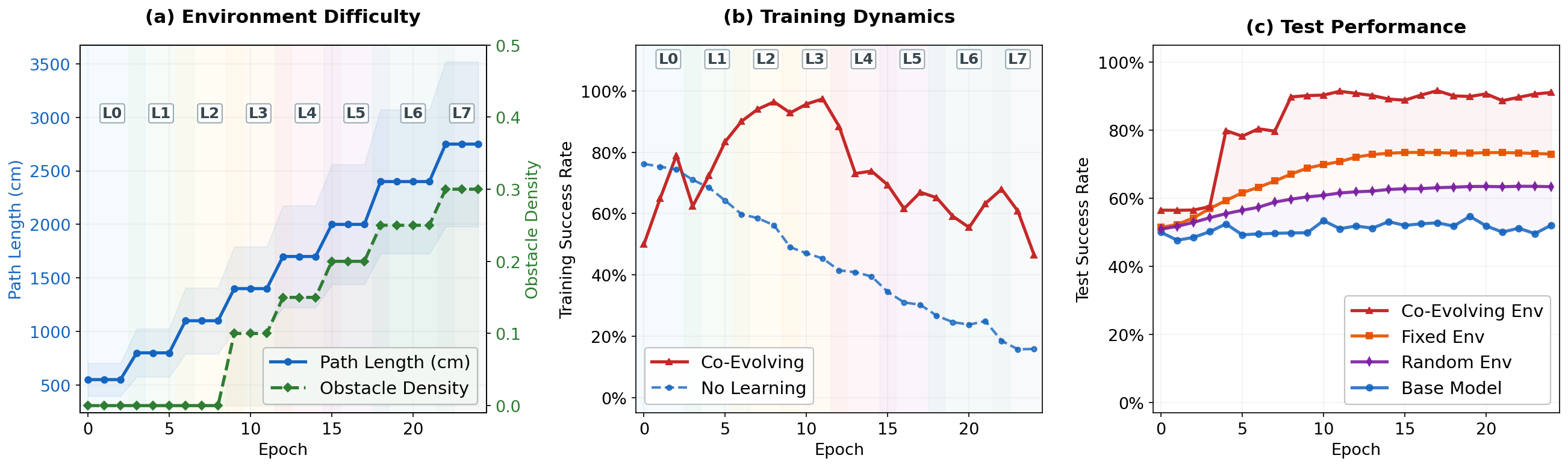}
    \vspace{-4mm}
    \caption{
        \textbf{Co-evolution of \textsc{SimCoder} and embodied agent.}
        \textbf{(a)}~Environment difficulty across 8~levels.
        \textbf{(b)}~Training dynamics: the co-evolving agent drops at each level transition then recovers.
        \textbf{(c)}~Test performance on the SimWorld-MMNav benchmark.
    }
    \label{fig:coevolve_dynamics}
    \vspace{-4mm}
\end{figure}

Case Studies 1 and 2 evaluate \textsc{SimCoder} and the embodied agent in isolation. As a demonstration of the platform's closed-loop capability, Case Study 3 instantiates a simple performance-gated adaptive curriculum. As shown in Figure~\ref{fig:coevolve_dynamics} (a), \textsc{SimCoder} generates progressively harder environments while the embodied agent updates its policy from experience, with each agent's output steering the other's next step via the co-evolution framework of \S\ref{sec:coevolution}.
\vspace{-2mm}

\paragraph{Method.}
We instantiate a specific adaptive curriculum under the co-evolution framework of \S\ref{sec:coevolution}. Navigation difficulty is parameterized along two axes: path length and obstacle density, which are jointly quantized into eight levels of increasing challenge. Each co-evolution round proceeds in three explicit steps: \textbf{(i) Environment Generation:} \textsc{SimCoder} generates a batch of navigation episodes at the current difficulty level. \textbf{(ii) Embodied Agent Learning:} The embodied agent attempts the batch and records any failed episodes. It then distills these failure trajectories into a small set of prioritized decision rules (e.g., \textit{``if the goal lies behind the agent, turn around before exploring forward''}). Crucially, it \emph{appends} these to its existing rule set. Unlike GEPA~\citep{li2025gepa}, which rewrites the full prompt each round, this incremental accumulation preserves proven strategies from earlier levels while adding new failure-derived corrections. \textbf{(iii) Environment Adaptation:} \textsc{SimCoder} evaluates the agent's mean Success rate on the batch. If the agent clears a level-specific mastery threshold, \textsc{SimCoder} advances the next batch of environments to a higher difficulty; otherwise, it holds the current level until the agent catches up. Full details are in Appendix~\ref{app:case_study_3}.
\vspace{-4mm}
\paragraph{Setting.}
We evaluate on the external SimWorld-MMNav benchmark~\citep{zhuang2026simworldrobotics} and compare four conditions using Qwen3.5-9B: \textbf{(1)}~co-evolving environment (our full closed-loop system), \textbf{(2)}~fixed-difficulty environment (held constant at level 3), \textbf{(3)}~random-difficulty environment (random level from 0 to 7), and \textbf{(4)}~base model (no embodied agent learning).

\vspace{-3mm}
\subsubsection{Results}
\vspace{-2mm}

\vspace{2pt}

\vspace{-1mm}

\textbf{Adaptive curricula drive continuous improvement and prevent early saturation.} The training dynamics of the co-evolving system (Figure~\ref{fig:coevolve_dynamics}b) exhibit a characteristic \emph{drop-and-recover} pattern~\citep{vygotsky1978mind}: the agent's Success rate dips exactly when \textsc{SimCoder} introduces harder tasks, but recovers rapidly as distilled strategies transfer to the new difficulty tier. Ultimately, this closed-loop co-evolution yields a 90\% test Success rate on SimWorld-MMNav. This represents a massive 18-point gain over the fixed-environment learning baseline (72\%), which saturates midway through training because it is not pushed to its capability frontier, and a 40-point gain over the untrained baseline (50\%; Figure~\ref{fig:coevolve_dynamics}c).


%% file: Tables/case_study_2/main.tex
\definecolor{mettask}{HTML}{0961F6}   
\definecolor{meteff}{HTML}{3A7D44}    
\definecolor{mettraj}{HTML}{7A5CCF}   

\newcommand{\metrichead}[2]{\textcolor{#1}{\scriptsize #2}}

\begin{table}[h]
\centering
\begingroup
\setlength{\abovecaptionskip}{0pt}
\setlength{\belowcaptionskip}{-3pt}
\setlength{\aboverulesep}{0.25ex}
\setlength{\belowrulesep}{0.25ex}
\setlength{\cmidrulesep}{0.15ex}
\setlength{\tabcolsep}{3pt}
\renewcommand{\arraystretch}{0.88}
\scriptsize

\caption{\textbf{Embodied navigation on generated environments.}
We report SR, SPL, SoftSPL, and nDTW. Metric colors denote
\textcolor{mettask}{task success}, \textcolor{meteff}{path efficiency}, and
\textcolor{mettraj}{trajectory fidelity}. \textbf{Bold} marks the best result
within each column; \colorbox{green!15}{green} marks improvement >0.05 over baseline.}
\label{tab:cs2_main_results}


\resizebox{\textwidth}{!}{%
\begin{tabular}{@{}ll cccc cccc@{}}
\toprule
\textbf{LLM} & \textbf{Setting}
& \multicolumn{4}{c}{\textbf{Point Navigation}}
& \multicolumn{4}{c}{\textbf{Object Navigation}} \\
\cmidrule(lr){3-6} \cmidrule(lr){7-10}

&
& \metrichead{mettask}{SR(\%, $\uparrow$)}
& \metrichead{meteff}{SPL($\uparrow$)}
& \metrichead{meteff}{SoftSPL($\uparrow$)}
& \metrichead{mettraj}{nDTW($\uparrow$)}
& \metrichead{mettask}{SR(\%, $\uparrow$)}
& \metrichead{meteff}{SPL($\uparrow$)}
& \metrichead{meteff}{SoftSPL($\uparrow$)}
& \metrichead{mettraj}{nDTW($\uparrow$)} \\

\midrule
\multirow{2}{*}{Qwen3.5-27B}
& Baseline
& 20.21 & 0.0931 & 0.0882 & 0.2322
& 22.56 & 0.0795 & 0.0782 & 0.2486 \\

& + Memory
& \cellcolor{green!15}\textbf{26.76}
& \textbf{0.1266}
& \textbf{0.1036}
& \textbf{0.2676}
& \cellcolor{green!15}\textbf{31.59}
& \cellcolor{green!15}\textbf{0.1429}
& \textbf{0.1161}
& \cellcolor{green!15}\textbf{0.3159} \\

\midrule
\multirow{2}{*}{Qwen3.5-9B}
& Baseline
& 0.40 & 0.0039 & 0.1137 & 0.1137
& 0.00 & 0.0000 & 0.1160 & 0.1160 \\

& + Memory
& \cellcolor{green!15}\textbf{7.43}
& \cellcolor{green!15}\textbf{0.0722}
& \cellcolor{green!15}\textbf{0.2617}
& \cellcolor{green!15}\textbf{0.2617}
& \cellcolor{green!15}\textbf{14.62}
& \cellcolor{green!15}\textbf{0.1408}
& \cellcolor{green!15}\textbf{0.3059}
& \cellcolor{green!15}\textbf{0.3059} \\

\midrule
\multirow{2}{*}{Qwen3.5-2B}
& Baseline
& 1.38 & 0.0138 & 0.1030 & \textbf{0.0034}
& 0.24 & 0.0024 & 0.1080 & \textbf{0.0041} \\

& + Memory
& \textbf{6.23}
& \textbf{0.0528}
& \cellcolor{green!15}\textbf{0.2119}
& 0.0022
& \textbf{4.58}
& \textbf{0.0389}
& \cellcolor{green!15}\textbf{0.1841}
& 0.0036 \\

\bottomrule
\end{tabular}%
}
\vspace{-2mm}
\endgroup
\end{table}

%% file: Section/2-Related_Work.tex
\vspace{-2mm}
\section{Related Work}
\vspace{-1mm}
\label{sec:related}

\paragraph{Embodied simulation platforms.}\label{sec:related-sim}
Embodied AI research relies on three families of interactive simulators, each with structural limitations. \emph{Hand-built indoor and outdoor platforms} support navigation, manipulation, driving, urban robotics, and language-grounded games~\citep{kolve2022ai2thorinteractive3denvironment, li2021igibson, puig2023habitat30cohabitathumans, li2024behavior1khumancenteredembodiedai, cheng-etal-2024-legent, dosovitskiy2017carlaopenurbandriving, wang2024grutopiadreamgeneralrobots, wu2024metaurbanembodiedaisimulation, gao2024embodied, shridhar2021alfworldaligningtextembodied, fan2022minedojobuildingopenendedembodied, zhong2025unrealzooenrichingphotorealisticvirtual, swain2026virtualenvplatformembodiedai}, but rely on manually authored scene catalogs that are fixed and expensive to extend. \emph{Procedural generators}~\citep{deitke2022procthorlargescaleembodiedai, raistrick2023infinite} scale environment count but remain constrained by hand-designed templates and rules. \emph{LLM-based scene synthesizers}~\citep{yang2024holodecklanguageguidedgeneration, hu2024scenecraftllmagentsynthesizing, xia2026sagescalableagentic3d, zhang2026code2worlds, pfaff2026scenesmith} enable open-ended diversity, but output static 3D content only evaluated as visual artifacts without task definitions, agent interfaces, or learning signals. \ours{} combines the photorealism of hand-built UE5 platforms with the open-ended diversity of agentic generation, and closes the loop between scene generation and embodied training: every generated scene is exported through a standard Gym interface and adapted based on downstream agent feedback (Table~\ref{tab:comparison}).
\vspace{-3mm}
\paragraph{Agent--environment co-evolution.}\label{sec:related-coevol}
Prior work on \emph{unsupervised environment design} edits parametric environments to keep difficulty near the agent's frontier~\citep{dennis2020emergent, jiang2020prioritized, parker2022evolving, wang2019poet, samvelyan2023maestro}; LLM-based extensions such as EnvGen and Eureka revise reward functions, configurations, or task programs from agent feedback~\citep{guo2025genenvdifficultyalignedcoevolutionllm, ma2023eureka, faldor2024omniepic, liang2024eurekaverse}, but only tune parameters of pre-built simulators rather than construct scenes. Closer to our setting, EnvGen~\citep{zala2024envgengeneratingadaptingenvironments} co-trains an LLM generator with an RL agent in a controlled gridworld, whereas \textsc{SimCoder} synthesizes full photorealistic UE5 scenes from scratch through engine-level tool calls and reusable skills; and Agent-World~\citep{dong2026agentworldscalingrealworldenvironment} mines MCP databases for \emph{digital}-agent training, whereas our generated environments expose RGB-D observations, agent pose, and physical reward through a standard Gym contract for \emph{embodied} policies. This gap is especially pronounced for embodied training, where useful environments must be not only adaptive but also photorealistic, physically plausible, and scalable. See additional related work in Appendix~\ref{app:relatedwork}.

\begin{table}[t]
\centering
\small
\caption{Comparison with representative embodied simulation platforms, generative scene-construction systems, and environment-agent co-evolution methods.
\textbf{Diverse Gen.}: supports diverse environment generation at scale.
\textbf{Phys./Vis.\ Realism}: fidelity of physics simulation and visual rendering.
\textbf{Gym Interface}: provides a standard Gymnasium-compatible API.
\textbf{Self-Evolution}: autonomous improvement of the platform's own generation or operation behavior from verifier feedback.
\textbf{Co-Evolution}: environment generation adapts based on downstream embodied agent performance.}
\label{tab:comparison}
\resizebox{\textwidth}{!}{%
\begin{tabular}{llccccc}
\toprule
\textbf{Platform} & \textbf{Engine} & \textbf{Diverse Gen.} & \textbf{Phys./Vis. Realism} & \textbf{Gym Interface} & \textbf{Self-Evolution} & \textbf{Co-Evolution} \\
\midrule
CARLA \citep{dosovitskiy2017carlaopenurbandriving}                 & UE4          & \xmark & {+++} & \xmark & \xmark & \xmark \\
ThreeDWorld \citep{gan2021threedworldplatforminteractivemultimodal} & Unity        & \xmark & {++}  & \xmark & \xmark & \xmark \\
AI2-THOR \citep{kolve2022ai2thorinteractive3denvironment}          & Unity        & \xmark & {++}  & \xmark & \xmark & \xmark \\
MineDojo \citep{fan2022minedojobuildingopenendedembodied}          & Minecraft    & \xmark & {+}   & \cmark & \xmark & \xmark \\
ProcTHOR \citep{deitke2022procthorlargescaleembodiedai}            & Unity        & \cmark & {++}  & \xmark & \xmark & \xmark \\
Habitat 3.0 \citep{puig2023habitat30cohabitathumans}               & Habitat-Sim  & \xmark & {++}  & \cmark & \xmark & \xmark \\
MetaUrban \citep{wu2024metaurbanembodiedaisimulation}              & PyBullet     & \cmark & {++}  & \cmark & \xmark & \xmark \\
GRUtopia \citep{wang2024grutopiadreamgeneralrobots}                & Isaac Sim    & \cmark & {++}  & \cmark & \xmark & \xmark \\
EmbodiedCity \citep{gao2024embodied}                               & UE4          & \xmark & {+++} & \xmark & \xmark & \xmark \\
UnrealZoo \citep{zhong2025unrealzooenrichingphotorealisticvirtual} & UE4/5        & \xmark & {+++} & \cmark & \xmark & \xmark \\
Virtual Community \citep{zhou2026virtualcommunityopenworld}        & Genesis      & \cmark & {++}  & \xmark & \xmark & \xmark \\
VirtualEnv \citep{swain2026virtualenvplatformembodiedai}           & UE5          & \xmark & {+++} & \xmark & \xmark & \xmark \\
Holodeck \citep{yang2024holodecklanguageguidedgeneration}                                  & AI2-THOR/Unity & \cmark & {++}  & \xmark & \xmark & \xmark \\
SAGE \citep{xia2026sagescalableagentic3d}                                           & Isaac Sim      & \cmark & {+++} & \xmark & \cmark & \xmark \\
GenEnv \citep{guo2025genenvdifficultyalignedcoevolutionllm} & AlfWorld/Text-only & \xmark & {+} & \xmark & \xmark & \cmark \\
\midrule
\textsc{\textbf{SimWorld Studio}}                                  & \textbf{UE5} & \cmark & {+++} & \cmark & \cmark & \cmark \\
\bottomrule
\end{tabular}%
}
\vspace{-2mm}
\end{table}

%% file: Section/5-Conclusion.tex
\vspace{-2mm}
\section{Conclusion}
\vspace{-1mm}

\label{sec:conclusion}
We presented \ours{}, a platform that overcomes the bottleneck of static scene generation by synthesizing scalable, interactive 3D environments for embodied learning. Driven by a self-evolving coding agent, \ours{} automatically translates prompts into Gymnasium-compatible worlds and adapts their difficulty based on the embodied agent's performance. Our results demonstrate that this closed-loop co-evolution prevents training saturation and boosts zero-shot generalization, establishing a self-improving paradigm for embodied AI research.


%% file: Appendix/A-SimWorld_Studio.tex
\newpage
\section{Limitation}
\label{app:conclusion}
A current limitation is that the effectiveness of \ours{} is still bounded by the capability of the underlying coding agent.
In particular, generating complex 3D environments requires strong spatial reasoning: the agent must understand object placement, geometric constraints, physical plausibility, navigability, and long-range layout consistency.
While the current system can already produce useful scenes, failures may still occur when the task requires fine-grained spatial planning or precise multi-object arrangement.
Improving spatial reasoning for coding agents, especially in 3D interactive environments, is therefore an important direction for future work.

\section{Broader Impact}
\label{app:broader_impacts}
\ours{} has the potential to improve productivity in 3D environment creation, especially for complex engines such as Unreal Engine where existing coding-agent support remains limited.
By allowing coding agents to directly edit, validate, and reuse scene-building skills, the system can reduce repetitive engineering effort and make interactive environment construction more accessible to researchers and developers.
It also provides a scalable route for embodied AI research, where the lack of diverse, controllable, and interactive environments is often a major bottleneck.

At the same time, more capable automated scene-generation tools may affect existing workflows in game development, simulation design, and digital-content production.
Some routine environment-authoring tasks could become increasingly automated, which may shift the role of human creators from manual construction toward supervision, design specification, and quality control.
We believe such systems should be developed as assistive tools that augment human creativity and engineering productivity, while preserving human oversight over artistic direction, safety, and deployment decisions.

Potential misuse includes generating restricted, unsafe, or policy-violating simulated environments, using automated scene construction for surveillance-like or tactical planning scenarios, or recombining licensed assets outside their permitted terms. Our release is intended for research use. We do not release trained harmful policies, scraped personal data, or human-subject datasets. The Python execution interface is local/self-hosted, and users are expected to follow the licenses and deployment policies of the engine, model providers, and asset libraries.

\section{\ours{}: Platform Details}
\label{app:platform}

\subsection{MCP Tool Reference}
\label{app:mcp_tools}

Table~\ref{tab:simworld-mcp-tools} lists all 14 MCP tools exposed by the \ours{} server. Tools are organized into four functional groups: actor management, environment and asset management, scene evaluation, and a Python escape hatch for operations not covered by the predefined API.

\begin{table}[h]
\centering
\scriptsize
\caption{\textbf{\ours{} MCP server tool reference.} \texttt{*} = required parameter.}
\label{tab:simworld-mcp-tools}
\setlength{\tabcolsep}{3pt}
\renewcommand{\arraystretch}{1.12}

\begin{tabularx}{\linewidth}{@{}p{2.1cm} p{2.75cm} X p{3.0cm}@{}}
\toprule
\textbf{Group} & \textbf{Tool} & \textbf{Purpose} & \textbf{Key Parameters} \\
\midrule
\multirow{8}{=}{\textit{Actor management}}
 & \nolinkurl{spawn_blueprint_actor} & Spawn a CityDatabase Blueprint, such as a building, tree, vehicle, or prop; accepts full path or shorthand. & \texttt{actor\_name}*, \texttt{blueprint\_id}*, \texttt{location}*, \texttt{rotation}, \texttt{scale} \\
 & \nolinkurl{spawn_actor} & Spawn a static-mesh actor, including engine primitives or any \texttt{SM\_} asset. & \texttt{name}*, \texttt{static\_mesh}*, \texttt{location}*, \texttt{rotation}, \texttt{scale} \\
 & \nolinkurl{delete_actor} & Delete a named actor. & \texttt{name}* \\
 & \nolinkurl{delete_all_spawned} & Delete all session-spawned actors. & --- \\
 & \nolinkurl{get_actors_in_level} & List every actor currently in the UE level. & --- \\
 & \nolinkurl{find_actors_by_name} & Search actors by name pattern. & \texttt{pattern}* \\
 & \nolinkurl{set_actor_transform} & Move, rotate, or scale an actor. & \texttt{name}*, \texttt{location}, \texttt{rotation}, \texttt{scale} \\
 & \nolinkurl{take_screenshot} & Capture the UE viewport as PNG. & \texttt{filename} \\
\midrule
\multirow{2}{=}{\textit{Environment \& assets}}
 & \nolinkurl{setup_environment} & Initialize lighting, sky, fog, ground plane, and view distance. Must be called before spawning assets. & \texttt{ground\_size}, \texttt{time\_of\_day} \\
 & \nolinkurl{list_assets} & List available SimWorld assets by category, including buildings, trees, vehicles, street furniture, and roads. & \texttt{category} \\
\midrule
\multirow{3}{=}{\textit{Scene evaluation}}
 & \nolinkurl{verify_scene} & VLM verifier: captures the scene and evaluates placement against the original request; returns \texttt{PASS}, \texttt{NEEDS\_IMPROVEMENT}, or \texttt{FAIL} with actionable suggestions. & \texttt{original\_request}*, \texttt{focus\_areas} \\
 & \nolinkurl{check_collisions} & Geometric overlap check: computes world-space AABBs and returns intersecting pairs with area in cm$^2$. & \texttt{names}, \texttt{scope}, \texttt{min\_area\_cm2} \\
 & \nolinkurl{check_vertical_support} & Detects floating objects by reporting actors whose AABB bottom is above ground and not supported by another actor. & \texttt{names}, \texttt{scope}, \texttt{ground\_z}, \texttt{tolerance\_cm} \\
\midrule
\textit{Python escape hatch}
 & \nolinkurl{execute_python_script} & Execute arbitrary Unreal Engine Python for operations not covered by dedicated tools; successful scripts can be saved as reusable skills. & \texttt{script}* \\
\bottomrule
\end{tabularx}
\end{table}

\subsection{\textsc{SimCoder} Generation Pipeline}
\label{app:simcoder_pipeline}

Figure~\ref{fig:platform} illustrates the overall \textsc{SimCoder} architecture. Here we describe each stage of the generation pipeline in detail.

\paragraph{Stage 1: Context acquisition.}
\textsc{SimCoder} begins every generation episode by querying its context through tool calls rather than a manually engineered prompt. It calls \texttt{list\_assets} to enumerate available mesh categories and asset counts, queries the \texttt{SkillRegistry} to retrieve applicable skills (e.g., \textit{building-placement}, \textit{city-layout}), and calls \texttt{setup\_environment} to initialize the scene. This tool-driven context acquisition ensures that \textsc{SimCoder} always operates with an accurate, up-to-date view of its action space.

\paragraph{Stage 2: Layout planning.}
Given the natural-language prompt and optional reference image, \textsc{SimCoder} drafts a high-level layout plan specifying spatial zones (e.g., street grid, park, commercial row), asset categories per zone, and approximate density. For image-guided generation (S2), the plan is anchored to spatial structure inferred from the reference. Layout planning is performed in-context; the plan is not externalized but is used to sequence subsequent tool calls.

\paragraph{Stage 3: Iterative scene construction.}
\textsc{SimCoder} builds the scene incrementally, spawning and arranging actors via actor-management tools. After each spawn batch, it calls \texttt{check\_collisions} and \texttt{check\_vertical\_support}; violations are returned inline as tool-call responses and resolved before proceeding. This per-call verification loop (\S\ref{sec:verifier}) prevents error compounding across a long construction trajectory.

\paragraph{Stage 4: VLM-based semantic verification.}
After completing a logical block of construction, \textsc{SimCoder} calls \texttt{verify\_scene} with the original prompt. The VLM (Claude in our experiments) receives six multi-angle screenshots and the current actor list, scores semantic alignment, and returns structured feedback (PASS / NEEDS\_IMPROVEMENT / FAIL) with specific issue descriptions. If the verdict is not PASS, \textsc{SimCoder} performs targeted corrections and re-verifies. The generation episode terminates on PASS or after a maximum of three verification rounds.

\paragraph{Stage 5: Skill authoring (self-evolution).}
When verifier feedback reveals a failure pattern that recurs across episodes (e.g., commercial buildings colliding due to under-estimated footprints), \textsc{SimCoder} authors a corrective skill or tool wrapper via \texttt{execute\_python\_script} and writes it to the skill directory. On the next skill-registry refresh, the new skill is indexed and becomes available to all future generations, permanently extending the agent's capability without manual intervention.

\subsection{Skill Library Structure}
\label{app:skill_library}

Each skill is a Markdown document with a YAML front-matter header and an optional companion Python utility:

\begin{verbatim}
---
name: building-placement
version: 1.2
tags: [placement, spacing, buildings]
dependencies: [city-layout]
python_util: building_placement_utils.py
---
## Summary
Footprint-aware spacing rules for urban building placement ...
## Usage
Call `compute_building_grid(n_buildings, style)` to get
a list of (location, rotation) placements respecting
minimum inter-building clearance per size category.
\end{verbatim}

The five built-in skills cover: \textbf{building-placement} (size categories, spacing tables, rotation patterns across 127 building classes), \textbf{city-layout} (grid blocks, street-facing rows, mixed-use neighborhoods), \textbf{street-furniture} (placement heuristics for trees, benches, lamps, and signs), \textbf{weather-and-mood} (lighting and time-of-day presets), and \textbf{screenshot-tour} (multi-angle capture patterns for the VLM verifier). Self-authored skills follow the same schema, ensuring seamless integration into the registry.

\input{Appendix/SimCoder_RunningCase}

\subsection{\ours{}'s Graphic User Interface}
\label{app:studio_interface}

Figure~\ref{fig:app_studio_ui_all} presents representative interface views of \textsc{SimWorld Studio}. 
The first panel shows the main studio interface in a light theme, illustrating the integrated workflow across user--agent interaction, UE rendering, asset/backend services, Gym APIs, and embodied-agent monitoring. 
The remaining dark-theme panels show specialized views for skill management, tool abstraction, and interactive user control inside the generated Unreal Engine environment.

\Needspace{0.35\textheight}
\begin{interfacecasebox}{Appendix Case B: Skills Panel}
    \vspace{0.5mm}
    \centering
    \includegraphics[width=\linewidth]{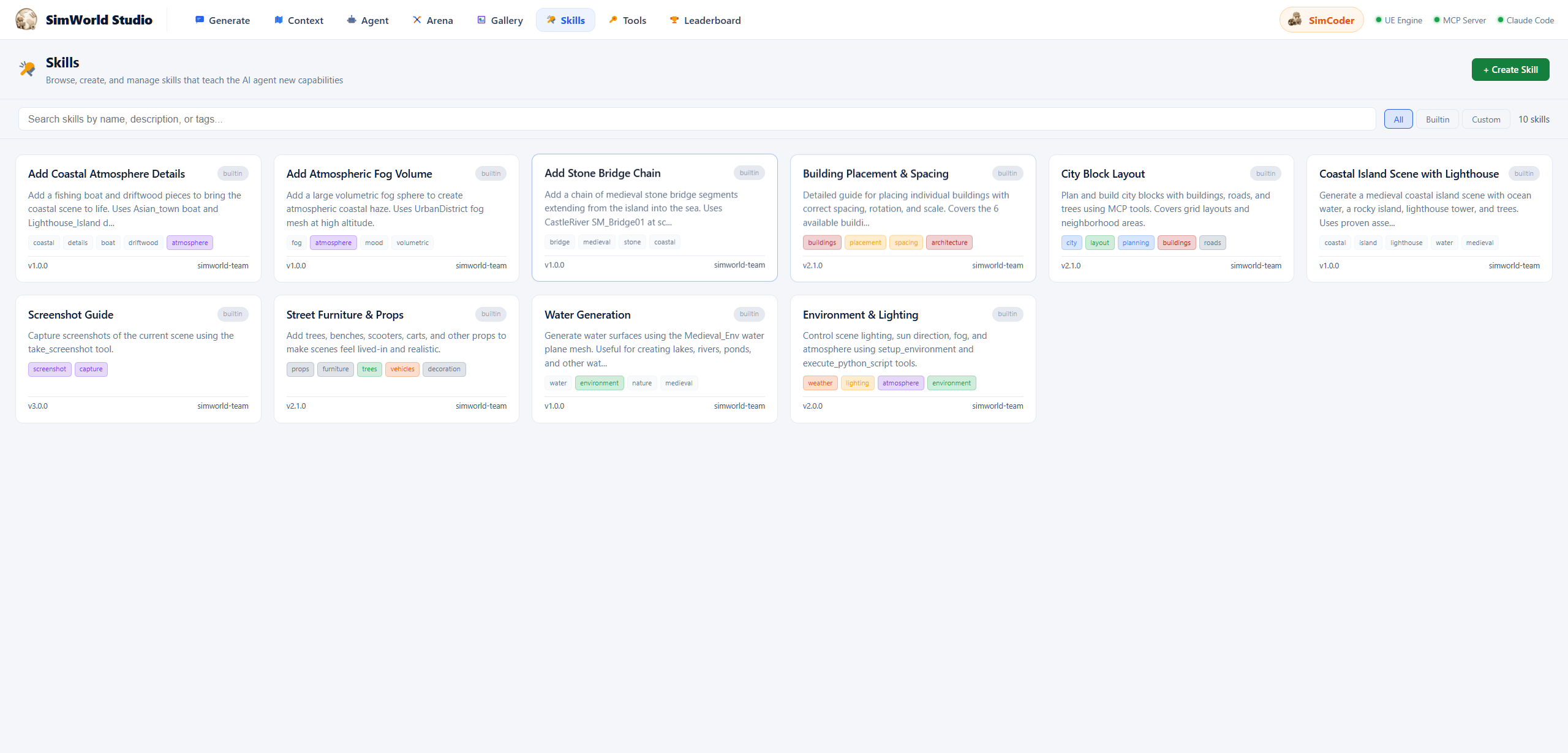}
    \par

    \vspace{1mm}
    \includegraphics[width=\linewidth]{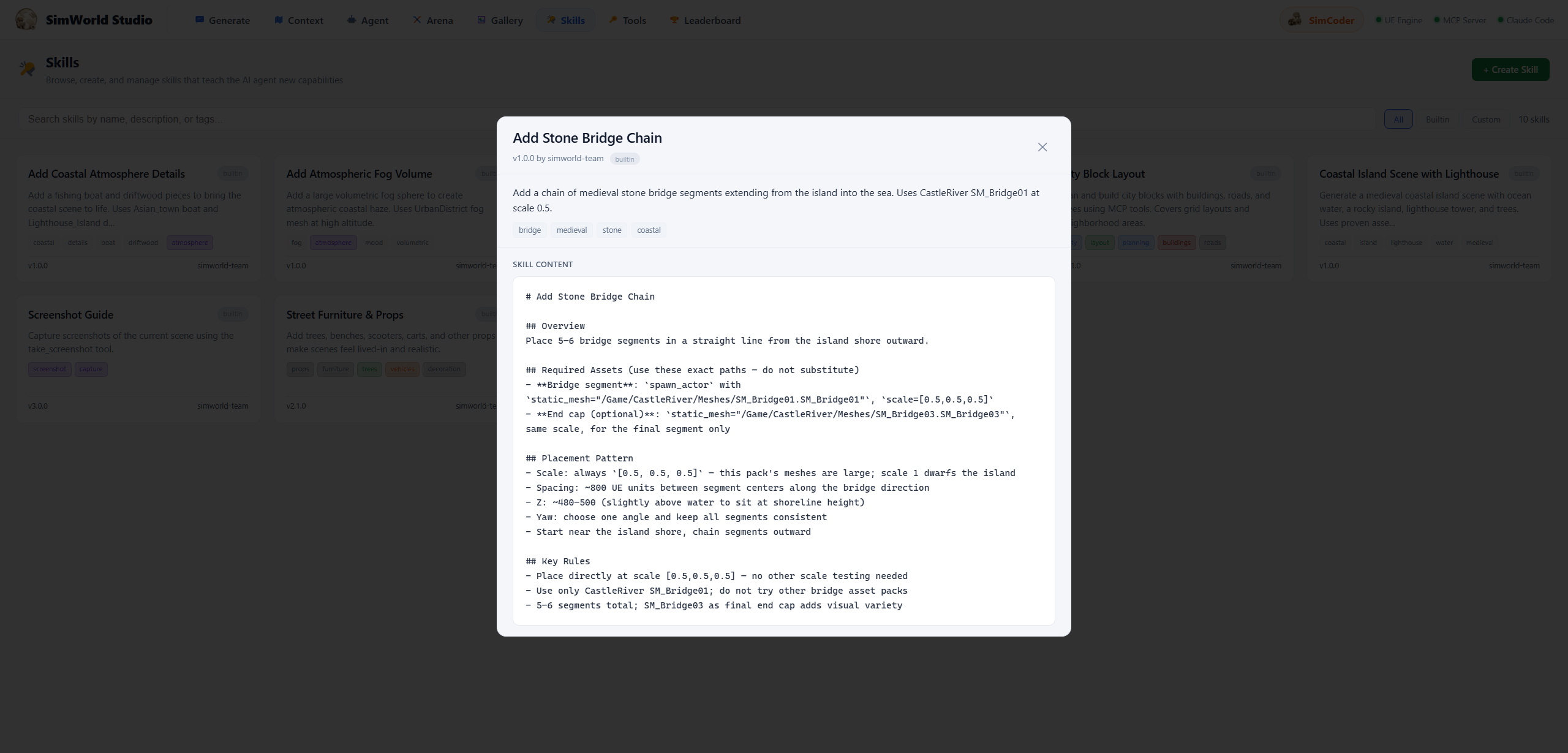}
    \par

    \vspace{1mm}
    \raggedright
    \noindent\textbf{(b) Skills panel.}
    Reusable high-level skills are organized as compositional capabilities that can be browsed, created, and reused by the coding agent across scene-generation episodes.
    The detailed view further exposes the internal structure of skill \texttt{add\_stone\_bridge\_chain}, including its description, callable interface, implementation details, and execution history, enabling the agent to inspect and reuse learned abstractions across tasks.
\end{interfacecasebox}

\Needspace{0.35\textheight}
\begin{interfacecasebox}{Appendix Case C: Tools Panel}
    \vspace{0.5mm}
    \centering
    \includegraphics[width=\linewidth]{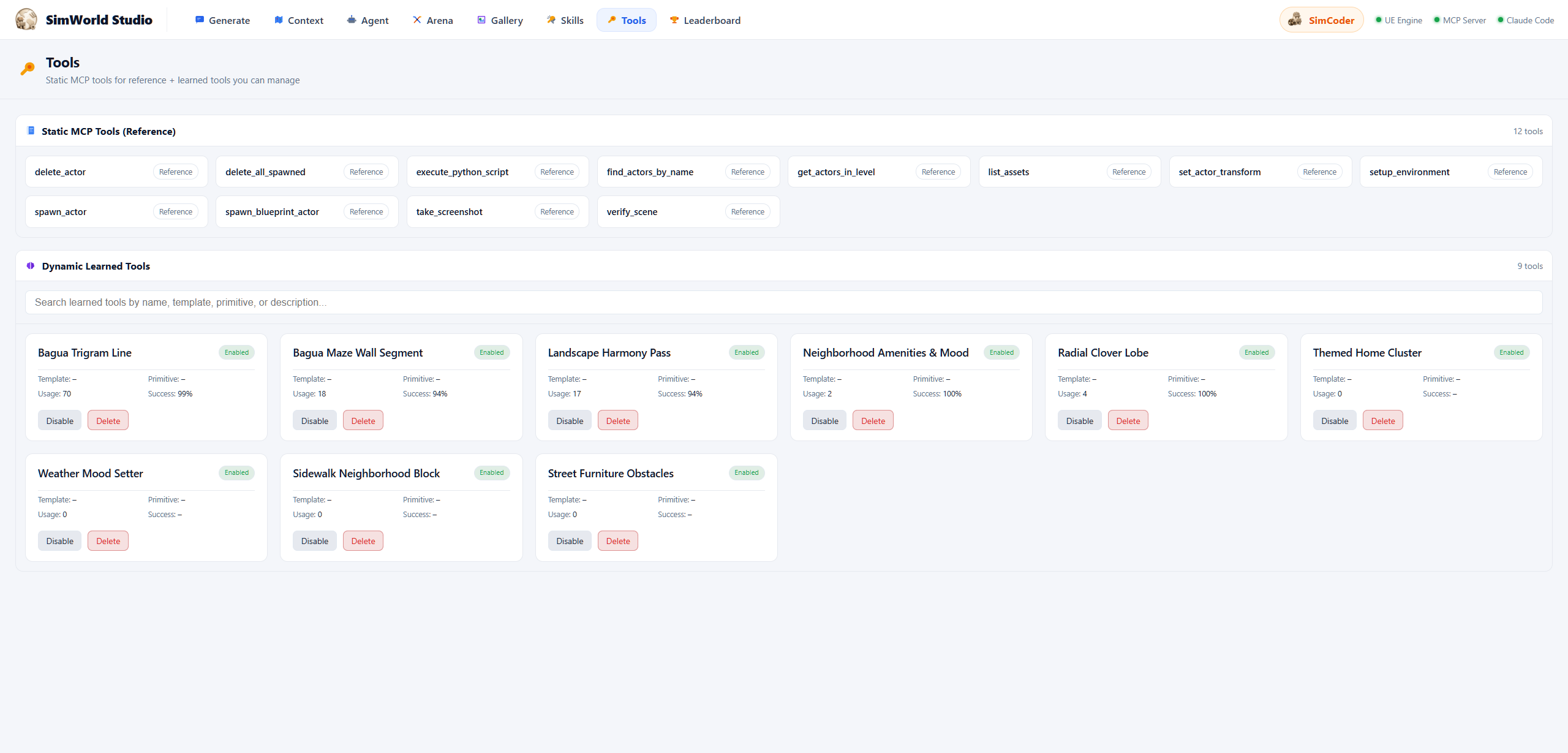}
    \par

    \vspace{1mm}
    \raggedright
    \noindent\textbf{(c) Tools panel.}
    The interface exposes both static MCP tools and dynamically learned tools, enabling the agent to combine primitive operations with reusable abstractions.
\end{interfacecasebox}

\Needspace{0.35\textheight}
\begin{interfacecasebox}{Appendix Case D: Interactive User-Control Mode}
    \vspace{0.5mm}
    \centering
    \includegraphics[width=\linewidth]{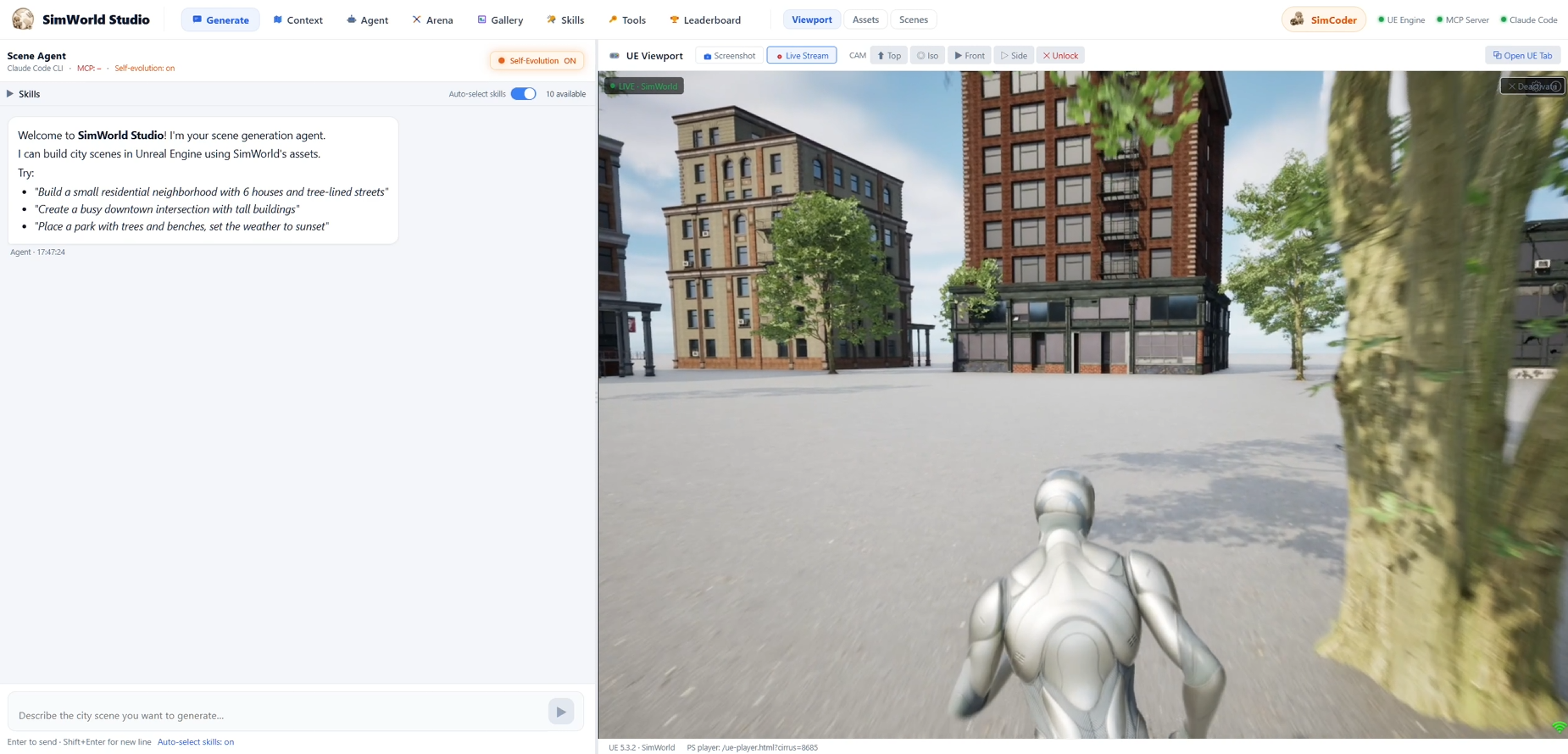}
    \par

    \vspace{1mm}
    \raggedright
    \noindent\textbf{(d) Interactive user-control mode.}
    Beyond text-based prompting, users can directly control an embodied character with keyboard input to explore, inspect, and test the generated Unreal Engine environment from an egocentric perspective.
\end{interfacecasebox}

\Needspace{7\baselineskip}
\vspace{1mm}
\captionof{figure}{
\textbf{Representative interface views of \textsc{SimWorld Studio}.}
The light-theme main interface provides an integrated workspace for user--agent interaction, UE scene rendering, asset/backend management, Gym environment APIs, and embodied-agent monitoring.
The dark-theme panels further show specialized views for skill management, tool abstraction, and direct embodied interaction, allowing users to move beyond text-only prompting and interact with generated environments through controllable agents.
}
\label{fig:app_studio_ui_all}
\subsection{\ours{} Running Configuration}
\label{app:configuration}
\begin{table}[H]
\centering
\caption{\ours{} Minimal System Requirements.}
\label{tab:system_requirements}
\small
\begin{tabular}{ll}
\toprule
\textbf{Component} & \textbf{Requirement} \\
\midrule
OS & Linux; Ubuntu 20.04+ recommended \\
GPU & NVIDIA GPU with 8GB+ VRAM; tested on L40S, T4, and A100 \\
NVIDIA Driver & Version 525+ with Vulkan support \\
Node.js & Version 18+ \\
Python & Version 3.9+ \\
Disk Space & 40 GB free; 15 GB download + 21 GB extracted \\
\bottomrule
\end{tabular}
\end{table}

\subsection{Assets, Licenses, and Model Access}
\label{app:assets_license}

\begin{table}[h]
\centering
\small
\caption{Existing software, model APIs, and assets used in \ours{}.}
\label{tab:assets_license}
\begin{tabular}{p{0.21\linewidth}p{0.23\linewidth}p{0.20\linewidth}p{0.25\linewidth}}
\toprule
\textbf{Asset / Software} & \textbf{Source / Owner} & \textbf{Version / Access} & \textbf{License / Terms} \\
\midrule
Unreal Engine & Epic Games & UE5.3.2 & Unreal Engine EULA \\
SimWorld asset library & SimWorld\citep{ye2025simworld} & Project asset library & Follows the original SimWorld asset license / terms \\
Qwen3.5-2B/9B/27B & Qwen Team & Hugging Face model cards & Apache-2.0 / model card terms \\
Claude Opus / Sonnet APIs & Anthropic & API model snapshots used in experiments & Anthropic API terms and model documentation \\
Claude Code & Anthropic & Agentic coding framework & Anthropic product / API terms \\
Gymnasium and Python packages & Upstream open-source projects & Listed in repository environment file & Respective open-source licenses \\
\bottomrule
\end{tabular}
\end{table}

We do not scrape personal data or release human-subject data. Generated scenes are created from prompts and licensed engine/assets. Users of the released code are expected to comply with the licenses and terms of the underlying engine, model providers, and asset libraries.

%% file: Appendix/SimCoder_RunningCase.tex
\subsection{Detailed Running Case of \textsc{SimCoder}}
\label{app:simcoder_running_case}

This section provides a detailed running case of \textsc{SimCoder} editing a container-yard maze scene through iterative tool use, feedback injection, and skill accumulation.
The task is to build a navigable container maze near \texttt{PlayerStart}.
Across four shown rounds, \textsc{SimCoder} incrementally constructs the maze entrance, extends it into a T-junction, adds a diagonal obstacle, and introduces dead-end formations.
The case illustrates three core mechanisms of \ours{}: external verifier feedback, collision-aware correction, and reusable skill formation.


\newcommand{\skilltag}[1]{%
    \tcbox[
        on line,
        colback=black!5,
        colframe=black!18,
        boxrule=0.35pt,
        arc=1mm,
        left=0.6mm,
        right=0.6mm,
        top=0.25mm,
        bottom=0.25mm,
        nobeforeafter
    ]{\scriptsize\ttfamily\detokenize{#1}}%
}

\newcommand{\objectstabletitle}{%
    \par\vspace{1mm}
    \noindent\textbf{Objects spawned.}
    \par\vspace{0.5mm}
}

\newtcolorbox{runcasebox}[1]{
    enhanced,
    breakable,
    width=\textwidth,
    colback=black!2,
    colframe=black!18,
    boxrule=0.6pt,
    arc=1.5mm,
    left=1.5mm,
    right=1.5mm,
    top=1.5mm,
    bottom=1.5mm,
    before skip=6pt,
    after skip=6pt,
    title=#1,
    coltitle=black,
    colbacktitle=black!8,
    fonttitle=\small\bfseries
}

\newtcolorbox{roundbox}[1]{
    enhanced,
    breakable,
    width=\textwidth,
    colback=black!1,
    colframe=black!22,
    boxrule=0.6pt,
    arc=1.5mm,
    left=1.5mm,
    right=1.5mm,
    top=1.5mm,
    bottom=1.5mm,
    before skip=7pt,
    after skip=7pt,
    title=#1,
    coltitle=black,
    colbacktitle=black!10,
    fonttitle=\small\bfseries
}

\newtcolorbox{feedbackbox}[1]{
    enhanced,
    breakable,
    width=\linewidth,
    colback=blue!2,
    colframe=blue!18,
    boxrule=0.5pt,
    arc=1.2mm,
    left=1.2mm,
    right=1.2mm,
    top=1.0mm,
    bottom=1.0mm,
    before skip=4pt,
    after skip=4pt,
    title=#1,
    coltitle=black,
    colbacktitle=blue!7,
    fonttitle=\scriptsize\bfseries
}

\newtcolorbox{skillbox}[1]{
    enhanced,
    breakable,
    width=\linewidth,
    colback=green!2,
    colframe=green!20!black!25,
    boxrule=0.5pt,
    arc=1.2mm,
    left=1.2mm,
    right=1.2mm,
    top=1.0mm,
    bottom=1.0mm,
    before skip=4pt,
    after skip=4pt,
    title=#1,
    coltitle=black,
    colbacktitle=green!8,
    fonttitle=\scriptsize\bfseries
}

\lstdefinestyle{pythoncasestyle}{
    language=Python,
    basicstyle=\ttfamily\scriptsize,
    keywordstyle=\bfseries,
    commentstyle=\itshape,
    stringstyle=\ttfamily,
    showstringspaces=false,
    breaklines=true,
    breakatwhitespace=false,
    columns=fullflexible,
    keepspaces=true,
    frame=single,
    rulecolor=\color{black!18},
    backgroundcolor=\color{black!3},
    xleftmargin=1mm,
    xrightmargin=1mm,
    framexleftmargin=1mm,
    framexrightmargin=1mm,
    aboveskip=4pt,
    belowskip=6pt
}

\newcommand{\pythonfilebox}[2]{%
    \begin{tcolorbox}[
        enhanced,
        breakable,
        width=\textwidth,
        colback=black!1,
        colframe=black!18,
        boxrule=0.5pt,
        arc=1.2mm,
        left=1mm,
        right=1mm,
        top=1mm,
        bottom=1mm,
        before skip=5pt,
        after skip=6pt,
        title=#1,
        coltitle=black,
        colbacktitle=black!8,
        fonttitle=\scriptsize\bfseries
    ]
    \lstinputlisting[style=pythoncasestyle]{#2}
    \end{tcolorbox}
}


\begin{runcasebox}{Running Case Overview: Container-Yard Maze Editing}
\textbf{Run ID.} \skilltag{20260505_013558}

\textbf{Task.} Build a navigable container-yard maze near \skilltag{PlayerStart} using iterative scene editing.

\textbf{Model.} Claude Sonnet 4.5 via Claude Code SDK.

\textbf{Rounds shown.} 4.

\textbf{Skills learned in full run.} 6.

\textbf{Final recorded VLM score.} 6.5/10 at Round 3.

\vspace{1mm}
Each round follows a fixed loop. The agent receives scene context and feedback from the previous round, plans the next spatial edit, invokes scene-editing tools, and emits newly learned skills when a reusable pattern is discovered.
After the agent round ends, the pipeline captures screenshots, runs AABB collision checking, performs VLM evaluation in a separate SDK session, and injects the resulting feedback into the next prompt.
\end{runcasebox}

\begin{runcasebox}{Pipeline Loop}
\small
\setlength{\tabcolsep}{4pt}
\renewcommand{\arraystretch}{1.15}
\begin{tabular}{@{}p{0.42\linewidth}p{0.53\linewidth}@{}}
\toprule
\textbf{Agent side: Claude SDK} & \textbf{Pipeline side: subprocess} \\
\midrule
1. Call \skilltag{get_scene_summary}. & Inject VLM and collision feedback from the previous round into the prompt. \\
2. Apply prior VLM/collision fixes. & After the round ends, call \skilltag{take_screenshot}. \\
3. Plan the next maze element. & Run AABB collision checking over spawned actors. \\
4. Call \skilltag{spawn_actor} two to three times. & Run VLM evaluation in a separate SDK session. \\
5. Output JSON with \skilltag{new_skills}. & Inject screenshot, collision, and VLM results into the next round. \\
\bottomrule
\end{tabular}
\end{runcasebox}

\begin{runcasebox}{Skill Library: Pre-seeded and Learned Skills}
\small
\setlength{\tabcolsep}{3pt}
\renewcommand{\arraystretch}{1.15}
\begin{tabular}{@{}p{0.19\linewidth}p{0.34\linewidth}p{0.42\linewidth}@{}}
\toprule
\textbf{Source} & \textbf{Skill} & \textbf{Description} \\
\midrule
Pre-seeded & \skilltag{get_scene_summary} & Get \skilltag{PlayerStart} location and compact summaries of \skilltag{Demo_} actors. \\
Pre-seeded & \skilltag{search_ue_assets} & Search the UE asset registry by keyword for mesh paths. \\
Pre-seeded & \skilltag{spawn_container_row} & Spawn a row of containers with configurable spacing and yaw. \\
Pre-seeded & \skilltag{collision_check_demo_actors} & Run AABB collision checking over actors with the \skilltag{Demo_} prefix. \\
\midrule
Round 1 learned & \skilltag{spawn_maze_l_junction} & L-shaped junction with two parallel walls and one perpendicular blocker. \\
Round 2 learned & \skilltag{spawn_perpendicular_blocker} & Container at 90$^\circ$ yaw to close a corridor end. \\
Round 3 learned & \skilltag{spawn_tjunction_maze_segment} & T-junction with parallel corridors and a diagonal obstacle. \\
Later learned & \skilltag{spawn_maze_t_junction} & T-junction using perpendicular and lengthwise containers. \\
Later learned & \skilltag{spawn_maze_dead_end} & U-shaped dead-end with one back wall and two side walls. \\
Later learned & \skilltag{spawn_dead_end_u_formation} & Three-container U dead-end with configurable offsets. \\
\bottomrule
\end{tabular}
\normalsize
\end{runcasebox}


\begin{roundbox}{Round 1: Maze Entrance Corridor}
\begin{center}
    \includegraphics[width=0.995\linewidth]{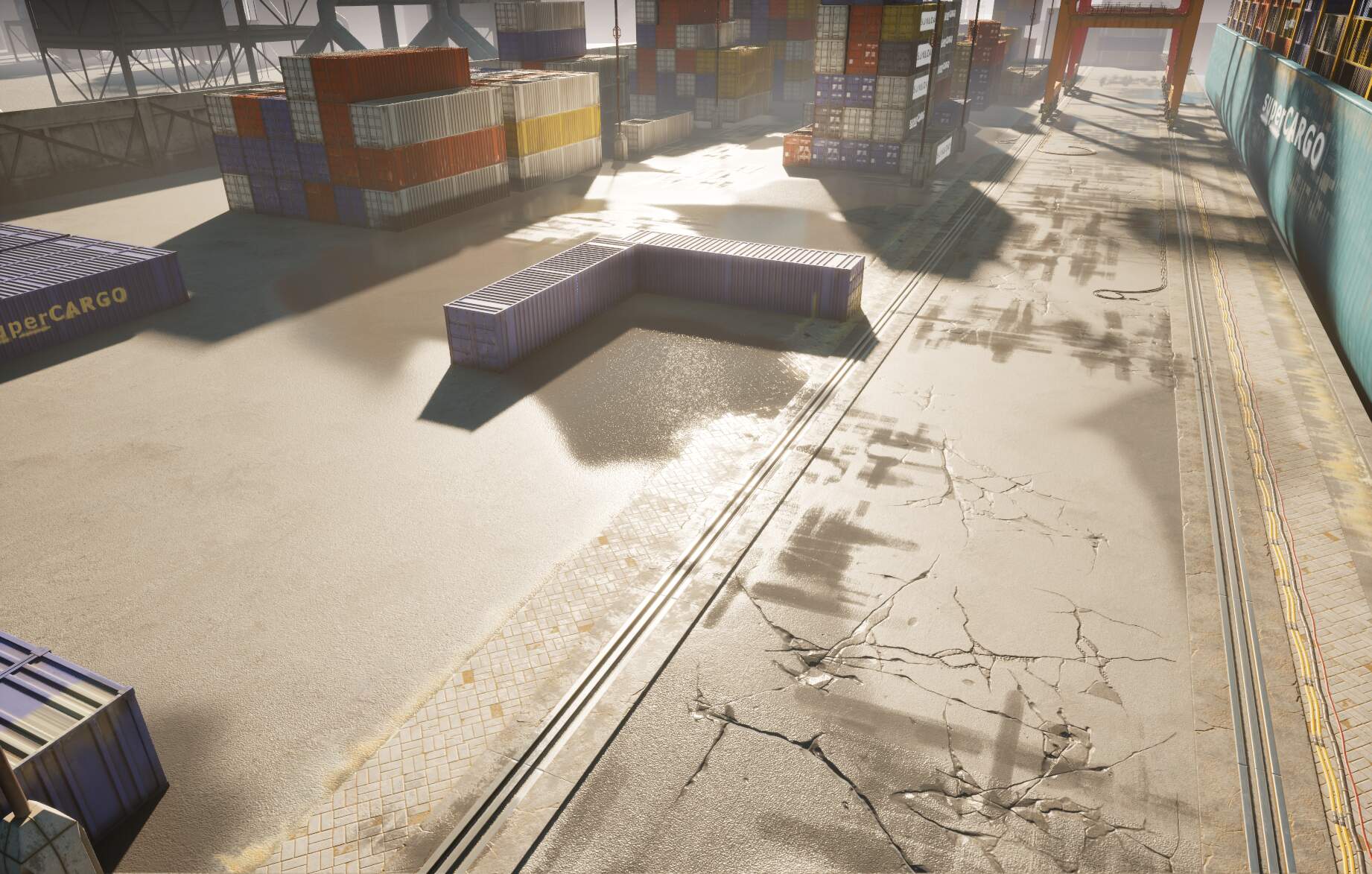}
\end{center}
\vspace{-1mm}

\begin{feedbackbox}{Agent reasoning}
\small
``I will place two parallel lengthwise containers as corridor walls and one perpendicular container as the first turn blocker.
Using rotation=[0,0,0] for walls and rotation=[0,0,90] for the turn---this establishes the maze entrance.''
\normalsize
\end{feedbackbox}

\objectstabletitle
\small
\setlength{\tabcolsep}{3pt}
\renewcommand{\arraystretch}{1.12}
\begin{tabular}{@{}p{0.34\linewidth}p{0.25\linewidth}p{0.10\linewidth}p{0.25\linewidth}@{}}
\toprule
\textbf{Actor name} & \textbf{Mesh} & \textbf{Yaw} & \textbf{Purpose} \\
\midrule
\skilltag{Demo_MazeWall_Left_XXXX}  & \skilltag{SM_Container12m} & 0$^\circ$  & Left corridor wall. \\
\skilltag{Demo_MazeWall_Right_XXXX} & \skilltag{SM_Container12m} & 0$^\circ$  & Right corridor wall. \\
\skilltag{Demo_MazeTurn_Block_XXXX} & \skilltag{SM_Container12m} & 90$^\circ$ & First turn blocker. \\
\bottomrule
\end{tabular}
\normalsize

\begin{skillbox}{Skill saved}
\skilltag{spawn_maze_l_junction}: spawn an L-shaped junction with two parallel walls and one perpendicular blocker.
\end{skillbox}

\begin{feedbackbox}{VLM verifier feedback}
\small
\textbf{Score: 4.5/10.}
The verifier finds a good foundation with rotation variety and industrial realism, but the maze is still too sparse.
Only three containers are visible, so the structure reads as a simple corridor rather than a recognizable maze.
The next round should add branching paths and small navigation obstacles.
\normalsize
\end{feedbackbox}

\textbf{Collision check.}
Not triggered in Round 1 because there are no prior maze actors to conflict with.
\end{roundbox}

\pythonfilebox{Saved Python Utility: \texttt{spawn\_maze\_l\_junction}}{Appendix/Script/spawn_maze_l_junction.py}


\begin{roundbox}{Round 2: T-Junction Extension}
\begin{center}
    \includegraphics[width=0.995\linewidth]{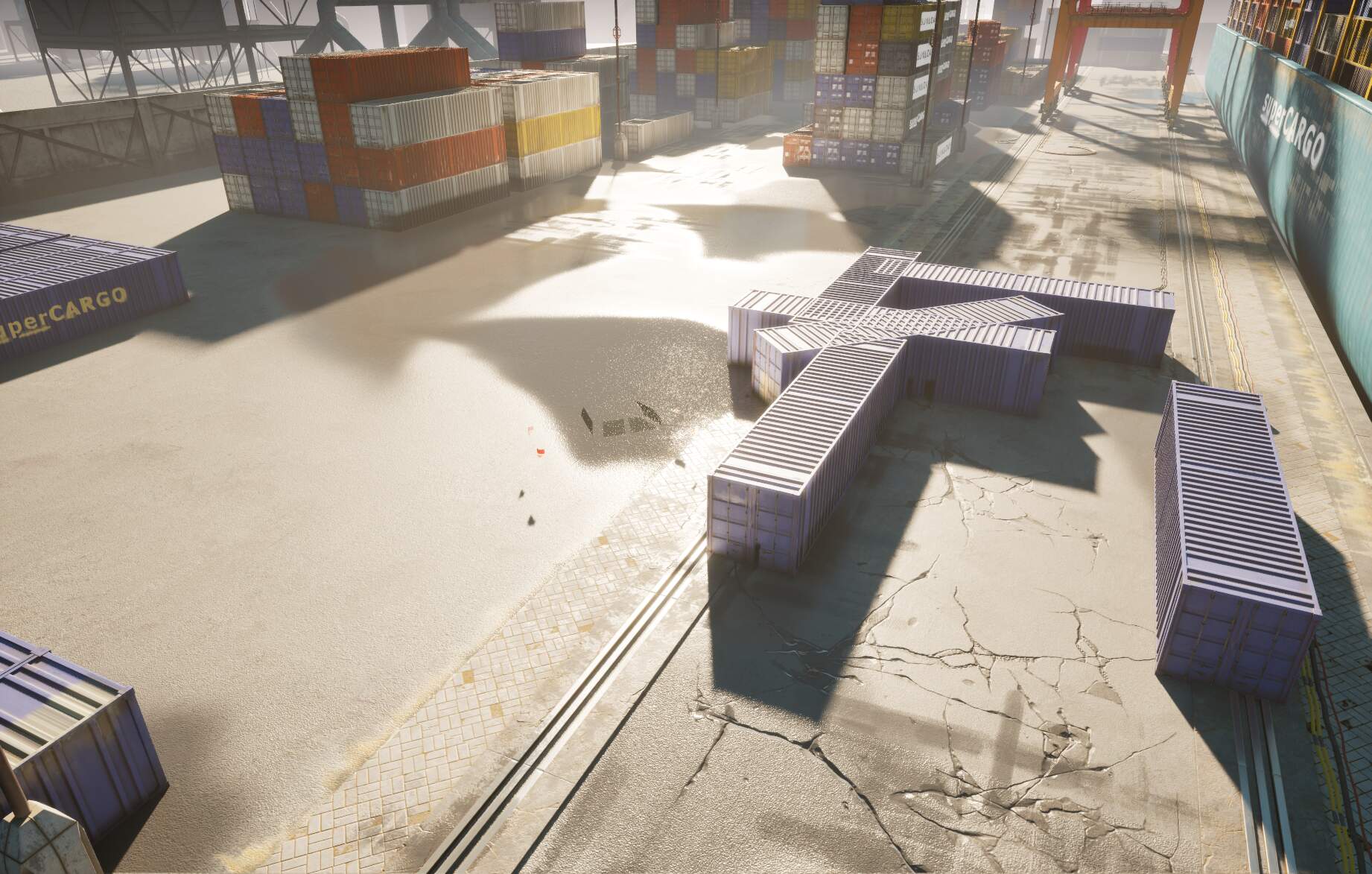}
\end{center}
\vspace{-1mm}

\begin{feedbackbox}{injected context}
\small
\textbf{VLM Score: 4.5/10.}
Add branching paths, such as a T-junction or dead-end fork, and include small navigation obstacles.
\normalsize
\end{feedbackbox}

\begin{feedbackbox}{Agent reasoning}
\small
``VLM says density is too low.
I will extend the corridor into a T-junction by adding two more lengthwise walls parallel to the existing corridor, plus a perpendicular blocker at the junction to force the turn decision.''
\normalsize
\end{feedbackbox}

\objectstabletitle
\small
\setlength{\tabcolsep}{3pt}
\renewcommand{\arraystretch}{1.12}
\begin{tabular}{@{}p{0.34\linewidth}p{0.25\linewidth}p{0.10\linewidth}p{0.25\linewidth}@{}}
\toprule
\textbf{Actor name} & \textbf{Mesh} & \textbf{Yaw} & \textbf{Purpose} \\
\midrule
\skilltag{Demo_WallWest_XXXX}   & \skilltag{SM_Container12m} & 0$^\circ$  & West branch wall. \\
\skilltag{Demo_WallEast_XXXX}   & \skilltag{SM_Container12m} & 0$^\circ$  & East branch wall. \\
\skilltag{Demo_TjuncBlock_XXXX} & \skilltag{SM_Container12m} & 90$^\circ$ & T-junction blocker. \\
\skilltag{Demo_TjuncBlock_XXXX} & \skilltag{SM_Container12m} & 90$^\circ$ & Retry spawn to confirm junction closure. \\
\bottomrule
\end{tabular}
\normalsize

\begin{skillbox}{Skill saved}
\skilltag{spawn_perpendicular_blocker}: spawn a container at 90$^\circ$ yaw to close a corridor end or force a turn.
\end{skillbox}

\begin{feedbackbox}{VLM verifier feedback}
\small
\textbf{Score: 5.5/10.}
The verifier confirms visible rotation variety and a clearer L-turn geometry.
However, the scene is still closer to a simple hallway than a branching maze.
The next round should add a U-shaped dead end, close the north gap with a perpendicular blocker, and add small props inside corridors.
\normalsize
\end{feedbackbox}

\textbf{Collision check.}
One collision pair is detected between \skilltag{Demo_MazeWall_01} and \skilltag{Demo_EntranceWall}.
The agent fixes this by deleting the overlapping actor before continuing.
\end{roundbox}


\begin{roundbox}{Round 3: Diagonal Obstacle and North Extension}
\begin{center}
    \includegraphics[width=0.995\linewidth]{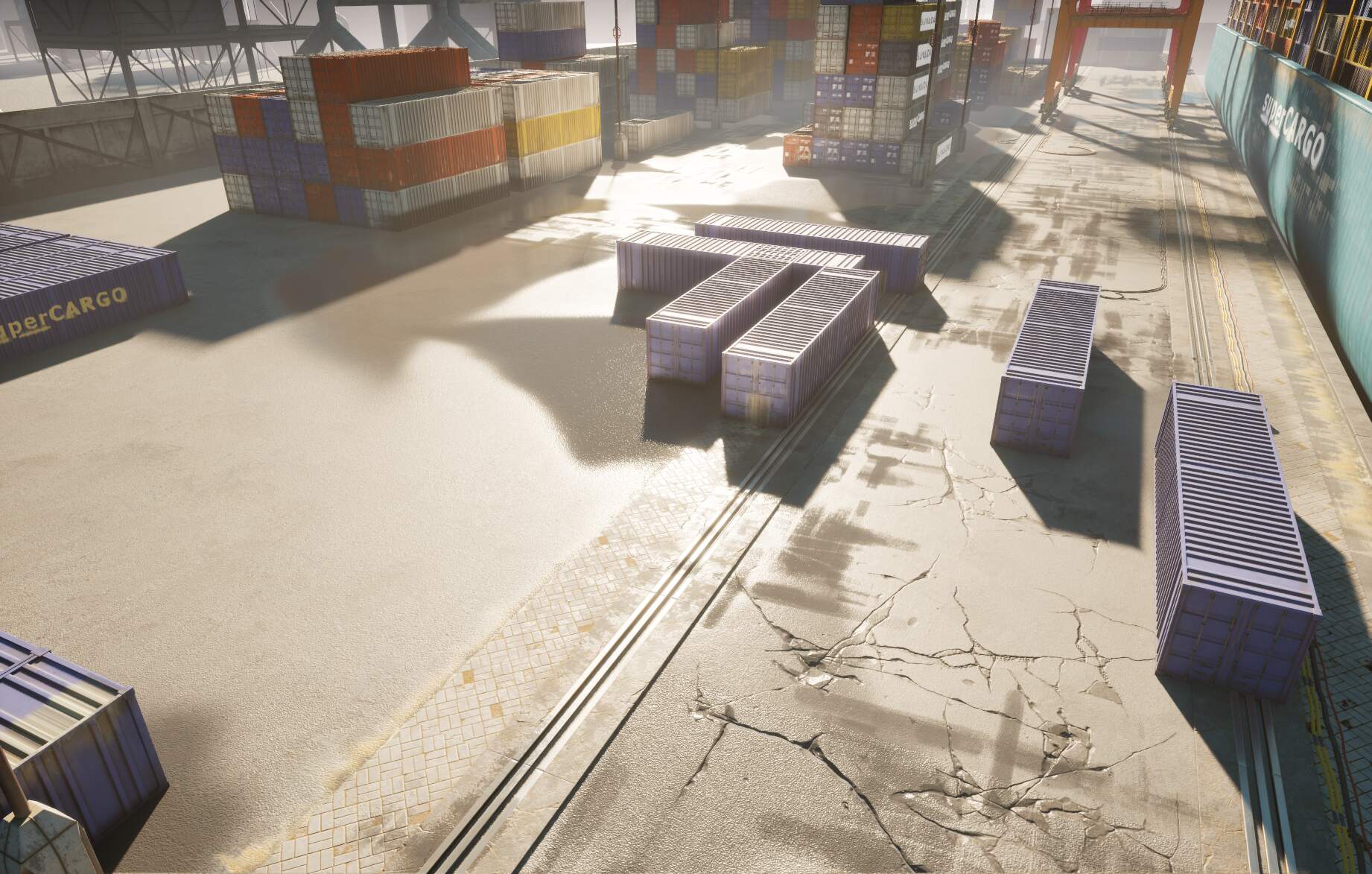}
\end{center}
\vspace{-1mm}

\begin{feedbackbox}{injected context}
\small
\textbf{VLM Score: 5.5/10.}
Add a U-shaped dead end, close the north gap with a perpendicular blocker, and spawn two to three small props inside corridors.
\normalsize
\end{feedbackbox}

\begin{feedbackbox}{Agent reasoning}
\small
``I will extend north with two more parallel walls, add a 45-degree diagonal container as a navigation obstacle inside the corridor to break straight-line visibility, saving as a T-junction skill for future reuse.''
\normalsize
\end{feedbackbox}

\objectstabletitle
\small
\setlength{\tabcolsep}{3pt}
\renewcommand{\arraystretch}{1.12}
\begin{tabular}{@{}p{0.34\linewidth}p{0.25\linewidth}p{0.10\linewidth}p{0.25\linewidth}@{}}
\toprule
\textbf{Actor name} & \textbf{Mesh} & \textbf{Yaw} & \textbf{Purpose} \\
\midrule
\skilltag{Demo_NorthWallWest_XXXX} & \skilltag{SM_Container12m} & 0$^\circ$  & North corridor wall. \\
\skilltag{Demo_NorthWallEast_XXXX} & \skilltag{SM_Container12m} & 0$^\circ$  & North corridor wall. \\
\skilltag{Demo_DiagObstacle_XXXX}  & \skilltag{SM_Container12m} & 45$^\circ$ & Diagonal obstacle inside corridor. \\
\bottomrule
\end{tabular}
\normalsize

\begin{skillbox}{Skill saved}
\skilltag{spawn_tjunction_maze_segment}: spawn a T-junction maze segment with parallel corridors and a diagonal obstacle.
\end{skillbox}

\begin{feedbackbox}{VLM verifier feedback}
\small
\textbf{Score: 6.5/10.}
The verifier confirms rotation diversity through diagonal container placement and a wider branching footprint.
Navigable aisles remain visible and the industrial context is preserved.
The scene is still below full maze complexity, so future rounds should add more containers, dead-end paths, and small interior props.
\normalsize
\end{feedbackbox}

\textbf{Collision check.}
No collisions are detected.
\end{roundbox}


\begin{roundbox}{Round 4: Dead-End Formations}
\begin{center}
    \includegraphics[width=0.995\linewidth]{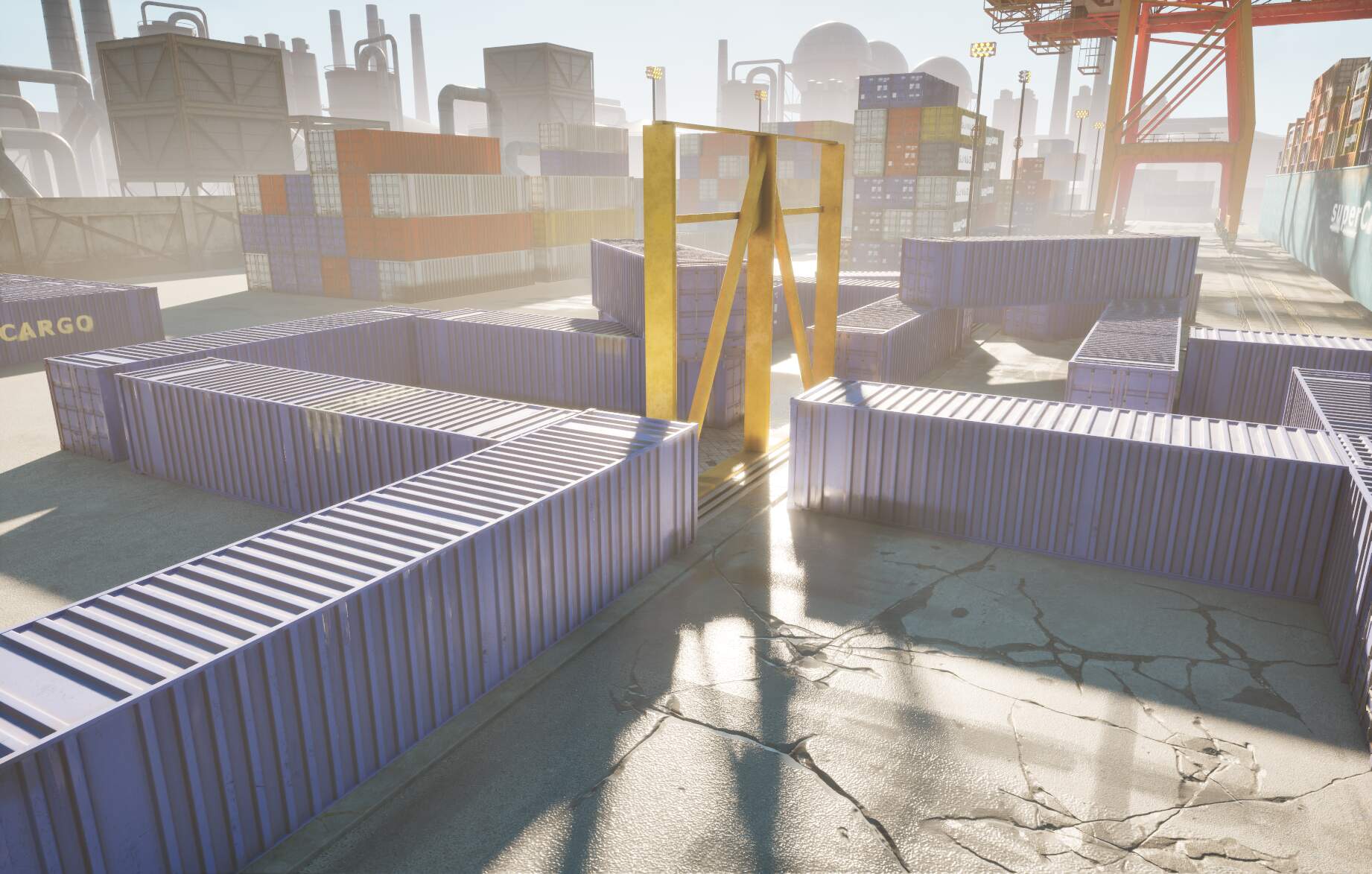}
\end{center}
\vspace{-1mm}

\begin{feedbackbox}{Injected context}
\small
\textbf{VLM Score: 6.5/10.}
Add U-shaped dead-end formations and close open gaps.
\normalsize
\end{feedbackbox}

\begin{feedbackbox}{Agent reasoning}
\small
``I will build two U-shaped dead-ends to force the agent into backtracking decisions.
This round uses six spawns, which is above the usual limit, but it is useful for increasing maze complexity.''
\normalsize
\end{feedbackbox}

\objectstabletitle
Six containers are spawned to form two U-shaped dead-end structures.
The formation uses lengthwise walls at 0$^\circ$ yaw and perpendicular blockers at 90$^\circ$ yaw.

\begin{feedbackbox}{VLM verifier feedback}
\small
Round 4 VLM feedback is not recorded because of a pipeline screenshot path error.
\normalsize
\end{feedbackbox}

\textbf{Collision check.}
No collisions are detected.
\end{roundbox}


\begin{runcasebox}{Score Progression}
\small
\setlength{\tabcolsep}{3pt}
\renewcommand{\arraystretch}{1.12}
\begin{tabular}{@{}p{0.10\linewidth}p{0.16\linewidth}p{0.16\linewidth}p{0.18\linewidth}p{0.34\linewidth}@{}}
\toprule
\textbf{Round} & \textbf{VLM Score} & \textbf{Collisions} & \textbf{Skills Learned} & \textbf{Key Action} \\
\midrule
1 & 4.5/10 & 0 & 1 & Entrance corridor with L-shaped turn. \\
2 & 5.5/10 & 1 $\rightarrow$ 0 & 1 & T-junction extension and collision repair. \\
3 & 6.5/10 & 0 & 1 & North extension with 45$^\circ$ diagonal obstacle. \\
4 & --- & 0 & 0 & Dead-end formations for backtracking behavior. \\
\bottomrule
\end{tabular}
\normalsize
\end{runcasebox}

\begin{runcasebox}{Evidence of Pipeline Components}
\textbf{Skill reuse.}
Later rounds explicitly retrieve and reuse earlier skills such as \skilltag{spawn_perpendicular_blocker} and \skilltag{spawn_maze_l_junction}, showing that learned scene-editing patterns are persisted across rounds.

\vspace{1mm}
\textbf{VLM-to-agent feedback loop.}
The VLM feedback is injected into the next prompt as hard context.
For example, after Round 1, the agent receives a score of 4.5/10 and the instruction to add branching paths.
The next response directly acts on this feedback by extending the corridor into a T-junction.

\vspace{1mm}
\textbf{Collision-aware auto-fix.}
When the pipeline detects a collision between \skilltag{Demo_MazeWall_01} and \skilltag{Demo_EntranceWall}, the next prompt requires the agent to delete or move the overlapping actor before adding new objects.
This turns geometric validation into an explicit correction signal.

\vspace{1mm}
\textbf{Rotation diversity.}
The run gradually increases geometric complexity by using 0$^\circ$, 45$^\circ$, 90$^\circ$, and later 135$^\circ$ yaw placements.
This produces corridors, perpendicular blockers, diagonal obstacles, and dead-end formations.
\end{runcasebox}

\begin{runcasebox}{Level Collision Checking}
The collision checker runs after every round over actors with the \skilltag{Demo_} prefix.
It performs a lightweight AABB overlap test and injects detected collisions into the next prompt as mandatory correction context.
\end{runcasebox}

\pythonfilebox{Level AABB Collision Check}{Appendix/Script/collision_check.py}

\noindent
The resulting message is injected into the next round as mandatory correction context:
\begin{quote}
\small\ttfamily
COLLISIONS: N pair(s) detected -> DELETE or MOVE before spawning.
\end{quote}

%% file: Appendix/Additional-Related-Work.tex
\newpage
\section{Additional Related Work}
\label{app:relatedwork}

\paragraph{3D scene and world generation.}
\label{sec:related-scenegen}

\textit{Generative 3D methods} synthesize explicit 3D scene representations such as object layouts, meshes, or 3D Gaussian splats. Data-driven indoor methods model object arrangements through autoregressive, diffusion-based, or scene-graph-conditioned generation~\citep{paschalidou2021atiss, tang2024diffuscene, zhai2023commonscenesgeneratingcommonsense3d, bokhovkin2025scenefactor}, while 2D-prior-based methods construct scenes through depth estimation, inpainting, and multi-view fusion~\citep{hollein2023text2room, chung2023luciddreamerdomainfreegeneration3d, yu2025wonderworldinteractive3dscene, zhang2023scenewiz3d}.
\textit{Video-based world models} instead treat world generation as conditioned video synthesis, producing controllable visual experiences without exposing editable 3D assets, persistent scene graphs, or physics-simulator state~\citep{bruce2024genie, yang2023learning, nvidia2025cosmosworldfoundationmodel, hong2025relic}.
\textit{Procedural and agentic pipelines} generate simulation-ready environments through procedural generators or LLM/VLM-driven agents that combine layout planning, asset retrieval or synthesis, placement, and iterative validation~\citep{deitke2022procthorlargescaleembodiedai, raistrick2023infinite, yang2024holodecklanguageguidedgeneration, hu2024scenecraftllmagentsynthesizing, xia2026sagescalableagentic3d, pfaff2026scenesmith, kuang2026vulcantoolaugmentedmultiagents, zhang2026code2worlds, liu2026imaginecitycitygenagentprocedural}.
 SimWorld Studio targets high-fidelity open-ended world generation in Unreal Engine~5, provides a standardized gym-like interface for embodied-agent training, and incorporates a persistent verifier-driven skill library for continual improvement of the coding agent.

\paragraph{Tool-augmented coding agents.}
\label{sec:related-agent}

Large language models have evolved from passive code generators into agents that reason, invoke tools, execute code, and interact with external environments. 
Recent work has developed this paradigm through interleaved reasoning and acting, large-scale API invocation and orchestration, executable code actions, and reflective feedback~\citep{yao2022react, schick2023toolformerlanguagemodelsteach, qin2023toolllmfacilitatinglargelanguage, wang2024executable, qian2023creatortoollibcreationllmagent, cai2024largelanguagemodelstool, shinn2023reflexion}. 
Evaluation has shifted from function-level code synthesis~\citep{chen2021evaluating, austin2021program} to compositional library use~\citep{zhuo2024bigcodebench}, repository-level patch generation with executable test harnesses~\citep{jimenez2023swe, zan2025multi, yang2024swemultimodal}, and holistic agent evaluation across operating systems and interactive environments~\citep{liu2023agentbench, xie2024osworld}.
Another line develops reusable agent infrastructure, including executable skill libraries, generated tools, role-specialized agents, agent--computer interfaces, sandboxed harnesses, and multi-agent coordination frameworks for reproducible verifier-guided iteration~\citep{wang2023voyageropenendedembodiedagent, cai2024largelanguagemodelstool, wang2024trove, hong2023metagpt, wu2024autogen, yang2024swe, wang2024openhands, xi2025agentgym}. 
SimWorld Studio extends tool-augmented coding agents beyond software repair and digital automation to verified game-engine construction of simulation-ready embodied environments.

\paragraph{Self-evolving agents.}
\label{sec:related-selfevolve}

Recent work on self-evolving agents studies closed-loop systems that autonomously improve from environmental feedback, self-generated curricula, and agent--environment co-evolution~\citep{gao2025survey, fang2025comprehensive}.
In language and software domains, key mechanisms include verbal self-reflection, experience accumulation, self-editing code, and automated search over agent designs~\citep{shinn2023reflexion, madaan2023self, zhao2024expel, yin2024g, hu2024automated, robeyns2025self, zhang2025agentic, zhang2025darwin}.
In web and general interactive settings, agents improve through failure-driven online curricula, multi-turn reinforcement learning, or iterative co-evolution with world models~\citep{qi2024webrl, he2025openwebvoyager, fang2025webevolver, xi2025agentgym, dong2026agentworldscalingrealworldenvironment}.
In embodied and robotic settings, self-evolution appears as adaptive training loops that expand an agent's skills and training distribution through automatic curricula, executable skill libraries, and generated tasks, rewards, or environments~\citep{wang2023voyageropenendedembodiedagent,faldor2024omniepic,wang2023robogen,ma2023eureka,xie2023text2reward,liang2024eurekaverse, luo2025self}.
SimWorld Studio is distinct in that it self-evolves the 3D world generator itself, guided by executable scene verifiers and downstream embodied-agent usability feedback, rather than primarily improving the agent, policy, or task curriculum.

\paragraph{Agent--environment co-evolution.}
\label{sec:related-coevol}

In unsupervised environment design, environments are generated, selected, or edited near the agent's capability frontier to induce automatic curricula~\citep{dennis2020emergent,jiang2020prioritized,parker2022evolving}, with open-ended and multi-agent variants jointly adapting environments, solvers, or co-players~\citep{wang2019poet,wang2020enhanced,samvelyan2023maestro}.
LLM-based methods move co-evolution into code space, using language models to generate and iteratively revise environment configurations, terrains, rewards, or task programs from agent feedback or learnability signals~~\citep{zala2024envgengeneratingadaptingenvironments,ma2023eureka,faldor2024omniepic,liang2024eurekaverse}; GenEnv~\citep{guo2025genenvdifficultyalignedcoevolutionllm} further co-trains the agent and a generative environment policy through RL difficulty-aligned curriculum.
A parallel world-model line uses learned simulators for VLA and RL policy improvement and iteratively co-refines the world model and policy from imagined rollouts, real interactions, or policy failures~\citep{jiang2026wovr,sharma2026worldgymnasttrainingrobotsreinforcement, liu2026worldvlaloop,guo2026vlaw,gigabrainteam2026gigabrain}.
R-Zero demonstrates an analogous Challenger--Solver loop in which task generation and reasoning ability co-evolve without human-curated data~\citep{huang2025rzeroselfevolvingreasoningllm}.
SimWorld Studio differs by generating simulation-ready 3D UE5 worlds via code and adapting future worlds from downstream embodied-agent feedback.

%% file: Appendix/B-CaseStudy_1.tex
\newpage
\section{Case Study 1}
\label{app:case_study_1}

\subsection{Full Results by Difficulty Level}
\label{app:full_results}

Table~\ref{tab:full_results} reports per-metric performance broken down by difficulty level (easy/medium/hard) for all four LLM backbones and three generation settings.

\definecolor{metphys}{HTML}{1565C0}
\definecolor{metquant}{HTML}{2E7D32}
\definecolor{metsem}{HTML}{E65100}
\definecolor{metaes}{HTML}{AD1457}

\begin{table}[H]
\centering
\caption{\textbf{Scene generation quality across LLM backbones, settings, and difficulty levels.} All metrics $\in [0,1]$ ($\uparrow$). Bold = best per column within each setting. E/M/H = easy/medium/hard; Avg = mean across difficulties. Metric colors: \textcolor{metquant}{quantity}, \textcolor{metphys}{physical validity}, \textcolor{metsem}{semantic}, \textcolor{metaes}{aesthetic}.}
\label{tab:full_results}
\vspace{2pt}
\scriptsize
\setlength{\tabcolsep}{3.5pt}
\renewcommand{\arraystretch}{1.12}
\resizebox{\textwidth}{!}{%
\begin{tabular}{@{}ll cccc cc c cccc cc c cccc cc c@{}}
\toprule
& & \multicolumn{7}{c}{\textbf{S1: Text-to-Scene}} & \multicolumn{7}{c}{\textbf{S2: Image+Text-to-Scene}} & \multicolumn{7}{c}{\textbf{S3: Scene Editing}} \\
\cmidrule(lr){3-9} \cmidrule(lr){10-16} \cmidrule(lr){17-23}
& & \multicolumn{4}{c}{\textit{Rule-Based}} & \multicolumn{2}{c}{\textit{VLM}} &
  & \multicolumn{4}{c}{\textit{Rule-Based}} & \multicolumn{2}{c}{\textit{VLM}} &
  & \multicolumn{4}{c}{\textit{Rule-Based}} & \multicolumn{2}{c}{\textit{VLM}} & \\
\cmidrule(lr){3-6} \cmidrule(lr){7-8} \cmidrule(lr){10-13} \cmidrule(lr){14-15} \cmidrule(lr){17-20} \cmidrule(lr){21-22}
\textbf{LLM} & \textbf{Diff.}
  & \rotatebox{70}{\textcolor{metquant}{Count}} & \rotatebox{70}{\textcolor{metquant}{Diversity}} & \rotatebox{70}{\textcolor{metphys}{No Collision}} & \rotatebox{70}{\textcolor{metphys}{Gravity}}
  & \rotatebox{70}{\textcolor{metsem}{Fidelity}} & \rotatebox{70}{\textcolor{metaes}{Aesthetics}} & \rotatebox{70}{\textbf{Avg}}
  & \rotatebox{70}{\textcolor{metquant}{Count}} & \rotatebox{70}{\textcolor{metphys}{No Collision}} & \rotatebox{70}{\textcolor{metphys}{Gravity}} & \rotatebox{70}{\textcolor{metphys}{In-Bounds}}
  & \rotatebox{70}{\textcolor{metsem}{Fidelity}} & \rotatebox{70}{\textcolor{metsem}{Style}} & \rotatebox{70}{\textbf{Avg}}
  & \rotatebox{70}{\textcolor{metphys}{Preserve}} & \rotatebox{70}{\textcolor{metquant}{Edit Count}} & \rotatebox{70}{\textcolor{metphys}{No Collision}} & \rotatebox{70}{\textcolor{metsem}{Coherence}}
  & \rotatebox{70}{\textcolor{metsem}{Edit Compl.}} & \rotatebox{70}{\textcolor{metaes}{Layout}} & \rotatebox{70}{\textbf{Avg}} \\
\midrule
\multirow{4}{*}{\makecell[l]{Qwen\\9B}}
  & E & .00 & .70 & .60 & .80 & .30 & .20 & .43
    & .80 & .50 & .70 & 1.0 & .20 & .20 & .57
    & 1.0 & .00 & .60 & .30 & .30 & .30 & .42 \\
  & M & .00 & .50 & .30 & .50 & .20 & .10 & .27
    & .60 & .30 & .50 & 1.0 & .10 & .10 & .43
    & 1.0 & .00 & .30 & .23 & .20 & .20 & .32 \\
  & H & .00 & .30 & .20 & .43 & .10 & .10 & .19
    & .40 & .20 & .40 & 1.0 & .10 & .10 & .37
    & 1.0 & .00 & .20 & .17 & .20 & .10 & .28 \\
\rowcolor{rowgray}
  & Avg & .00 & .50 & .37 & .58 & .20 & .13 & .36
    & .60 & .33 & .53 & 1.0 & .13 & .13 & .45
    & 1.0 & .00 & .37 & .23 & .23 & .20 & .34 \\
\midrule
\multirow{4}{*}{\makecell[l]{Qwen\\27B}}
  & E & .50 & .60 & 1.0 & 1.0 & .60 & .50 & .70
    & 1.0 & .95 & 1.0 & 1.0 & .50 & .37 & .80
    & 1.0 & .00 & .99 & .50 & .20 & \textbf{.60} & .55 \\
  & M & .00 & .46 & .99 & 1.0 & .40 & .30 & .53
    & .70 & .80 & .90 & 1.0 & .37 & .30 & .68
    & 1.0 & .00 & .99 & .47 & .10 & .50 & .51 \\
  & H & .00 & .32 & .98 & 1.0 & .30 & .30 & .48
    & .50 & .50 & .60 & 1.0 & .27 & .23 & .52
    & 1.0 & .00 & .99 & .43 & .10 & .40 & .49 \\
\rowcolor{rowgray}
  & Avg & .17 & .46 & .99 & 1.0 & .43 & .37 & .59
    & .73 & .75 & .83 & 1.0 & .38 & .30 & .67
    & 1.0 & .00 & .99 & .47 & .13 & .50 & .52 \\
\midrule
\multirow{4}{*}{Sonnet 4}
  & E & 1.0 & \textbf{1.0} & 1.0 & 1.0 & .70 & .50 & .87
    & 1.0 & \textbf{.99} & \textbf{1.0} & 1.0 & .63 & .47 & .85
    & 1.0 & 1.0 & .99 & .50 & \textbf{.90} & .50 & \textbf{.82} \\
  & M & .00 & \textbf{.80} & \textbf{1.0} & 1.0 & .50 & .30 & .60
    & .80 & \textbf{.85} & .90 & 1.0 & .50 & .37 & .74
    & 1.0 & 1.0 & .98 & .40 & \textbf{.60} & .40 & .73 \\
  & H & .00 & .60 & \textbf{1.0} & 1.0 & \textbf{.40} & \textbf{.30} & .55
    & .60 & .60 & \textbf{.70} & 1.0 & .37 & .27 & .59
    & 1.0 & 1.0 & .97 & .30 & \textbf{.50} & .30 & .68 \\
\rowcolor{rowgray}
  & Avg & .33 & \textbf{.80} & \textbf{1.0} & 1.0 & .53 & .37 & .70
    & .80 & .81 & .87 & 1.0 & .50 & .37 & .73
    & 1.0 & 1.0 & .98 & .40 & \textbf{.67} & .40 & .74 \\
\midrule
\multirow{4}{*}{Opus 4}
  & E & \textbf{1.0} & .80 & 1.0 & 1.0 & \textbf{.80} & \textbf{.60} & \textbf{.87}
    & \textbf{1.0} & .99 & 1.0 & 1.0 & \textbf{.80} & \textbf{.60} & \textbf{.90}
    & 1.0 & 1.0 & .99 & \textbf{.60} & .70 & .60 & .82 \\
  & M & \textbf{1.0} & .72 & .99 & 1.0 & \textbf{.70} & \textbf{.50} & \textbf{.82}
    & \textbf{.80} & .90 & .95 & 1.0 & \textbf{.60} & \textbf{.47} & \textbf{.79}
    & 1.0 & 1.0 & .98 & \textbf{.50} & .50 & \textbf{.53} & \textbf{.75} \\
  & H & \textbf{.25} & \textbf{.64} & .98 & 1.0 & .40 & .30 & \textbf{.60}
    & \textbf{.70} & \textbf{.70} & .80 & 1.0 & \textbf{.47} & \textbf{.33} & \textbf{.67}
    & 1.0 & 1.0 & .97 & \textbf{.40} & .30 & \textbf{.47} & \textbf{.69} \\
\rowcolor{rowgray}
  & Avg & \textbf{.75} & .72 & .99 & 1.0 & \textbf{.63} & \textbf{.47} & \textbf{.77}
    & \textbf{.83} & \textbf{.86} & \textbf{.92} & 1.0 & \textbf{.62} & \textbf{.47} & \textbf{.79}
    & 1.0 & 1.0 & .98 & \textbf{.50} & .50 & \textbf{.53} & \textbf{.75} \\
\bottomrule
\end{tabular}
}
\end{table}

\subsection{Detailed Metric Specifications}
\label{app:metrics}

Table~\ref{tab:all_metrics} provides the complete metric inventory across all three evaluation settings.

\begin{table}[H]
\centering
\caption{Complete metric inventory. ``R'' = Rule-based, ``V'' = VLM-as-Judge. \textbf{Bold} indicates metrics novel to this work.}
\label{tab:all_metrics}
\small
\begin{tabular}{llcccl}
\toprule
\textbf{Metric} & \textbf{Type} & \textbf{S1} & \textbf{S2} & \textbf{S3} & \textbf{Adapted From} \\
\midrule
CNT (Object Count) & R & \cmark & \cmark & --- & SceneEval, Holodeck \\
\textbf{DIV (Diversity)} & R & \cmark & --- & --- & \textbf{Novel} \\
COL (Collision Rate) & R & \cmark & \cmark & \cmark & SAGE, VULCAN, SceneEval \\
GRAV (Gravity Validity) & R & \cmark & \cmark & --- & SceneEval, VULCAN \\
OOB (In-Bounds Rate) & R & \cmark & --- & --- & SceneEval \\
PRES (Preservation) & R & --- & --- & \cmark & Adapted \\
\textbf{ECNT (Edit Count)} & R & --- & --- & \cmark & \textbf{Novel} \\
\midrule
PF (Prompt Fidelity) & V & \cmark & \cmark & --- & Code2Worlds \\
SRF (Spatial Fidelity) & V & \cmark & --- & --- & SceneEval \\
LAES (Aesthetics) & V & \cmark & --- & --- & WorldCraft, SAGE \\
ILC (Image Correspondence) & V & --- & \cmark & --- & Adapted \\
STY (Style Consistency) & V & --- & \cmark & --- & Adapted \\
\textbf{EC (Edit Completeness)} & V & --- & --- & \cmark & \textbf{Novel} \\
\textbf{SC (Scene Coherence)} & V & --- & --- & \cmark & \textbf{Novel} \\
\textbf{LQ (Layout Quality)} & V & --- & --- & \cmark & \textbf{Novel} \\
\bottomrule
\end{tabular}
\end{table}

\paragraph{VLM scoring rubric.}
All VLM metrics use a 0--10 integer rubric with anchored descriptors:
\textbf{10}~= perfect realization;
\textbf{7}~= most elements correct, minor deviations;
\textbf{5}~= partial match, significant deviations;
\textbf{3}~= vaguely related, major elements missing;
\textbf{0}~= no relationship.
Scores are normalized to $[0,1]$ by dividing by 10 for aggregation.
The VLM receives the original prompt, six multi-angle screenshots, and the scene graph as context.

\paragraph{Rule-based metric details.}
\textbf{Collision detection} uses axis-aligned bounding box (AABB) overlap tests on all non-environment actor pairs where both actors have extent $> 100$ units.
\textbf{Gravity checking} tests whether each actor's bounding box bottom face lies within $\pm 200$ units of the ground plane ($z=0$).
\textbf{In-bounds checking} tests $|x|, |y| \leq 9500$ units (the $190\text{m} \times 190\text{m}$ ground plane).

\subsection{Ablation Results}
\label{app:cs1_abl_res}
\input{Tables/case_study_1/Ablation}

%% file: Tables/case_study_1/Ablation.tex
\begin{table}[H]
\centering
\caption{Ablation study on verification loop and self-evolving skill accumulation across four models. All methods use the same minimal system prompt without domain-specific hints. \textbf{Bold} = best per model--difficulty group. \colorbox{green!15}{Green} = improvement $\geq 0.10$ over baseline; \colorbox{red!15}{red} = degradation $\geq 0.10$.}
\label{tab:ablation}
\vspace{1mm}
\resizebox{\textwidth}{!}{%
\begin{tabular}{ll l ccc ccc c}
\toprule
\textbf{Model} & \textbf{Method} & \textbf{Task} & \textbf{CNT$\uparrow$} & \textbf{DIV$\uparrow$} & \textbf{COL$\uparrow$} & \textbf{PF$\uparrow$} & \textbf{SRF$\uparrow$} & \textbf{LAES$\uparrow$} & \textbf{Tools} \\
\midrule


\multirow{9}{*}{\rotatebox[origin=c]{90}{\textbf{Qwen3.5-9B}}}

 & \multirow{3}{*}{Baseline}
   & Easy & \textbf{1.00} & \textbf{1.00} & \textbf{1.00} & 0.40 & 0.40 & 0.30 & 8 \\
 & & Mid  & \textbf{1.00} & \textbf{0.90} & 0.92 & 0.40 & 0.40 & 0.40 & 21 \\
 & & Hard & 0.00 & 0.40 & \textbf{1.00} & 0.00 & 0.00 & 0.00 & 10 \\
\cmidrule(lr){2-10}

 & \multirow{3}{*}{+ Verify}
   & Easy & \textbf{1.00} & \cellcolor{red!15}0.43 & \textbf{1.00} & \cellcolor{green!15}\textbf{0.60} & \cellcolor{green!15}\textbf{0.60} & \cellcolor{green!15}\textbf{0.40} & 8 \\
 & & Mid  & \textbf{1.00} & 0.85 & 0.94 & 0.40 & \cellcolor{red!15}0.30 & \cellcolor{red!15}0.20 & 21 \\
 & & Hard & \cellcolor{green!15}\textbf{0.25} & \cellcolor{green!15}0.54 & \textbf{1.00} & \cellcolor{green!15}0.20 & \cellcolor{green!15}0.20 & \cellcolor{green!15}0.30 & 129 \\
\cmidrule(lr){2-10}

 & \multirow{3}{*}{+ Self-Evolve}
   & Easy & \textbf{1.00} & \textbf{1.00} & \textbf{1.00} & \cellcolor{green!15}0.50 & 0.40 & 0.30 & 8 \\
 & & Mid  & \textbf{1.00} & 0.86 & 0.92 & \cellcolor{green!15}\textbf{0.50} & 0.40 & \cellcolor{green!15}\textbf{0.50} & 22 \\
 & & Hard & 0.00 & \cellcolor{green!15}\textbf{1.00} & \textbf{1.00} & \cellcolor{green!15}\textbf{0.10} & \cellcolor{green!15}\textbf{0.10} & \cellcolor{green!15}\textbf{0.20} & 27 \\

\midrule


\multirow{9}{*}{\rotatebox[origin=c]{90}{\textbf{Qwen3.5-27B}}}

 & \multirow{3}{*}{Baseline}
   & Easy & \textbf{1.00} & \textbf{1.00} & 0.80 & 0.50 & 0.30 & 0.30 & 8 \\
 & & Mid  & \textbf{1.00} & \textbf{0.81} & 0.99 & 0.40 & 0.40 & 0.30 & 22 \\
 & & Hard & 0.25 & \textbf{0.50} & 0.88 & 0.20 & 0.20 & 0.30 & 26 \\
\cmidrule(lr){2-10}

 & \multirow{3}{*}{+ Verify}
   & Easy & \textbf{1.00} & \textbf{1.00} & 0.80 & 0.50 & \cellcolor{green!15}0.50 & 0.30 & 18 \\
 & & Mid  & \textbf{1.00} & 0.77 & 0.99 & \cellcolor{green!15}\textbf{0.50} & \cellcolor{green!15}\textbf{0.60} & \cellcolor{green!15}\textbf{0.50} & 23 \\
 & & Hard & \textbf{0.25} & 0.38 & \cellcolor{green!15}\textbf{0.99} & \cellcolor{green!15}\textbf{0.40} & \cellcolor{green!15}\textbf{0.40} & \cellcolor{green!15}\textbf{0.50} & 37 \\
\cmidrule(lr){2-10}

 & \multirow{3}{*}{+ Self-Evolve}
   & Easy & \textbf{1.00} & \textbf{1.00} & 0.80 & 0.40 & \cellcolor{green!15}0.40 & 0.30 & 8 \\
 & & Mid  & \textbf{1.00} & \textbf{0.81} & \textbf{1.00} & \cellcolor{green!15}\textbf{0.50} & \cellcolor{green!15}0.50 & \cellcolor{green!15}0.40 & 22 \\
 & & Hard & \textbf{0.25} & \textbf{0.50} & 0.95 & \cellcolor{green!15}0.30 & \cellcolor{green!15}0.30 & \cellcolor{green!15}0.40 & 31 \\

\midrule


\multirow{9}{*}{\rotatebox[origin=c]{90}{\textbf{Sonnet 4}}}

 & \multirow{3}{*}{Baseline}
   & Easy & \textbf{1.00} & \textbf{1.00} & 0.67 & 0.40 & 0.20 & 0.20 & 10 \\
 & & Mid  & \textbf{1.00} & \textbf{0.79} & \textbf{1.00} & 0.60 & 0.50 & 0.50 & 27 \\
 & & Hard & \textbf{0.50} & \textbf{0.44} & \textbf{0.99} & 0.40 & 0.30 & 0.30 & 60 \\
\cmidrule(lr){2-10}

 & \multirow{3}{*}{+ Verify}
   & Easy & \cellcolor{red!15}0.00 & \textbf{1.00} & \cellcolor{green!15}\textbf{1.00} & 0.40 & \cellcolor{green!15}0.30 & \cellcolor{green!15}0.30 & 21 \\
 & & Mid  & \textbf{1.00} & \cellcolor{red!15}0.65 & \textbf{1.00} & \textbf{0.60} & \cellcolor{green!15}\textbf{0.60} & 0.50 & 29 \\
 & & Hard & 0.25 & 0.36 & \textbf{0.99} & \cellcolor{green!15}\textbf{0.60} & \cellcolor{green!15}\textbf{0.50} & \cellcolor{green!15}\textbf{0.60} & 73 \\
\cmidrule(lr){2-10}

 & \multirow{3}{*}{+ Self-Evolve}
   & Easy & \textbf{1.00} & \textbf{1.00} & \cellcolor{green!15}0.80 & \cellcolor{green!15}\textbf{0.50} & \cellcolor{green!15}\textbf{0.50} & \cellcolor{green!15}\textbf{0.40} & 10 \\
 & & Mid  & \textbf{1.00} & 0.78 & \textbf{1.00} & 0.50 & 0.50 & 0.50 & 26 \\
 & & Hard & 0.25 & 0.47 & \textbf{0.99} & \cellcolor{green!15}0.50 & \cellcolor{green!15}0.40 & \cellcolor{green!15}\textbf{0.50} & 58 \\

\midrule


\multirow{9}{*}{\rotatebox[origin=c]{90}{\textbf{Opus 4}}}

 & \multirow{3}{*}{Baseline}
   & Easy & \textbf{1.00} & \textbf{1.00} & 0.80 & 0.40 & 0.40 & 0.30 & 11 \\
 & & Mid  & \textbf{1.00} & \textbf{0.85} & 0.91 & 0.30 & 0.20 & 0.20 & 23 \\
 & & Hard & 0.25 & 0.37 & \textbf{1.00} & 0.40 & 0.30 & 0.30 & 89 \\
\cmidrule(lr){2-10}

 & \multirow{3}{*}{+ Verify}
   & Easy & \textbf{1.00} & \cellcolor{red!15}0.88 & \cellcolor{green!15}\textbf{1.00} & \cellcolor{green!15}\textbf{0.70} & \cellcolor{green!15}\textbf{0.80} & \cellcolor{green!15}\textbf{0.50} & 12 \\
 & & Mid  & \textbf{1.00} & 0.77 & 0.98 & \cellcolor{green!15}\textbf{0.60} & \cellcolor{green!15}\textbf{0.70} & \cellcolor{green!15}\textbf{0.60} & 24 \\
 & & Hard & 0.25 & 0.31 & \textbf{1.00} & \cellcolor{green!15}\textbf{0.60} & \cellcolor{green!15}\textbf{0.50} & \cellcolor{green!15}\textbf{0.60} & 97 \\
\cmidrule(lr){2-10}

 & \multirow{3}{*}{+ Self-Evolve}
   & Easy & \textbf{1.00} & \textbf{1.00} & 0.80 & \cellcolor{red!15}0.00 & 0.40 & \cellcolor{green!15}0.40 & 11 \\
 & & Mid  & \textbf{1.00} & 0.76 & \cellcolor{green!15}\textbf{1.00} & \cellcolor{green!15}\textbf{0.60} & \cellcolor{green!15}0.50 & \cellcolor{green!15}0.50 & 28 \\
 & & Hard & \cellcolor{green!15}\textbf{0.50} & \textbf{0.42} & \textbf{1.00} & \cellcolor{green!15}0.50 & \cellcolor{green!15}0.50 & \cellcolor{green!15}0.50 & 62 \\

\bottomrule
\end{tabular}
}

\vspace{2mm}
{\footnotesize
\textit{Metrics:} CNT = object count accuracy, DIV = asset diversity, COL = collision avoidance, PF = prompt fidelity (VLM), SRF = spatial relation fidelity (VLM), LAES = layout aesthetics (VLM). All $\in [0,1]$.
\textit{Tools} = total MCP tool calls across all rounds. Self-evolving runs tasks sequentially (easy$\to$mid$\to$hard), accumulating skills between tasks. Verification uses up to 3 harness-driven fix rounds per task.
}
\end{table}

%% file: Appendix/C-CaseStudy_2.tex
\section{Case Study 2: Experimental Details}
\label{app:case_study_2_full}

\subsection{Compute Setup for Case Study 2}

For Case Study~2, we use an 8$\times$H100 server to run the embodied-agent rollout pipeline. 
We host eight \ours{} instances in parallel, each responsible for executing a subset of the navigation episodes. 
The instances are synchronized with the agent-inference workers for observation collection, action execution, reward computation, and trajectory logging. 
With this setup, processing the 1.2K-episode training set requires approximately 8 hours per full pass.

\subsection{Experiment Design}
\label{app:case_study_2}

\paragraph{Environment generation.}
Training environments are generated by \textsc{SimCoder} using text-to-scene prompts spanning five urban archetypes: downtown intersections, residential neighborhoods, industrial districts, commercial avenues, and mixed-use blocks. Each environment is generated on a fixed $190\text{m}\times190\text{m}$ ground plane and automatically exported as a Gymnasium environment via the \ours{} embodied interface (\S\ref{sec:embodied_agent}). To study the effect of environmental diversity on downstream learning, we vary the number of distinct training environments from 1 to 30 while holding the total training-episode budget fixed at 200 episodes.

\paragraph{Episode generation.}
Navigation episodes are generated automatically from each environment's UE5 NavMesh via the following procedure:
\begin{enumerate}[leftmargin=*, itemsep=2pt]
    \item \textbf{NavMesh computation.} UE5's NavMesh is built from the static geometry of the generated scene, producing a walkable region that inherits all obstacles and building footprints automatically.
    \item \textbf{Start--goal sampling.} Start and goal positions are sampled uniformly from the walkable region. For PointNav, the goal is a 2D coordinate; for ObjectNav, the goal is a semantic category drawn from objects present in the scene.
    \item \textbf{Path computation.} The geodesic shortest path between start and goal is computed via NavMesh A* search, recording the path waypoints and length $L^*$.
    \item \textbf{Filtering.} An episode is retained if: (i) the path is fully reachable, (ii) the path length satisfies $3\text{m} \leq L^* \leq 20\text{m}$, and (iii) for ObjectNav, the target object is visible from at least one waypoint along the shortest path.
\end{enumerate}
We generate 1,200 training episodes distributed across all training environments and 329 held-out test episodes on unseen \ours{}-generated maps.

\paragraph{Agent observation and action space.}
At each step the agent receives a structured observation tuple: an RGB image ($224\times224$), bearing to goal (degrees), geodesic distance to goal (meters), and elapsed step count. The discrete action space consists of four primitives: \texttt{move\_forward} (0.25\,m), \texttt{turn\_left} ($15^{\circ}$), \texttt{turn\_right} ($15^{\circ}$), and \texttt{stop}. An episode terminates on \texttt{stop} or after 40 steps.

\paragraph{Training protocol.}
Rather than gradient-based policy optimization, the agent learns in a training-free manner by accumulating and retrieving episodic memory (Appendix~\ref{app:memory}) across training episodes. For each training episode the agent attempts navigation; the full action--observation trajectory and outcome are stored in the memory module at three granularities (step, trajectory, task). During test-time evaluation, the feedback signal is removed and the agent relies solely on retrieved memory to condition its policy.

\paragraph{Cross-benchmark evaluation.}
Beyond held-out \ours{} test environments, all agents are evaluated on SimWorld-MMNav~\citep{zhuang2026simworldrobotics}, an independently constructed navigation benchmark spanning three difficulty tiers (easy, medium, hard) in unseen UE5 environments. This evaluation directly measures whether navigation knowledge acquired in \ours{}-generated worlds transfers beyond the training distribution.

\subsection{Metric Definitions}
\label{app:cs2_metric_def}
We report four complementary metrics covering task success, path efficiency, and trajectory fidelity.

\paragraph{Notation.}
Let $N$ denote the number of evaluation episodes. For episode $i$, let $P_i = (p_1, \ldots, p_T)$ be the executed trajectory and $P_i^*$ the geodesic shortest-path reference, with lengths $L_i$ and $L_i^* = d_i^0$ respectively. Let $d_i$ be the geodesic distance from the agent's final position to the goal and $\delta = 1.0$\,m the success threshold.

\paragraph{Success Rate (SR).}
\begin{equation}
    \mathrm{SR} = \frac{1}{N}\sum_{i=1}^N \mathbf{1}(d_i < \delta).
\end{equation}
Binary indicator of task completion. The primary metric reported in the main paper.

\paragraph{Success weighted by Path Length (SPL).}
\begin{equation}
    \mathrm{SPL} = \frac{1}{N}\sum_{i=1}^N \mathbf{1}(d_i < \delta)\cdot\frac{L_i^*}{\max(L_i,\, L_i^*)}.
\end{equation}
Penalizes inefficient paths; a successful agent that takes a much longer route than necessary scores below 1.

\paragraph{SoftSPL.}
\begin{equation}
    \mathrm{SoftSPL} = \frac{1}{N}\sum_{i=1}^N \max\!\left(0,\,1-\frac{d_i}{d_i^0}\right)\cdot\frac{L_i^*}{\max(L_i,\, L_i^*)}.
\end{equation}
A continuous relaxation of SPL that rewards partial progress toward the goal rather than requiring binary success.

\paragraph{Normalized Dynamic Time Warping (nDTW).}
\begin{equation}
    \mathrm{nDTW} = \frac{1}{N}\sum_{i=1}^N \exp\!\left(-\frac{\mathrm{DTW}(P_i,\,P_i^*)}{\eta\cdot|P_i^*|}\right),
\end{equation}
where $\mathrm{DTW}(\cdot,\cdot)$ is the dynamic time warping alignment cost, $|P_i^*|$ the number of waypoints in the reference path, and $\eta$ a normalization constant ($\eta=5$ in our experiments). nDTW measures trajectory fidelity beyond endpoint success, rewarding agents that closely follow the reference path.

\subsection{Hierarchical Memory Design}
\label{app:memory}

The memory module extends the Generative Agents~\citep{park2023generativeagentsinteractivesimulacra} and ExpeL~\citep{zhao2024expel} frameworks with \emph{multi-level} updates that distill experience at step, trajectory, and task granularities, allowing strategies to be retrieved at the appropriate scope during inference.

\paragraph{Three-level memory structure.}
\begin{itemize}[leftmargin=*, itemsep=3pt]
    \item \textbf{Step-level memory (L1).} Stores fine-grained action--observation pairs within a single episode. Primarily used for within-episode self-correction (e.g., detecting a position revisit and triggering a recovery maneuver).
    \item \textbf{Trajectory-level memory (L2).} After each episode, the full trajectory is summarized into a compact natural-language record capturing navigation strategy, key decision points, and outcome. L2 records are indexed by environment features (obstacle density, path length tier) to enable retrieval of relevant past trajectories in new episodes.
    \item \textbf{Task-level memory (L3).} Periodically distills patterns across multiple L2 records into abstract navigation principles (e.g., ``in dense urban environments, maintain a $45^{\circ}$ offset from the goal bearing to avoid building corners''). L3 knowledge generalizes across environments and is injected into the system prompt as a high-priority context prefix.
\end{itemize}

\paragraph{Memory update.}
After each training episode, the module performs three sequential updates:
\begin{enumerate}[leftmargin=*, itemsep=2pt]
    \item \textbf{L1 flush.} Step-level records from the completed episode are compressed and stored.
    \item \textbf{L2 summarization.} An LLM summarizer condenses the episode trajectory and outcome into an L2 record, tagging it with environment metadata.
    \item \textbf{L3 distillation.} Every 10 episodes, an LLM distiller identifies recurring patterns across recent L2 records and appends new L3 principles, replacing outdated ones.
\end{enumerate}

\paragraph{Memory retrieval.}
At the start of each inference episode, the agent retrieves the top-$k$ most relevant L2 records (by environment similarity) and the current L3 knowledge base. These are formatted into a structured memory block injected into the agent's context before the first action, giving the agent a head start grounded in prior experience.

\subsection{Observation Modality Ablation}
\label{app:obs_abl}

Table~\ref{tab:ablation_obs} reports the ablation over observation modalities. RGB-D consistently outperforms depth-only or text-only inputs across model scales, confirming that complementary geometric (depth) and semantic (RGB) cues both contribute to navigation performance. Text-only (bearing + distance scalars) provides a surprisingly competitive baseline for larger models but degrades sharply at smaller scales, suggesting that visual grounding becomes more important as in-context reasoning capacity decreases. Three models is trained on 100 episodes training set and tested on 30 episode unseen test set.

\input{Tables/case_study_2/ablation_observation}

%% file: Tables/case_study_2/ablation_observation.tex
\begin{table}[H]
\centering
\caption{\textbf{Ablation on observation modalities across model scales.}
Each Qwen3.5 model is evaluated on ObjectNav with three vision configurations
(RGB, Depth, RGB+Depth) and a text-only baseline where the agent receives
goal distance and bearing scalars but no image. Metrics are success
rate (SR) and SoftSPL (SS), both in $[0,1]$ ($\uparrow$). Best values within
each (model scale, split, metric) group are highlighted in \textbf{bold}; tied
values share the bold marker; tied zeros are not bolded.}
\label{tab:ablation_obs}
\vspace{4pt}
\small
\setlength{\tabcolsep}{4.5pt}
\renewcommand{\arraystretch}{1.2}
\resizebox{\textwidth}{!}{%
\begin{tabular}{@{}l cc cc cc cc@{}}
\toprule
& \multicolumn{4}{c}{\textbf{Qwen3.5-27B}}
& \multicolumn{4}{c}{\textbf{Qwen3.5-9B}} \\
\cmidrule(lr){2-5} \cmidrule(lr){6-9}
\textbf{Observation}
& \multicolumn{2}{c}{\textit{Seen}} & \multicolumn{2}{c}{\textit{Unseen}}
& \multicolumn{2}{c}{\textit{Seen}} & \multicolumn{2}{c}{\textit{Unseen}} \\
\cmidrule(lr){2-3} \cmidrule(lr){4-5}
\cmidrule(lr){6-7} \cmidrule(lr){8-9}
& \textbf{SR}$\uparrow$ & \textbf{SS}$\uparrow$
& \textbf{SR}$\uparrow$ & \textbf{SS}$\uparrow$
& \textbf{SR}$\uparrow$ & \textbf{SS}$\uparrow$
& \textbf{SR}$\uparrow$ & \textbf{SS}$\uparrow$ \\
\midrule
RGB
& .636{\scriptsize$\pm$.487} & .775{\scriptsize$\pm$.227}
& \textbf{.833}{\scriptsize$\pm$.408} & \textbf{.721}{\scriptsize$\pm$.373}
& \textbf{.136}{\scriptsize$\pm$.347} & \textbf{.279}{\scriptsize$\pm$.332}
& 0 & .070{\scriptsize$\pm$.113} \\

Depth
& \textbf{.659}{\scriptsize$\pm$.479} & \textbf{.784}{\scriptsize$\pm$.198}
& .667{\scriptsize$\pm$.516} & .694{\scriptsize$\pm$.356}
& .045{\scriptsize$\pm$.211} & .194{\scriptsize$\pm$.308}
& 0 & .179{\scriptsize$\pm$.255} \\

RGB + Depth
& .614{\scriptsize$\pm$.493} & .696{\scriptsize$\pm$.308}
& .667{\scriptsize$\pm$.516} & .614{\scriptsize$\pm$.386}
& .068{\scriptsize$\pm$.255} & .202{\scriptsize$\pm$.290}
& \textbf{.167}{\scriptsize$\pm$.408} & \textbf{.223}{\scriptsize$\pm$.331} \\

Text only
& .455{\scriptsize$\pm$.504} & .693{\scriptsize$\pm$.267}
& .500{\scriptsize$\pm$.548} & .619{\scriptsize$\pm$.369}
& .068{\scriptsize$\pm$.255} & .235{\scriptsize$\pm$.276}
& 0 & .154{\scriptsize$\pm$.162} \\
\bottomrule
\end{tabular}
}
\end{table}

%% file: Appendix/D-CaseStudy_3.tex
\section{Case Study 3: Experimental Details}
\label{app:case_study_3}

\subsection{Structured Prompt Evolution}

The embodied agent in Case Study 3 optimizes its navigation policy via a structured variant of GEPA~\citep{li2025gepa}. Standard GEPA rewrites the entire system prompt at each update step, which risks overwriting effective strategies already in place. Our structured variant instead maintains an ordered list of prioritized decision rules $\mathcal{R}_t = [r_1, r_2, \ldots, r_k]$ that are injected verbatim into the system prompt in fixed priority order.

\paragraph{Rule format.}
Each rule $r_i$ is a concise, actionable navigation heuristic expressed in natural language. Representative examples:
\begin{itemize}[leftmargin=2em, itemsep=2pt]
    \item \emph{``If the bearing to the goal is within $\pm30^{\circ}$ and no obstacle is visible, prefer \texttt{move\_forward}.''}
    \item \emph{``If the same grid cell has been visited three or more times in the last ten steps, execute two consecutive \texttt{turn\_right} actions to escape the loop.''}
    \item \emph{``When the reported distance to goal drops below 1.5\,m, issue \texttt{stop} immediately.''}
\end{itemize}
Rules are ordered by precedence: more specific rules (low-level reactive behaviors) appear first and override more general rules when both apply. A later rule takes effect only when no earlier rule fires.

\paragraph{Update mechanism.}
After each training episode, failure modes $\mathcal{F}_t$ are extracted from the trajectory and passed to the rule-synthesis LLM, which generates new rules $\Delta\mathcal{R}(\mathcal{F}_t)$ that address observed failures without contradicting existing rules:
\begin{equation}
    \mathcal{R}_{t+1} = \mathcal{R}_t \;\cup\; \Delta\mathcal{R}(\mathcal{F}_t).
\end{equation}
New rules are appended at the end of the list. A pruning step removes rules that have not been activated in the preceding 20 episodes. The list is capped at 30 active rules to prevent prompt bloat.

\paragraph{Failure mode taxonomy.}
The rule-synthesis prompt presents the full action--observation trajectory alongside a structured failure taxonomy:

\begin{enumerate}[leftmargin=2em, itemsep=2pt]
    \item \textbf{Directional errors.} Turning when a clear forward path exists; misalignment between bearing and action.
    \item \textbf{Loop behavior.} Revisiting the same position repeatedly; oscillating between two nearby cells.
    \item \textbf{Goal proximity failure.} Stopping too far from the goal; overshooting and failing to recover.
    \item \textbf{Obstacle avoidance failure.} Repeated collisions with the same obstacle type (e.g., building corner).
    \item \textbf{Timeout.} Reaching the 500-step budget without issuing \texttt{stop}.
\end{enumerate}
For each detected failure, the LLM synthesizes one to three targeted rules. The synthesis prompt also receives the current rule list $\mathcal{R}_t$ to enforce non-contradiction.

\subsection{Difficulty Curriculum Parameterization}

\textsc{SimCoder} controls episode difficulty via three axes: geodesic path length, initial heading offset (angular deviation from the goal direction at episode start), and dynamic obstacle density (fraction of navigable area occupied by movable obstacles).

Table~\ref{tab:difficulty_levels} lists the eight difficulty levels (L0--L7) with their exact parameter ranges. Levels are designed so that each axis increases monotonically: longer paths demand extended planning, larger heading offsets require deliberate reorientation before progress can begin, and higher obstacle density increases path variability and collision risk.

\begin{table}[H]
\centering
\caption{\textbf{Difficulty level parameterization.} Path length is the geodesic distance from start to goal; heading offset is the initial angular deviation from the goal direction; obstacle density is the fraction of navigable area occupied by dynamic obstacles.}
\label{tab:difficulty_levels}
\small
\setlength{\tabcolsep}{8pt}
\renewcommand{\arraystretch}{1.25}
\begin{tabular}{@{}c S[table-format=4.0] @{--} S[table-format=4.0] S[table-format=3.0] @{--} S[table-format=3.0] c@{}}
\toprule
\textbf{Level} & \multicolumn{2}{c}{\textbf{Path Length (cm)}} & \multicolumn{2}{c}{\textbf{Heading Offset (°)}} & \textbf{Obstacle Density} \\
\midrule
L0 & 400  & 800   & 0  & 15  & 0.00 \\
L1 & 600  & 1000  & 0  & 30  & 0.05 \\
L2 & 800  & 1400  & 0  & 45  & 0.10 \\
L3 & 1000 & 1800  & 0  & 60  & 0.15 \\
L4 & 1200 & 2200  & 0  & 90  & 0.20 \\
L5 & 1500 & 2600  & 0  & 120 & 0.25 \\
L6 & 2000 & 3000  & 0  & 150 & 0.30 \\
L7 & 2500 & 3500  & 0  & 180 & 0.35 \\
\bottomrule
\end{tabular}
\end{table}

\paragraph{Mastery-gating thresholds.}
\textsc{SimCoder} advances to the next level only when the agent's rolling success rate $\bar{S}_t$---averaged over the most recent 5 epochs---exceeds a level-specific threshold $\tau_\ell$:
\begin{equation}
    \ell_{t+1} = \ell_t + \mathbf{1}\!\bigl[\bar{S}_t \geq \tau_{\ell_t}\bigr].
\end{equation}
Thresholds decrease with difficulty to maintain challenge as tasks grow harder: $(\tau_0, \ldots, \tau_7) = (0.80,\,0.75,\,0.70,\,0.65,\,0.60,\,0.55,\,0.50,\,0.45)$. This mastery-gating scheme mirrors the zone-of-proximal-development principle~\citep{vygotsky1978mind}: the agent is exposed to the next challenge only once it has demonstrated reliable competence at the current level.

\subsection{Training and Evaluation Protocol}

\paragraph{Co-evolution loop.}
Each training epoch proceeds in four steps:
\begin{enumerate}[leftmargin=2em, itemsep=2pt]
    \item \textbf{Episode generation.} \textsc{SimCoder} generates 20 navigation episodes at the current difficulty level $\ell_t$ using the \ours{} embodied interface.
    \item \textbf{Agent rollout.} The embodied agent attempts all 20 episodes; action--observation trajectories and outcomes are recorded.
    \item \textbf{Policy update.} The prompt-evolution module updates $\mathcal{R}_t$ from extracted failure modes.
    \item \textbf{Difficulty advance.} The rolling success rate $\bar{S}_t$ is computed; if $\bar{S}_t \geq \tau_{\ell_t}$, \textsc{SimCoder} advances to $\ell_{t+1}$.
\end{enumerate}
The benchmark evaluation on SimWorld-MMNav is run every 5 epochs. Training runs for 25 epochs total (500 episodes).

\paragraph{Baseline conditions.}
\begin{itemize}[leftmargin=2em, itemsep=3pt]
    \item \textbf{Co-evolving environment + learning.} The full co-evolution loop as described above. \textsc{SimCoder} adapts difficulty; the agent accumulates rules.
    \item \textbf{Fixed-difficulty environment + learning.} The same prompt-evolution mechanism, but \textsc{SimCoder} holds difficulty fixed at L3 (medium) throughout. Isolates the contribution of the adaptive curriculum from the rule-learning mechanism.
    \item \textbf{No learning.} A fixed system prompt with no rule accumulation and no curriculum adaptation. Serves as the zero-shot LLM baseline.
\end{itemize}
All conditions use Qwen3.5-9B and the same initial system prompt with an empty rule list.

\paragraph{Evaluation.}
Final performance is measured on SimWorld-MMNav~\citep{zhuang2026simworldrobotics}, a held-out benchmark of 329 episodes spanning easy, medium, and hard difficulty tiers in independently authored UE5 environments unseen during training. We report Success Rate (SR) as the primary metric.

%% file: Appendix/E-Examples.tex
\section{Prompt Examples}
\label{app:prompts}

\paragraph{Text-to-Scene (Hard).}
\emph{``Design a full residential neighborhood. Build two parallel streets with buildings on both sides: place 3 buildings along the north side and 3 buildings along the south side of each street (total 6 buildings in two rows). Connect the streets with a cross road. Line all roads with trees on both sides. Create a central park between the two streets with tables, couches, and trash bins. Add a fire hydrant at every street intersection. Place scooters and carts parked along the roads. Mark one intersection with road cones and road blockers as a construction zone.''}

\paragraph{Image+Text-to-Scene (Hard).}
\emph{``Build a dense urban block matching this aerial photo: multiple buildings arranged in a grid pattern with streets between them, trees lining the streets, and vehicles and street furniture throughout.''}
(Accompanied by a real aerial photograph of a city block.)

\paragraph{Scene Editing (Hard).}
\emph{``Expand the plaza into a larger district: add 2 new buildings on the north side. Add roads connecting the new buildings to the existing ones. Plant 6 more trees to line the new roads. Create a marketplace area with 3 tables and 2 carts near the center. Add road cones and road blockers to mark a construction zone near the new buildings. Place fire hydrants at each road intersection and trash bins along the sidewalks.''}

\newpage
\section{Qualitative Examples}
The qualitative examples in this appendix use the model snapshots available at the time of figure generation. 
They are provided only as visual examples and are not included in the quantitative comparisons in Table~\ref{tab:main_results}.
\label{app:examples}

\subsection{Text-to-Scene}
\begin{figure}[htbp]
\centering

\begin{tcolorbox}[
    colback=gray!5,
    colframe=gray!40,
    boxrule=0.4pt,
    arc=2pt,
    left=5pt, right=5pt, top=5pt, bottom=5pt,
    title={\small \textbf{Prompt P1}},
    fonttitle=\bfseries,
    coltitle=black
]
\scriptsize
Build a downtown city-block scene at a 4-way cross intersection in Unreal Engine 5, set during the morning. The aesthetic, building heights, density, and how lively each side feels are completely your call - make it feel like a real downtown intersection.
\vspace{4pt}

It needs to be a proper 4-way cross, not a T-intersection, with one road running along the X axis and another cutting through it along Y. Make sure the seam where they meet is clean. Nothing should overlap the road or sidewalk - buildings stay outside the road footprint, cars stay in their lanes, pedestrians stay on sidewalks or in crosswalks. Run everything in a single script call, then take one screenshot, then stop.
\vspace{4pt}

Everything else is yours. How long each road is, which quadrants get tall glass towers versus shorter brick mid-rises, whether buildings are packed tight or have occasional plazas and setbacks - that's your composition. Same goes for sidewalk dressing like trees, café tables, and benches, how many people are out, car patterns, and whether the two roads feel like the same neighborhood or two different ones. Try to get specific and make real decisions rather than just defaulting to whatever looks average - the more deliberate the choices, the more the scene has a point of view.
\vspace{4pt}

The skill library is staged and ready to use. Before writing the script, briefly sketch out what kind of downtown this is, what the focal points are, and how the density and lighting support the mood - then write it.
\end{tcolorbox}

\vspace{8pt}

\begin{minipage}[t]{0.31\linewidth}
    \centering
    \includegraphics[width=\linewidth]{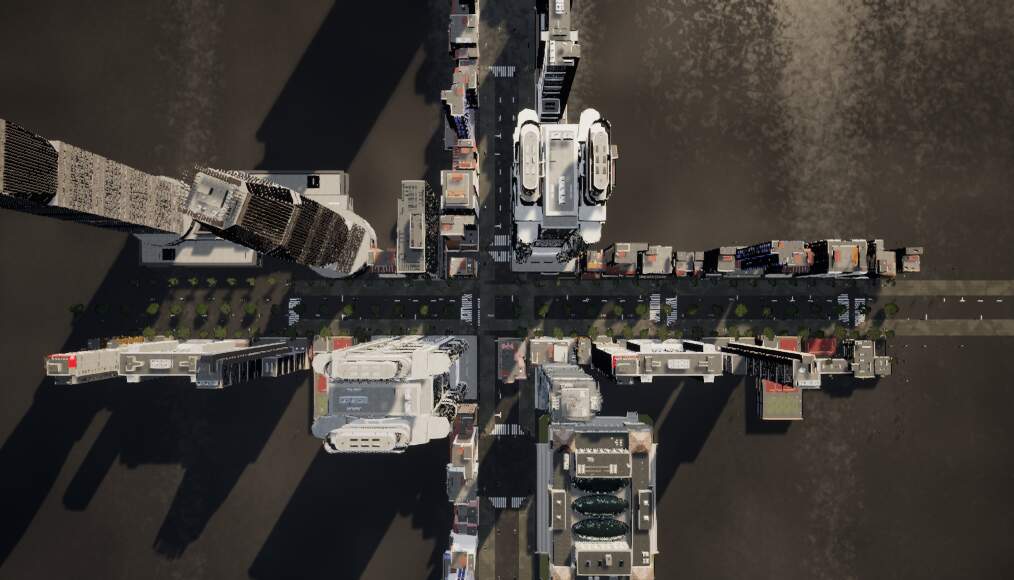}
    \captionof*{figure}{\small Claude Opus 4.7}
\end{minipage}
\hfill
\begin{minipage}[t]{0.31\linewidth}
    \centering
    \includegraphics[width=\linewidth]{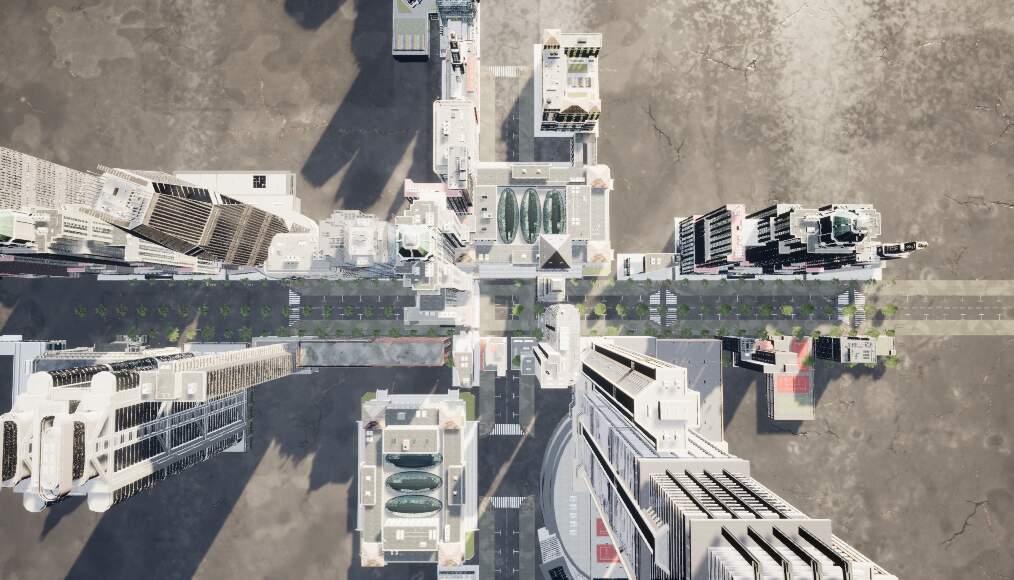}
    \captionof*{figure}{\small Qwen 3.5-27B}
\end{minipage}
\hfill
\begin{minipage}[t]{0.31\linewidth}
    \centering
    \includegraphics[width=\linewidth]{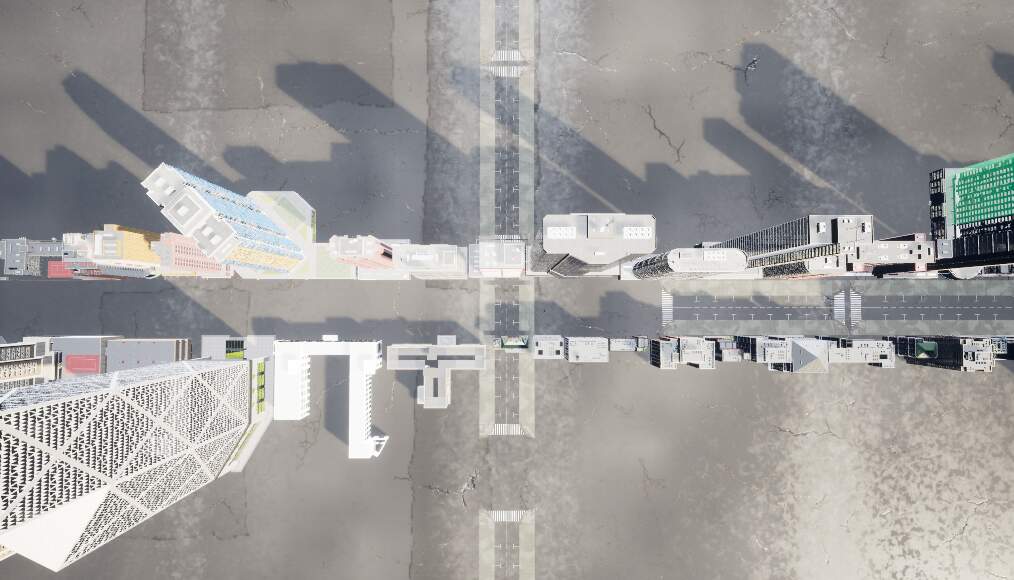}
    \captionof*{figure}{\small Qwen 3.5-9B}
\end{minipage}

\caption{\textbf{Qualitative Example P1.} Output scenes generated by three model backbones given the same downtown city-block intersection prompt.}
\label{fig:f1_p1_qualitative}
\end{figure}
\vspace{-5mm}
\newpage

\subsection{Image+Text-to-Scene}
\begin{figure}[htbp]
\centering

\begin{tcolorbox}[
    colback=gray!5,
    colframe=gray!40,
    boxrule=0.4pt,
    arc=2pt,
    left=5pt, right=5pt, top=5pt, bottom=5pt,
    title={\small \textbf{Prompt P1}},
    fonttitle=\bfseries,
    coltitle=black
]
\scriptsize
\begin{wrapfigure}{r}{0.35\linewidth}
    \centering
    \vspace{-10pt}
    \includegraphics[width=\linewidth]{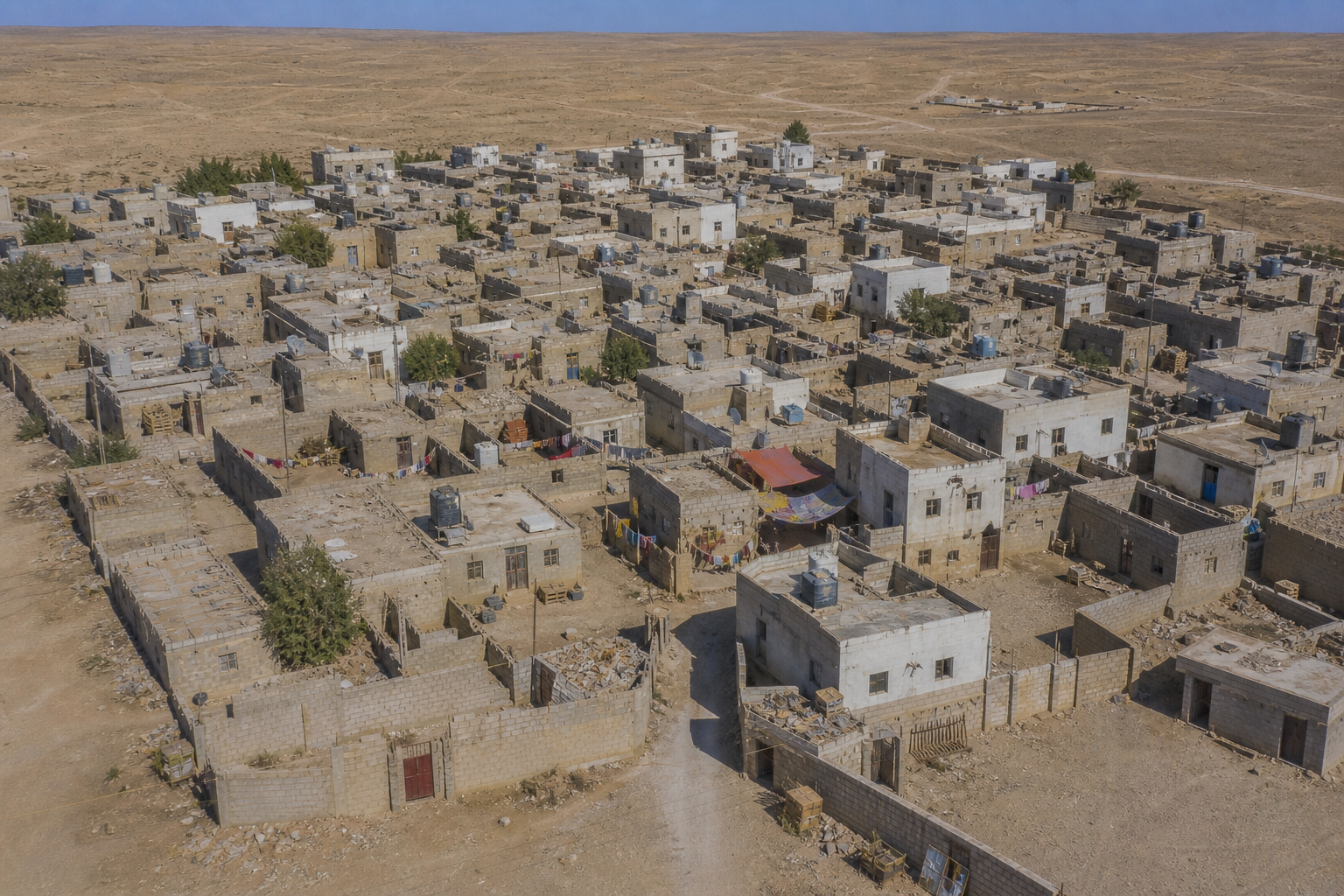}
    \captionof*{figure}{\small Reference Image}
\end{wrapfigure}
Create a large Middle Eastern desert village that matches the attached reference image. The reference shows a dense settlement of low-rise buildings spread across desert terrain - your job is to recreate that feeling using the tools available.

\vspace{4pt}
A few things are fixed: start by establishing the desert ground, then build the village with somewhere between 150 and 200 houses, then set up the lighting. The village builder already handles clutter internally so don't add random props on top of that. Pick your own random seed.

\vspace{4pt}
Everything else is a judgment call based on what you see in the reference. How many houses, how tightly packed they are, how organic or grid-like the layout feels - those are yours to decide. Think about whether rooftops should have water tanks and antennas, whether some houses get compound walls, and whether there's debris and clutter visible between buildings. Beyond the village itself, think about where dirt roads cut through, where fences sit, how many water tanks and market tents are scattered around the edges, and where they'd naturally fall. Study the reference and let it guide those decisions rather than just picking middle-of-the-road values for everything.
\end{tcolorbox}

\vspace{8pt}

\begin{minipage}[t]{0.31\linewidth}
    \centering
    \includegraphics[width=\linewidth]{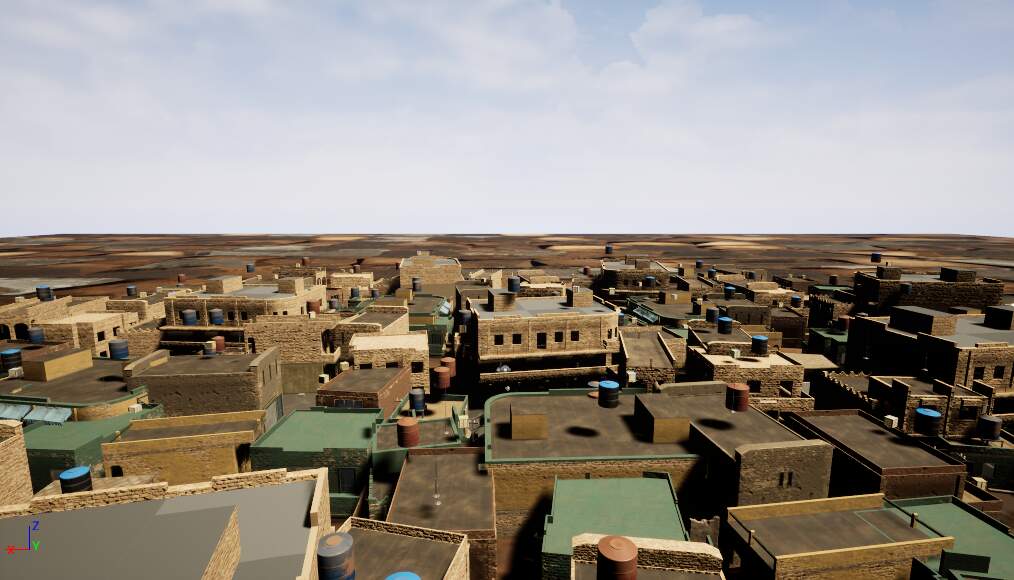}
    \captionof*{figure}{\small Claude Opus 4.7}
\end{minipage}
\hfill
\begin{minipage}[t]{0.31\linewidth}
    \centering
    \includegraphics[width=\linewidth]{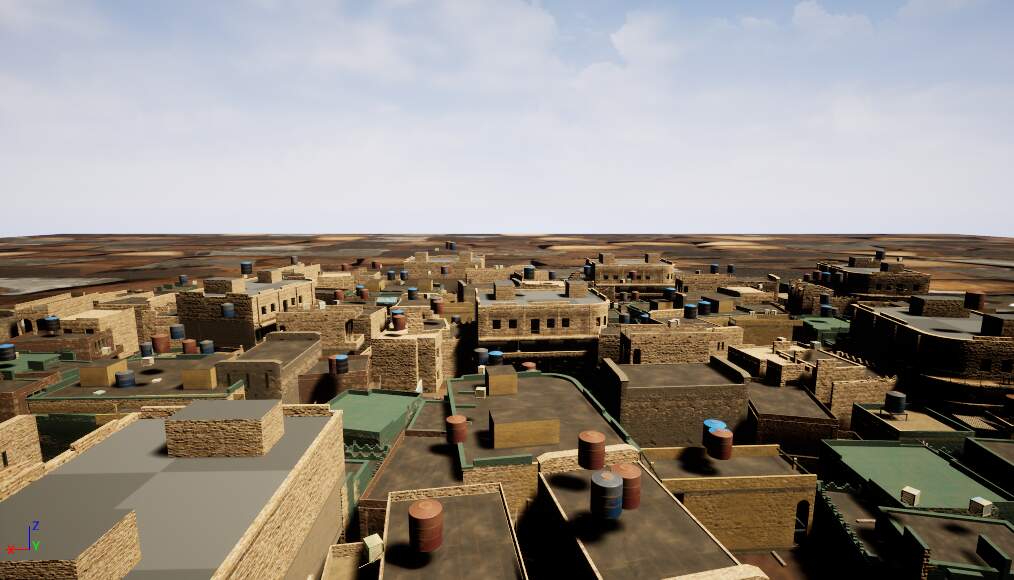}
    \captionof*{figure}{\small Qwen 3.5-27B}
\end{minipage}
\hfill
\begin{minipage}[t]{0.31\linewidth}
    \centering
    \includegraphics[width=\linewidth]{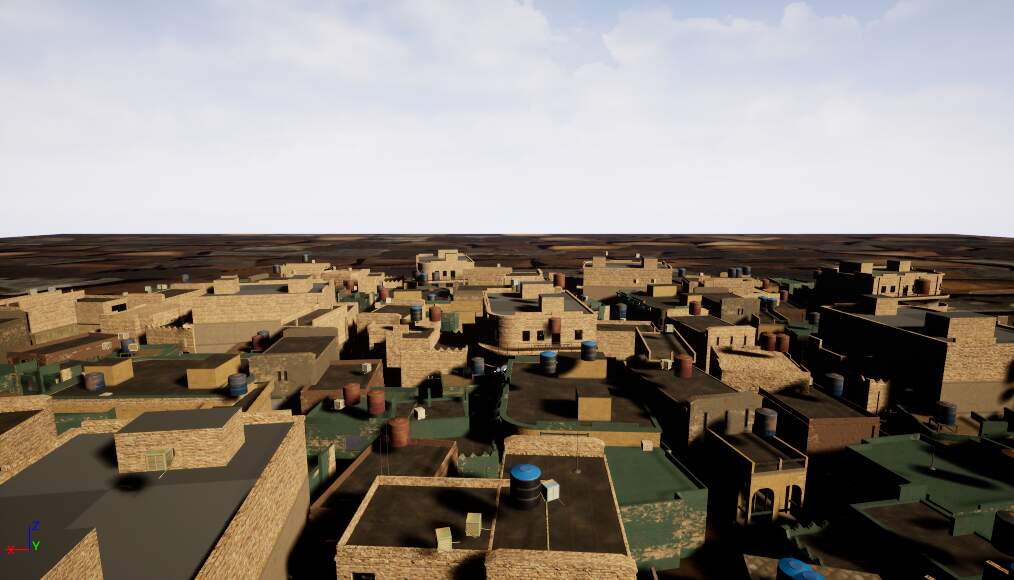}
    \captionof*{figure}{\small Qwen 3.5-9B}
\end{minipage}

\caption{\textbf{Qualitative Example P1.} Top: prompt and reference image. Bottom: rendered UE5 screenshots from each model backbone.}
\label{fig:f2_p1_qualitative}
\end{figure}

\subsection{Scene Editing}
\begin{figure}[htbp]
\centering

\begin{minipage}[t]{0.655\linewidth}
    \vspace{0pt}
    \begin{tcolorbox}[
        colback=gray!5,
        colframe=gray!40,
        boxrule=0.4pt,
        arc=2pt,
        left=5pt, right=5pt, top=5pt, bottom=5pt,
        title={\small \textbf{Prompt P1}},
        fonttitle=\bfseries,
        coltitle=black
    ]
    \scriptsize
    Build a two-sided residential street in the current scene, which already has six starting buildings and six trees. Keep one existing building, remove the others, then fill out both sides using exactly two building types, the kept one plus one other medium sized building of your choosing. Check the scene first, then remove the unwanted buildings, road is already laid but lay the sidewalk along it, and build outward. The north side incorporates the kept building with additional buildings extending west and east; the south side gets its own row facing north. Once buildings are placed, add pedestrians on both sides, vehicles in the road, and finish with a cloudy sky.
    \vspace{4pt}
    Street length between 18,000 and 28,000, road width 800–1,100, sidewalk width 400–700, road center Y between -2,500 and -1,700. North row: 3 to 6 buildings on each side of the existing house. South row: 8 to 14. Pedestrians: 40 to 70 per sidewalk. Vehicles: 10 to 16. Two building types only.
    \end{tcolorbox}
\end{minipage}
\hfill
\begin{minipage}[t]{0.31\linewidth}
    \vspace{0pt}
    \centering
    \includegraphics[width=\linewidth]{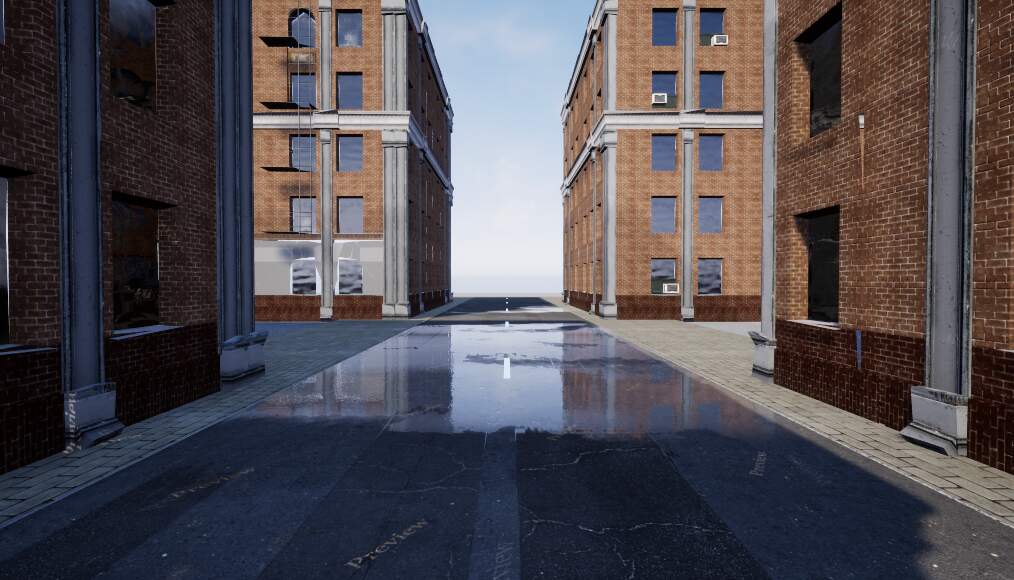}
    \captionof*{figure}{\small Original Scene}
\end{minipage}

\vspace{8pt}

\begin{minipage}[t]{0.31\linewidth}
    \centering
    \includegraphics[width=\linewidth]{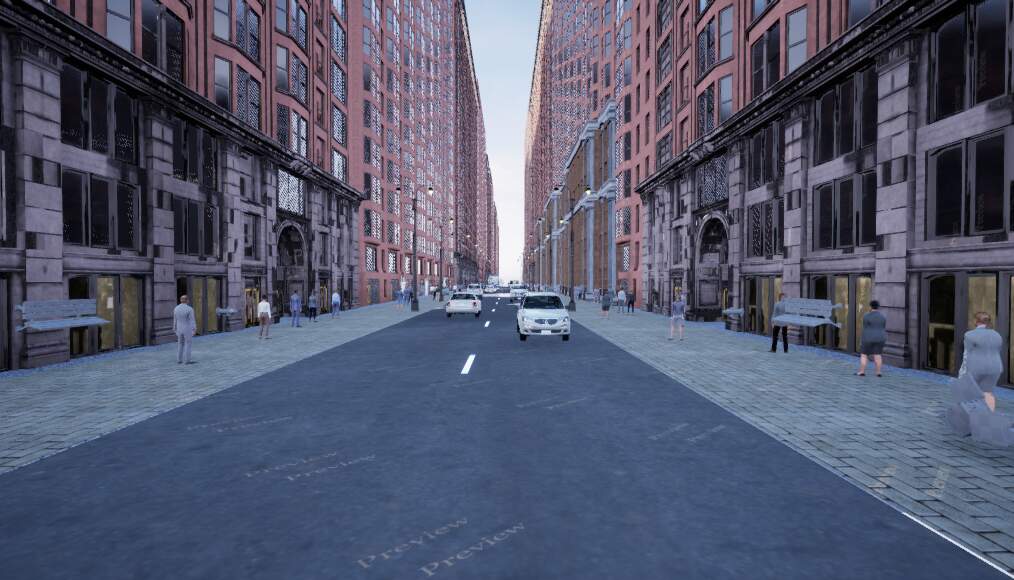}
    \captionof*{figure}{\small Claude Opus 4.7}
\end{minipage}
\hfill
\begin{minipage}[t]{0.31\linewidth}
    \centering
    \includegraphics[width=\linewidth]{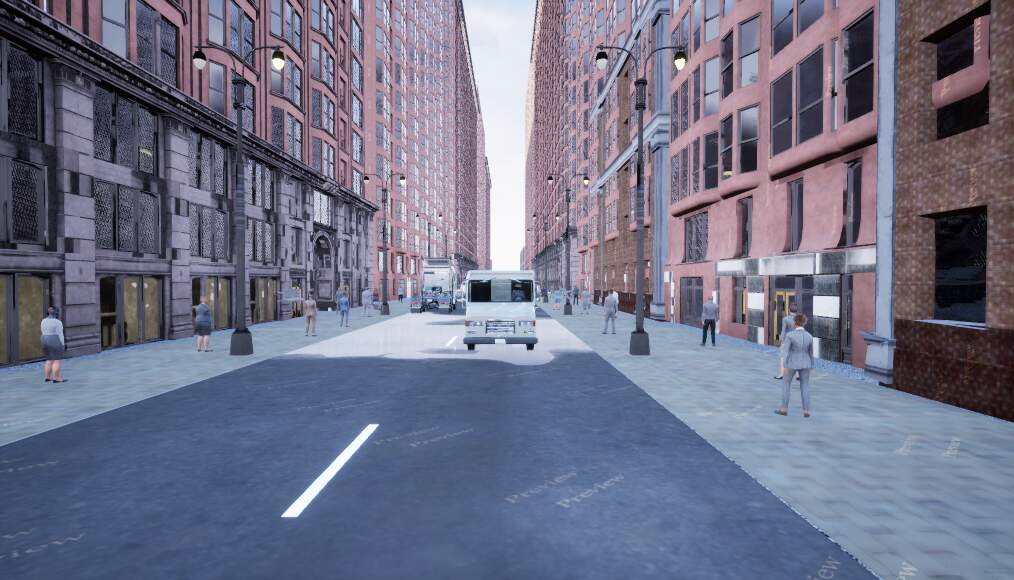}
    \captionof*{figure}{\small Qwen 3.5-27B}
\end{minipage}
\hfill
\begin{minipage}[t]{0.31\linewidth}
    \centering
    \includegraphics[width=\linewidth]{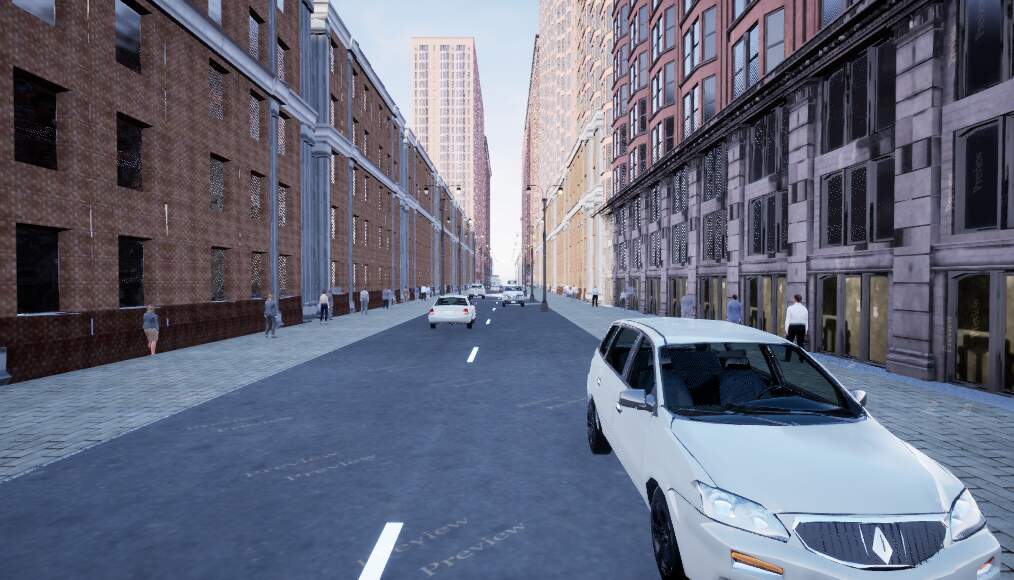}
    \captionof*{figure}{\small Qwen 3.5-9B}
\end{minipage}

\caption{\textbf{Qualitative Example P1.} Top: editing prompt and the original scene prior to modification. Bottom: rendered UE5 screenshots showing each model's edited scene, built on top of the same starting configuration.}
\label{fig:f3_p1_qualitative}
\end{figure}

\begin{figure}[htbp]
\centering

\begin{minipage}[t]{0.655\linewidth}
    \vspace{0pt}
    \begin{tcolorbox}[
        colback=gray!5,
        colframe=gray!40,
        boxrule=0.4pt,
        arc=2pt,
        left=5pt, right=5pt, top=5pt, bottom=5pt,
        title={\small \textbf{Prompt P2}},
        fonttitle=\bfseries,
        coltitle=black
    ]
    \scriptsize
    Build a two-sided residential street in the current scene, which already has six starting buildings and six trees. Remove all of them, keep one, and build out both sides using a fixed pool of four small 2-to-3 storey buildings, mixed randomly per tile. Check the scene first, remove the unwanted buildings, lay the sidewalk along the road, then build outward. North side incorporates the kept building with rows extending east and west; south side gets its own row facing north. Add pedestrians on both sides, vehicles in the road, and finish with a cloudy sky.
    \vspace{4pt}
    Pick real values: street length 18,000–28,000, road width 800–1,100, sidewalk width 400–700, road center Y between -2,500 and -1,700. North row: 3 to 6 buildings each side. South row: 8 to 14. Pedestrians: 40 to 70 per sidewalk. Vehicles: 10 to 16.
    \end{tcolorbox}
\end{minipage}
\hfill
\begin{minipage}[t]{0.31\linewidth}
    \vspace{0pt}
    \centering
    \includegraphics[width=\linewidth]{Figures/original_scene.png}
    \captionof*{figure}{\small Original Scene}
\end{minipage}

\vspace{8pt}

\begin{minipage}[t]{0.31\linewidth}
    \centering
    \includegraphics[width=\linewidth]{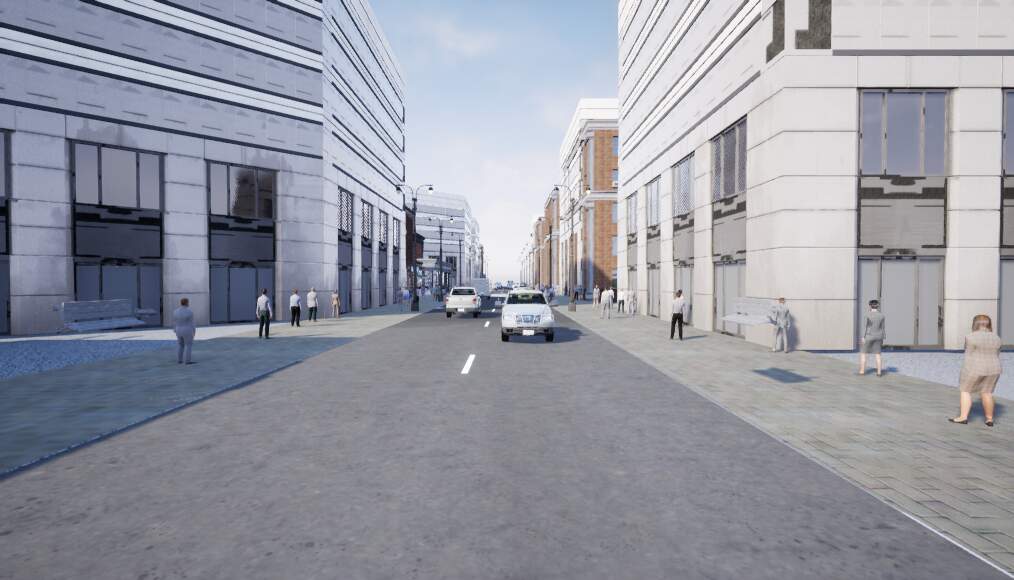}
    \captionof*{figure}{\small Claude Opus 4.7}
\end{minipage}
\hfill
\begin{minipage}[t]{0.31\linewidth}
    \centering
    \includegraphics[width=\linewidth]{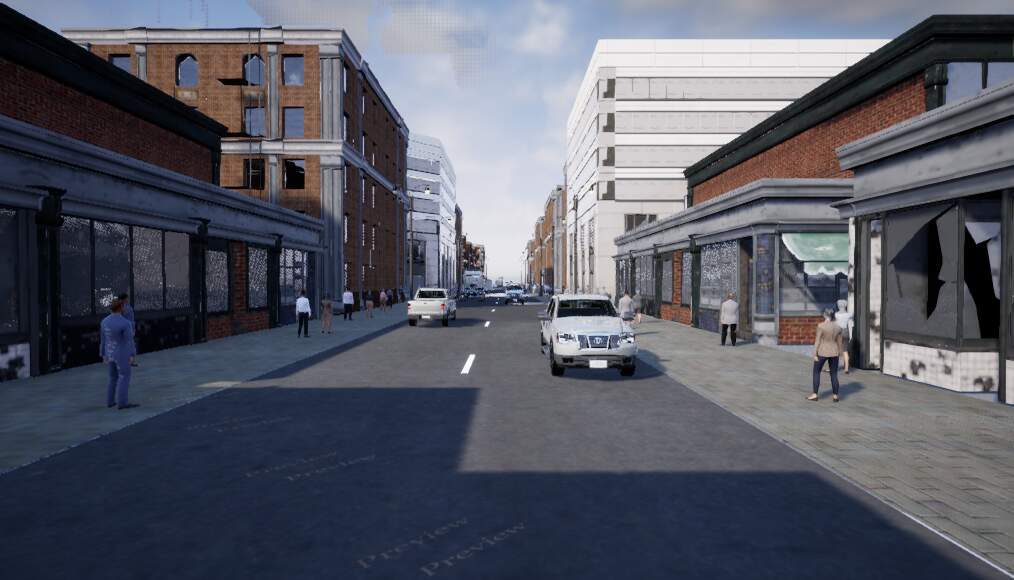}
    \captionof*{figure}{\small Qwen 3.5-27B}
\end{minipage}
\hfill
\begin{minipage}[t]{0.31\linewidth}
    \centering
    \includegraphics[width=\linewidth]{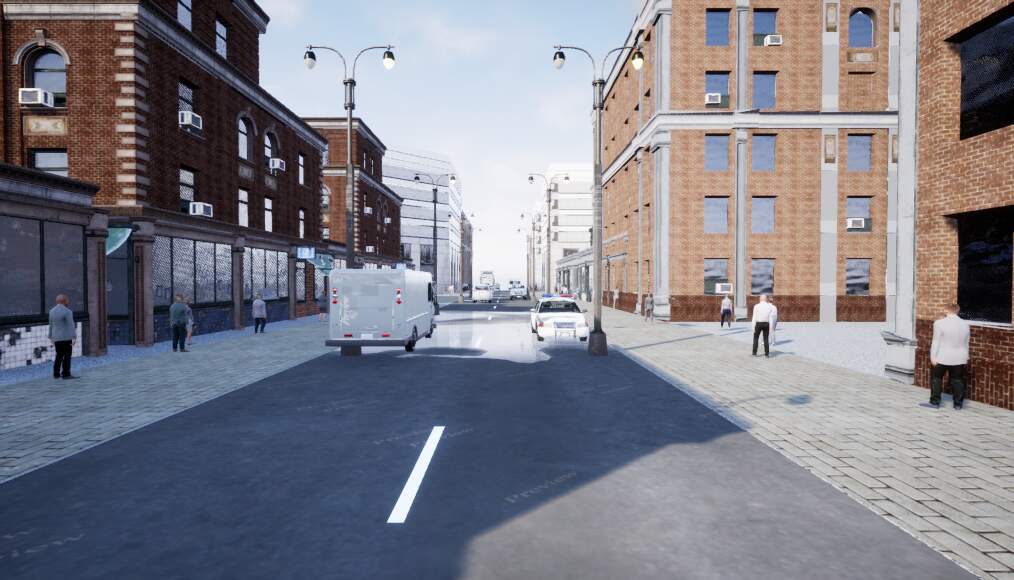}
    \captionof*{figure}{\small Qwen 3.5-9B}
\end{minipage}

\caption{\textbf{Qualitative Example P2.} Top: editing prompt and the original scene prior to modification. Bottom: rendered UE5 screenshots showing each model's edited scene, built on top of the same starting configuration.}
\label{fig:f3_p2_qualitative}
\end{figure}

\newpage
\subsection{Iterative Scene Development}
\begin{figure}[htbp]
    \centering
    \includegraphics[width=\linewidth]{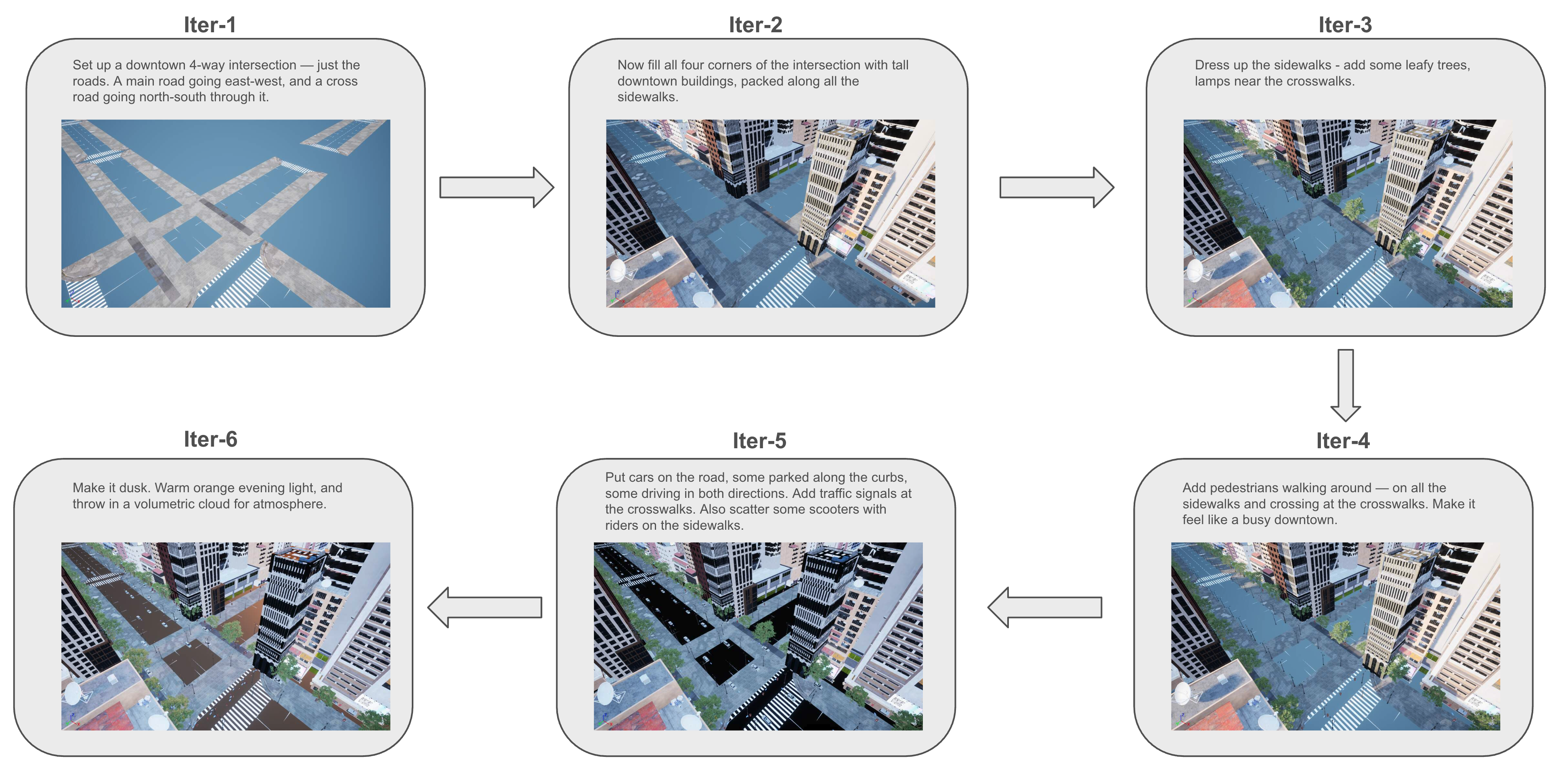}
    \caption{\textbf{Iterative scene development over six steps.} Starting from a bare 4-way road intersection (Iter-1), the scene is progressively enriched through a sequence of natural language editing instructions: tall downtown buildings are added at each corner (Iter-2), sidewalks are dressed with trees and lamps (Iter-3), pedestrians populate the crosswalks and sidewalks (Iter-4), cars, scooters, and traffic signals are introduced (Iter-5), and finally the lighting is shifted to dusk with warm orange tones and volumetric clouds (Iter-6). Each iteration modifies only what is specified without rebuilding the scene from scratch, demonstrating the platform's ability to support compositional, instruction-driven scene development.}
    \label{fig:f4_iterative_editing}
\end{figure}

